\documentclass[10pt,journal,compsoc]{IEEEtran}

\usepackage{amsmath,graphicx}
\usepackage{amsthm}
\usepackage{amssymb}
\usepackage{psfrag,subfig,comment,array}
\usepackage{latexsym}
\usepackage{multicol,xspace}
\usepackage{float}
\usepackage{multirow}
\usepackage{rotating}
\usepackage{bm}
\usepackage{tabularx}
\usepackage{mathtools}
\usepackage{stfloats}
\usepackage{picture,enumitem}
\usepackage{algorithm,algpseudocode}
\usepackage{pbox}
\usepackage{standalone}

\newcommand{\eall}{\emph{et al.}}
\newcommand{\argmin}{\operatornamewithlimits{argmin}}
\newcommand{\tr}{\operatornamewithlimits{tr}}

\newcommand{\cov}{\operatornamewithlimits{cov}}

\newcommand{\E}{\mathbb{E}}
\newcommand{\V}{\mathbb{V}}

\newcommand{\sL}{\mathcal{L}}
\newcommand{\sN}{\mathcal{N}}

\newcommand{\sH}{\mathcal{H}}
\newcommand{\sI}{\mathcal{I}}

\newcommand{\sV}{\mathcal{V}}
\newcommand{\sE}{\mathcal{E}}

\newcommand{\sO}{\mathcal{O}}
\renewcommand{\arraystretch}{1.5}

\allowdisplaybreaks[4]

\makeatletter
\long\def\@IEEEtitleabstractindextextbox#1{\parbox{0.922\textwidth}{#1}}

\begin{document}

\title{Efficient Variational Bayes Learning of Graphical Models with Smooth Structural Changes}

\author{Hang Yu,~\IEEEmembership{Member,~IEEE,}
        Songwei Wu,
        and Justin Dauwels,~\IEEEmembership{Senior Member,~IEEE}
\thanks{Part of the material in this paper was presented at IEEE 26th International Workshop on Machine Learning for Signal Processing (MLSP), Salerno, Italy, Sept. 2016~\cite{YD:2016}. This work is supported by MOE (Singapore) Tier 2 project MOE2017-T2-2-126.}
\thanks{H. Yu is with Ant Group, China. This work was mainly done when he was with Nanyang Technological University, Singapore (e-mail: HYU1@e.ntu.edu.sg).}
\thanks{S. Wu is with School of Electrical and Electronic Engineering, Nanyang Technological University, Singapore (e-mail: wuso002@e.ntu.edu.sg).}
\thanks{J. Dauwels is with Electrical Engineering, Mathematics \& Computer Science, Delft University of Technology, Netherlands (e-mail: J.H.G.Dauwels@tudelft.nl).}
}


%

\IEEEtitleabstractindextext{%
\begin{abstract}

Estimating a sequence of dynamic undirected graphical models, in which adjacent graphs share similar structures, is of paramount importance in various social, financial, biological, and engineering systems, since the evolution of such networks can be utilized for example to spot trends, detect anomalies, predict vulnerability, and evaluate the impact of interventions. Existing methods for learning dynamic graphical models require the tuning parameters that control the graph sparsity and the temporal smoothness to be selected via brute-force grid search. Furthermore, these methods are computationally burdensome with time complexity $\sO(NP^3)$ for $P$ variables and $N$ time points. As a remedy, we propose a low-complexity tuning-free Bayesian approach, named BASS. Specifically, we impose temporally dependent spike and slab priors on the graphs such that they are sparse and varying smoothly across time. An efficient variational inference algorithm based on natural gradients is then derived to learn the graph structures from the data in an automatic manner. Owing to the pseudo-likelihood and the mean-field approximation, the time complexity of BASS is only $\sO(NP^2)$. To cope with the local maxima problem of variational inference, we resort to simulated annealing and propose a method based on bootstrapping of the observations to generate the annealing noise. We provide numerical evidence that BASS outperforms existing methods on synthetic data in terms of structure estimation, while being more efficient especially when the dimension $P$ becomes high. We further apply the approach to the stock return data of 78 banks from 2005 to 2013 and find that the number of edges in the financial network as a function of time contains three peaks, in coincidence with the 2008 global financial crisis and the two subsequent European debt crisis. On the other hand, by identifying the frequency-domain resemblance to the time-varying graphical models, we show that BASS can be extended to learning frequency-varying inverse spectral density matrices, and further yields graphical models for multivariate stationary time series. As an illustration, we analyze scalp EEG signals of patients at the early stages of Alzheimer's disease (AD) and show that the brain networks extracted by BASS can better distinguish between the patients and the healthy controls.

\end{abstract}
}

\maketitle
\IEEEdisplaynontitleabstractindextext
\IEEEpeerreviewmaketitle

\section{Introduction}
\label{sec:intro}

The recent decades have witnessed a rapid development of graphical models, since they provide a refined language to describe complicated systems and further facilitate the derivation of efficient inference algorithms~\cite{KF:2009}. While an extensive literature revolves around learning static graphical models that are time-invariant (see~\cite{MB:2006}-\cite{YWXD:2020} and references therein), the change of interdependencies with a covariate (e.g. time or space) is often the rule rather than the exception for real-world data, such as friendships between individuals in a social community, communications between genes in a cell, equity trading between companies, and computer network traffic. Furthermore, such dynamic graphical models can be leveraged to spot trends, detect anomalies, classify events, evaluate the impact of interventions, and predict future behaviors of the systems. For instance, estimating time-varying functional brain networks during epileptic seizures can show how the dysrhythmia of the brain propagates, and analyzing the network evolution can help to detect epilepsy and assess the treatment of epilepsy~\cite{KEKZEC:2010}. We therefore focus on learning dynamic graphical models in this study.

In the case where all variables follow a joint Gaussian distribution, the graphical model structure is directly defined by the precision matrix (i.e., the inverse covariance matrix). Specifically, a zero element corresponds to the absence of an edge in the graphical model or, equivalently, the conditional independence between two variables. Therefore, our objective is to learn a time-varying precision matrix. Existing works on learning the time-varying precision matrices can be categorized into three groups. The first one~\cite{ZLW:2010}-\cite{QH:2016} considers the temporal dependence by smoothing the empirical covariance matrix across time using kernels. Given the temporally dependent covariance matrix, the sparse precision matrix is then estimated individually at each time point. The estimation problem can be solved by maximizing the likelihood with an $\ell_1$-norm penalty on the precision matrix. However, unexpected variability may arise between two adjacent networks since each network is estimated independently~\cite{MHSLAM:2014}. To mitigate this issue, the second group of dynamic network models~\cite{MHSLAM:2014}-\cite{HPBL:2017} further captures the temporal dependence by enforcing $\ell_1$, $\ell_2$, or Frobenius norm constraints on the difference between two consecutive precision matrices. 
As an alternative, instead of imposing separate constraints for the sparsity and the smoothness across time of the precision matrices, the third group~\cite{YP:2020} employs the local group lasso penalty to promote sparsity and smoothness together.

Unfortunately, the dynamic graphical models inferred by all three categories of methods are sensitive to the tuning parameters, including the kernel bandwidth and the penalty parameters that control the sparsity and smoothness. Classical brute-force grid search approaches for selecting these parameters are cross validation (CV), Akaike information criterion (AIC) and Bayesian information criterion (BIC)~\cite{KSAX:2010,KX:2011,MHSLAM:2014,AX:2009, WA:2015}-\cite{YP:2020}. However, heavy computational burdens come along with these methods; the learning algorithm needs to be run once for every combination of all possible values of the tuning parameters in a predefined candidate set before the one associated with the largest score is chosen. Moreover, it has been demonstrated in~\cite{LRW:2010} that these parameter selection approaches yield unsatisfactory results for graphical model selection, especially when the number of variables is large. Apart from the large number of runs for parameter selection, the computational cost in each run is also large. The time complexity of the current three groups of methods is $\sO(NP^3)$, where $P$ denotes the dimension (i.e., number of variables) and $N$ denotes the sample size. In practice, these methods are fraught with difficulties of daunting computational cost when tackling problems with more than 100 variables.

To address these problems, we propose a novel approach named BASS (BAyesian learning of graphical models with Smooth Structural changes) to learn the time-varying graphical models that is free of tuning while having a low time complexity of $\sO(NP^2)$. In particular, we focus on Gaussian graphical models, and consequently, our objective is to infer the time-varying precision matrix. To this end, we impose a temporally dependent spike and slab prior~\cite{ZS:2013,AWH:2014} on the off-diagonal entries of the precision matrix at each time point. Specifically, each off-diagonal entry of the precision matrix can be factorized as the product of a Bernoulli and a Gaussian distributed variable; the former is coupled over time via a binary Markov chain while the latter a Gauss-Markov chain (i.e., a thin-membrane model~\cite{YDJ:2014}). To facilitate the derivation of the variational inference algorithm, we replace the exact likelihood of the precision matrix at each time instant by the pseudo-likelihood that consists of the conditional distributions of one variable conditioned on the remaining variables. We then develop an efficient variational inference algorithm based on natural gradients to learn the variational distribution of the time-varying precision matrix. Due to the use of the pseudo-likelihood and the mean-field approximation in the variational inference, the time complexity of BASS is only $\sO(NP^2)$. To cope with the problem of local maxima during the variational inference, we resort to simulated annealing~\cite{GZP:2018} and propose a method based on bootstrapping to generate the annealing noise. Numerical results show that when compared with the three groups of frequentist methods, BASS achieves better performance in terms of structure estimation with significantly less amount of computational time. We further apply BASS to construct financial networks from the stock return data of 78 banks worldwide during the 2008 Great recession. We find that the network becomes denser during the crisis, with clear peaks during the Great financial crisis and each wave of the subsequent European debt crisis.

Interestingly, BASS can be extended to inferring graphical models for multiple \emph{stationary} time series in the frequency domain in a straightforward manner. Before explaining this approach, we briefly review the relevant literature on graphical models for stationary time series below. In~\cite{Dahlhaus:2000}, it is shown that for jointly Gaussian time series, the conditional independencies between time series are encoded by the common zeros in the inverse spectral density matrices at all frequencies. Given this insight, hypothesis tests are then performed in~\cite{Dahlhaus:2000}-\cite{Tugnait:2019} to test the conditional independence between every pair of time series. However, such methods are limited to problems with low dimensions and the true graphical model cannot be very sparse. On the other hand, Bach and Jordan~\cite{BJ:2004} further show that by leveraging the Whittle approximation~\cite{Whittle:1953} the Fourier transform of the time series at a certain frequency can be regarded as samples drawn from the complex Gaussian distribution whose covariance matrix is the spectral density matrix at the same frequency. As a result, an appealing approach is to first estimate the smoothed spectral density matrix given the Fourier transform of the time series and then to infer the sparse inverse spectral density matrix by maximizing the $\ell_1$ norm penalized likelihood~\cite{JHG:2015}. However, this approach requires extensive tuning. A variant of this approach for autoregressive processes is proposed in~\cite{SV:2010}. Apart from the frequentist methods, Bayesian methods have also been proposed in~\cite{TFF:2015}. Unfortunately, this method can only learn decomposable graphs from the data. It is also quite time-consuming since Monte-Carlo Markov Chain is used to learn the Bayesian model. Note that the time complexity of all aforementioned methods is at least $\sO(NP^3)$ for $P$-variate time series with length $N$. In this paper, in analogy to estimating the time-varying inverse covariance matrix, we learn the frequency-varying inverse spectral density matrix using BASS based on the Fourier transform of the time series, and then define the graphical model for the multivariate time series by identifying the common zero pattern of all inverse spectral density matrices. Different from the aforementioned methods, BASS is tuning free and scales gracefully with the dimension with time complexity $\sO(NP^2)$. We compare BASS with the frequentist method GMS (graphical model selection) proposed in~\cite{JHG:2015} on synthetic data. Similar to the results in the time domain, BASS can better recover the true graphs while being more efficient. We further apply BASS to the scalp EEG signals of patients at early stages of Alzheimer's disease (AD), and build a classifier based on the estimated graphical models to differentiate between the AD patients and age-matched control subjects.. The classification accuracy resulting from BASS is higher than that from GMS.

This paper is structured as follows. We present our Bayesian model for time-varying graphical models in Section~\ref{sec:dynmcGM} and derive the natural gradient variational inference algorithm in Section~\ref{sec:VBlearn}. We then extend the proposed model to the frequency domain to infer graphical models for stationary time series in Section~\ref{ssec:GMTS}. In Section~\ref{sec:results}, we show the numerical results for both synthetic and real data. Finally, we close this paper with conclusions in Section~\ref{sec:conclusion}.


\section{Time-varying Graphical Models}  
\label{sec:dynmcGM}

We are concerned with undirected graphical models $\mathcal G = (\mathcal V, \mathcal E)$ in this paper, where $\mathcal V$ denotes a set of vertices relating to variables and $\mathcal E$ denotes the edge set that encodes the conditional dependencies between the variables. Each node $j\in\mathcal{V}$ is associated with a random variable $x_j$. An edge $(j,k)\in\mathcal{E}$ is absent if and only if the corresponding two variables $x_j$ and $x_k$ are conditionally independent: $p(x_j,x_k|x_{-jk})=p(x_j|x_{-jk})p(x_k|x_{-jk})$, where $-jk$ denotes all the nodes in $\mathcal V$ except $j$ and $k$. When all variables $\bm x = [x_1,\cdots,x_P]'$ are jointly Gaussian distributed, the resulting graphical model is referred to as a Gaussian graphical model. Let $\mathcal N(\bm\mu, \Sigma)$ denote a Gaussian distribution with mean $\bm\mu$ and covariance $\Sigma$. The distribution can be equivalently parameterized as $\mathcal N(K^{-1}\bm h, K^{-1})$, where $K = \Sigma^{-1}$ is the precision matrix (the inverse covariance) and $\bm h = K\bm\mu$ is the potential vector. The density function is:
\begin{align} \label{eq:GGM_PDF_complete}
p(\bm x)\propto \det(K)^{\frac{1}{2}} \exp\Big(-\frac{1}{2}\bm x'K\bm x+\bm h'\bm x\Big),
\end{align}
where $\bm x'$ denotes the transpose of $\bm x$.
Under this scenario, the conditional dependencies are characterized by the precision matrix, that is, $x_j$ and $x_k$ are conditionally independent if and only if $K_{jk} = 0$. As a result, for Gaussian graphical models, we target at inferring a sparse precision matrix. 

For time-varying graphical models, we assume that the observation $\bm x^{(t)}$ at time $t$ is drawn from the graphical model with precision matrix $K^{(t)}$ for $t = 1, \cdots, N$, and $K^{(t)}$ changes smoothly with $t$. Without loss of generality, we further assume that $\bm\mu^{(t)} = 0$ in our model, and so $\bm h^{(t)} = 0$. The likelihood of $K^{(t)}$ can then be expressed as:
\begin{align} \label{eq:exact_llhd}
p(\bm x^{(t)}|K^{(t)}) \propto \det(K^{(t)})^{\frac{1}{2}}\exp\Big(-\frac{1}{2}{{\bm x}^{(t)}}'K^{(t)}{\bm x}^{(t)}\Big).
\end{align}

To facilitate the derivation of the variational inference algorithm, we propose to use the pseudo-likelihood instead of the exact likelihood~\eqref{eq:exact_llhd} in the Bayesian formulation. More specifically, the pseudo-likelihood is derived from the conditional distributions of one variable $x_j$ conditioned on the remaining variables $\bm x_{-j}$:,
\begin{align}
& p(x_j^{(t)}|\bm x_{-j}^{(t)}, K_{jj}^{(t)}, K_{j,-j}^{(t)}) \notag \\
\propto&\ \sqrt{K_{jj}^{(t)}}\exp\Big[-\frac{1}{2}K_{jj}^{(t)}(x_j^{(t)} + {K_{jj}^{(t)}}^{-1}K_{j,-j}^{(t)}\bm x_{-j}^{(t)})^2\Big], \notag \\
\propto&\ \sqrt{K_{jj}^{(t)}}\exp\Big[-\frac{1}{2}K_{jj}^{(t)}{x_j^{(t)}}^2 - x_j^{(t)} K_{j,-j}^{(t)}\bm x_{-j}^{(t)} \notag \\
&\, - \frac{1}{2}{K_{jj}^{(t)}}^{-1}(K_{j,-j}^{(t)}\bm x_{-j}^{(t)})^2\Big], \label{eq:pseudo_llhd}
\end{align}
where $K_{j,-j}$ denotes row $j$ in $K$ excluding $K_{jj}$, and $-{K_{jj}^{(t)}}^{-1}K_{j,-j}^{(t)}\bm x_{-j}^{(t)}$ and ${K_{jj}^{(t)}}^{-1}$ are respectively the mean and the variance of the conditional distribution $p(x_j|\bm x_{-j})$. Here, we regard~\eqref{eq:pseudo_llhd} as a Gaussian distribution of $x_j$ whose mean and variance are parameterized by $K_{jj}^{(t)}$ and $K_{j,-j}^{(t)}$. In other words, it is a likelihood function of $K_{jj}^{(t)}$ and $K_{j,-j}^{(t)}$. This pseudo-likelihood of $K$ has been frequently explored in the literature~\cite{MB:2006}-\cite{LW:2017},~\cite{AX:2009,YP:2020}, for Gaussian graphical model selection. It simplifies the determinant term in~\eqref{eq:exact_llhd} that typically leads to heavy computational burden of $\sO(P^3)$, and so improves the computational efficiency. Indeed, the time complexity of the proposed method BASS is only $\sO(P^2)$ w.r.t. (with regard to) $P$, owing to the pseudo-likelihood. Furthermore, the pseudo-likelihood typically results in more accurate and robust results when learning the graph structure~\cite{ODKB:2014, LW:2017,YP:2020}. Different from previous works~\cite{MB:2006}-\cite{LW:2017},~\cite{YP:2020} that allow $K_{jk}$ to be different from $K_{kj}$, we assume that $K_{jk}=K_{kj}$ in our paper. Additionally, we infer the distribution of the diagonal elements $K_{jj}$ explicitly from the observed data $\bm x$ rather than setting $K_{jj} = 1$ as in~\cite{MB:2006}-\cite{LW:2017},~\cite{YP:2020}. For compactness in notation, we denote the distribution in~\eqref{eq:pseudo_llhd} as $p(x_j^{(t)}| K_{jj}^{(t)}, K_{j,-j}^{(t)})$ in the sequel.

Next, we impose priors on both $K_{jj}^{(t)}$ and $K_{j,-j}^{(t)}$ in order to construct a full Bayesian model. We first focus on the off-diagonal elements $K_{jk}^{(t)}$. To guarantee that the off-diagonal parts of the precision matrices $K^{(t)}$ are sparse while varying smoothly across time, we resort to the temporally dependent spike and slab prior~\cite{ZS:2013,AWH:2014}. Concretely, a spike and slab prior on $K_{jk}^{(t)}$ can be defined as~\cite{IR:2005}:
\begin{align}
K_{jk}^{(t)} \sim \pi_{jk}^{(t)} \mathcal N(\mu_{jk}^{(t)},\nu_{jk}^{(t)}) + (1-\pi_{jk}^{(t)})\delta_0,
\end{align}
where $\mathcal N(\mu_{jk}^{(t)},\nu_{jk}^{(t)})$ is a Gaussian distribution with mean $\mu_{jk}^{(t)}$ and variance $\nu_{jk}^{(t)}$, $\delta_0$ is a Kronecker delta function, and $\pi_{jk}^t \in [0,1]$ determines the probability of $K_{jk}^{(t)} = 0$ (i.e., the spike probability). By decreasing $\pi_{jk}^{(t)}$ to $0$, this prior would shrink $K_{jk}^{(t)}$ to $0$, thus encouraging sparsity in $K^{(t)}$. The above expression can also be equivalently written as~\cite{ZS:2013}:
\begin{align}
K_{jk}^{(t)} &= s_{jk}^{(t)} J_{jk}^{(t)} \label{eq:equality}\\
J_{jk}^{(t)} &\sim \mathcal N(\mu_{jk}^{(t)},\nu_{jk}^{(t)}), \\
s_{jk}^{(t)} &\sim \text{Ber}(\pi_{jk}^{(t)}),
\end{align}
where $\text{Ber}(\pi_{jk}^{(t)})$ is a Bernoulli distribution with success probability $\pi_{jk}^{(t)}$. To obtain $K^{(t)}$ that changes smoothly with $t$, we impose smoothness priors on both $s_{jk}^{(t)}$ and $J_{jk}^{(t)}$. For $s_{jk}^{(t)}$, we assume that it is drawn a binary Markov chain defined by the initial state and the transition probabilities:
\begin{align}
p(\bm s_{jk}^{(1:N)}) = p(s_{jk}^{(1)}) \prod_{t = 2}^N p(s_{jk}^{(t)}|s_{jk}^{(t-1)}),
\end{align}
where
\begin{align}
&p(s_{jk}^{(1)}) = \pi_1^{\delta(s_{jk}^{(1)}=1)}\big(1-\pi_1\big)^{\delta(s_{jk}^{(1)}=0)}, \\
&p(s_{jk}^{(t)}|s_{jk}^{(t-1)}) = A_{00}^{\delta(s_{jk}^{(t-1)} = 0, s_{jk}^{(t)} = 0)} (1 - A_{00})_{}^{\delta(s_{jk}^{(t-1)} = 0, s_{jk}^{(t)} = 1)} \notag \\
&\cdot A_{11}^{\delta(s_{jk}^{(t-1)} = 1, s_{jk}^{(t)} = 1)} (1 - A_{11})_{}^{\delta(s_{jk}^{(t-1)} = 1, s_{jk}^{(t)} = 0)},
\end{align}
and $\delta(\cdot)$ denotes the indicator function that yields $1$ when the condition in the bracket is satisfied and $0$ otherwise. We further assume $\pi_1$, $A_{00}$, and $A_{11}$ follow uniform distributions $\text{Be}(1,1)$, where $\text{Be}(1,1)$ denotes a Beta distribution with shape parameters one.

\begin{figure}[t]
\centering
\includegraphics[width=0.67\columnwidth,origin=c]{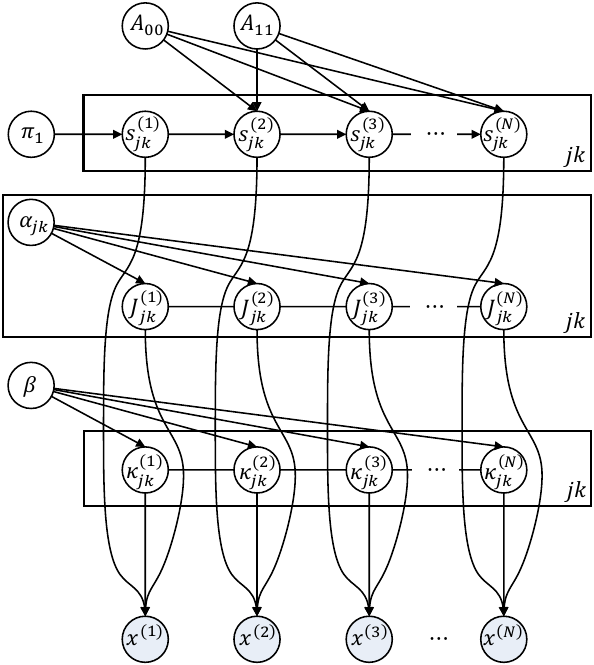}
\vspace{-1mm}
\caption{Graph representation of BASS.}
\label{fig:BASS}
\vspace{-6mm}
\end{figure}

On the other hand, we assume that $J_{jk}^{(t)}$ forms a Gauss-Markov chain, in particular, a thin-membrane model~\cite{YDJ:2014}. The resulting prior on $J_{jk}^{(t)}$ can be expressed as:
\begin{align}
p(J_{jk}^{(1:N)}) &\propto \alpha_{jk}^{\frac{N - 1}{2}}\exp\Big[-\frac{\alpha_{jk}}{2}\sum_{t=2}^N(J_{jk}^{(t)} - J_{jk}^{(t-1)})^2\Big],\notag \\
&\propto \alpha_{jk}^{\frac{N - 1}{2}}\exp(-\frac{\alpha_{jk}}{2}{J_{jk}^{(1:N)}}' K_{\text{TM}}J_{jk}^{(1:N)}),
\end{align}
where $\alpha_{jk}$ is the smoothness parameter and $\alpha_{jk}K_{\text{TM}}$ is the precision matrix of this Gaussian graphical model. We further impose a non-informative Jeffrey's prior on $\alpha_{jk}$, that is, $p(\alpha_{jk}) \propto 1 / \alpha_{jk}$. The difference between $J_{jk}^{(t-1)}$ and $J_{jk}^{(t)}$ at every two consecutive time points $t-1$ and $t$ is controlled by the smoothness parameter $\alpha_{jk}$, suggesting that $\alpha_{jk}$ determines the smoothness of $J_{jk}^{(t)}$ across $t$. We also notice that $K_{\text{TM}}$ is the graph Laplacian matrix corresponding to the Markov chain: the diagonal entry $[K_{\text{TM}}]_{jj}$ equals the number of neighbors of node $j$, while the off-diagonal entry $[K_{\text{TM}}]_{jk}$ equals $-1$ if node $j$ and $k$ are adjacent and $0$ otherwise. As a result, $K_{\text{TM}}$ is a tri-diagonal matrix in our case. Furthermore, it follows from the properties of the Laplacian matrix that $K_{\text{TM}}\bm 1 = 0$, where $\bm 1$ denotes a vector of all ones. In other words, the thin-membrane model is invariant to the addition of $c\bm 1$, where $c$ is an arbitrary constant, and it allows the deviation from any overall mean level without having to specify the overall mean level itself. Such desirable properties make the thin-membrane model a popular smoothness prior in practice.

For diagonal entries in the time-varying precision matrix, since they can only take positive values, we reparameterize $K_{jj}^{(t)}$ as $K_{jj}^{(t)} = \exp(\kappa_j^{(t)})$. To promote the smooth variation of $\kappa_j^{(t)}$ across $t$, we assume that $\bm \kappa_j^{(1:N)}$ follows a thin-membrane model with smoothness parameter $\beta$. We also impose the Jeffrey's prior on $\beta$.

Altogether, the proposed Bayesian model is summarized as a graphical model in Fig.~\ref{fig:BASS}. The joint distribution of all variables can be factorized as:
\begin{align}
&\, p(\bm x^{(1:N)}, \bm s^{(1:N)}, J^{(1:N)}, \bm\kappa^{(1:N)}, \pi_1, A_{00}, A_{11}, \bm\alpha, \beta) \notag \\
=&\, \prod_{j = 1}^P \prod_{t=1}^N p(x_j^{(t)}|\kappa_j^{(t)}, J_{j,-j}^{(t)}, \bm s_{j, -j}^{(t)})  \notag \\
&\, \cdot\prod_{j = 1}^P \prod_{k = j+1}^P \Big[p(\bm s_{jk}^{(1:N)}|\pi_1, A_{00}, A_{11}) p(J_{jk}^{(1:N)}|\alpha_{jk}) p(\alpha_{jk})\Big] \notag \\
&\, \cdot\prod_{j = 1}^P p(\bm\kappa_j^{(1:N)}|\beta) p(\pi_1) p(A_{00}) p(A_{11}) p(\beta). \label{eq:joint_dist}
\end{align}


\section{Variational Inference}
\label{sec:VBlearn}

In this section, we develop a variational inference algorithm to learn the above Bayesian model. We first derive the low-complexity variational inference algorithm for BASS. Since the variational inference algorithm is often sensitive to local maxima, we further present how to utilize simulated annealing to help the algorithm escape from local maxima. We refer the readers to the summary of variational inference and natural gradients in Section A in the appendix before diving into the calculations in the sequel.

\subsection{Variational Inference for BASS}
\label{ssec:vi}

Our objective is to approximate the intractable posterior distribution $p(\bm s^{(1:N)}, J^{(1:N)}, \bm\kappa^{(1:N)}, \pi_1, A_{00}, A_{11}, \bm\alpha, \beta|\bm x^{(1:N)})$ by a tractable variational distribution, that is, $q(\bm s^{(1:N)}, J^{(1:N)}, \bm\kappa^{(1:N)}, \pi_1, A_{00}, A_{11}, \bm\alpha, \beta)$. Specifically, we apply the mean-field approximation and factorize the variational distribution as:
\begin{align}
&\, q(\bm s^{(1:N)}, J^{(1:N)}, \bm\kappa^{(1:N)}, \pi_1, A_{00}, A_{11}, \bm\alpha, \beta) \notag \\
=&\, \prod_{j=1}^P\prod_{k = j+1}^P \Big[q(\bm s_{jk}^{(1:N)}) q(J_{jk}^{(1:N)})q(\alpha_{jk})\Big] \prod_{j=1}^P q(\bm\kappa_j^{(1:N)})q(\pi_1)\notag \\
&\, q(A_{00})q(A_{11})q(\beta).
\end{align}
For ease of exposition, we present the proposed algorithm from the perspective of message passing~\cite{Bishop:2006}. In a hierarchical Bayesian model, the update for each node only requires information (i.e., messages) from the nodes in its Markov blanket, including this node's parents, children, and co-parents of its children.

\begin{figure}[t]
\vspace{-2mm}
\centering
\subfloat[]{
\includegraphics[height = 5.8cm,origin=c]{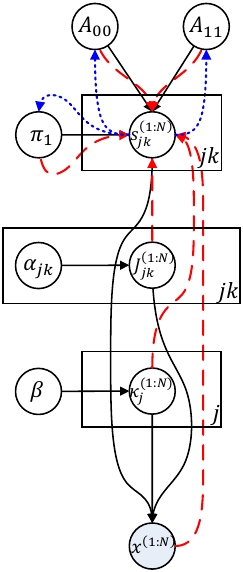}
\label{sfig:message_passing_s}}
\hfill
\subfloat[]{
\includegraphics[height = 5.8cm,origin=c]{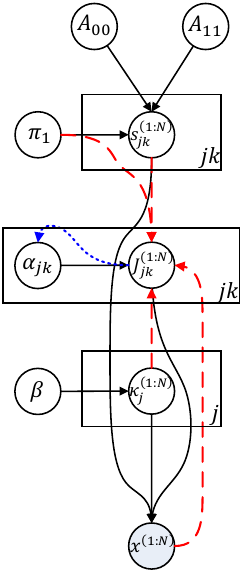}
\label{sfig:message_passing_J}}
\hfill
\subfloat[]{
\includegraphics[height = 5.8cm,origin=c]{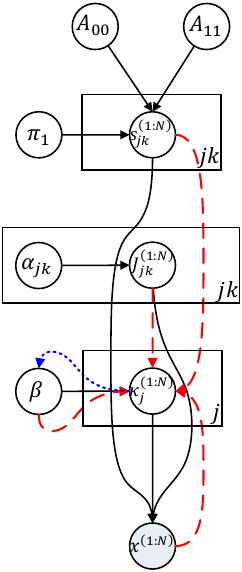}
\label{sfig:message_passing_kappa}}
\vspace{-1mm}
\caption{Message passing scheme for updating (a) $q(\bm s_{jk}^{(1:N)})$ (the red dashed arrows), $q(\pi_1)$, $q(A_{00})$, and $q(A_{11})$ (the blue dotted arrows), (b) $q(\bm J_{jk}^{(1:N)})$ (the red dashed arrows) and $q(\alpha_{jk})$ (the blue dotted arrow), and (c) $q(\bm\kappa_{j}^{(1:N)})$ (the red dashed arrows) and $q(\beta)$ (the blue dotted arrow).}
\label{fig:message_passing}
\vspace{-5mm}
\end{figure}

Given the Bayesian model in Fig.~\ref{fig:BASS}, the update rule of $q(\bm s_{jk}^{(1:N)})$ can be derived by passing messages in the directions of the red dashed arrows in Fig.~\ref{sfig:message_passing_s}, resulting in:
\begin{align}
q(\bm s_{jk}^{(1:N)}) \propto \exp\bigg[\sum_{t=1}^N\varphi_\sV^s(s_{jk}^{(t)}) + \sum_{t=2}^N\varphi_\sE^s(s_{jk}^{(t)}, s_{jk}^{(t-1)})\bigg], \label{eq:q_s}
\end{align}
where the node potential $\varphi_\sV^s$ and edge potential $\varphi_\sE^s$ are defined in (S.14)\footnote{S stands for equations, tables, and figures in the appendix.} and (S.15) in Table~S.1 in the appendix that are functions of those nodes in the Markov blanket of $\bm s_{jk}^{(1:N)}$. Moreover, it is apparent that $q(\bm s_{jk}^{(1:N)})$ can be factorized as a binary Markov Chain. As such, the marginal and pairwise densities, $q(s_{jk}^{(t)})$ and $q(s_{jk}^{(t)}, s_{jk}^{(t-1)})$, can then be computed via message passing in the binary Markov chain (i.e., the forward-backward algorithm) with time complexity $\sO(N)$. Given $q(s_{jk}^{(t)})$ and $q(s_{jk}^{(t)}, s_{jk}^{(t-1)})$, the variational distributions of the initial state $\pi_1$ and the transition probabilities $A_{00}$ and $A_{11}$ can be derived by receiving messages from $s_{jk}^{(t)}$ (see blue dotted arrows in Fig.~\ref{sfig:message_passing_s}):
\begin{align}
q(\pi_1) =&\, \text{Be}(a, b), \label{eq:q_pi1} \\
q(A_{00}) =&\, \text{Be}(c_0, d_0), \\
q(A_{11}) =&\, \text{Be}(c_1, d_1). \label{eq:q_A11}
\end{align}
where the shape parameters $a$, $b$, $c_0$, $d_0$, $c_1$, and $d_1$ of the Beta distributions can be updated as in~(S.22)-(S.27).  

Similarly, as denoted by the red dashed arrows in Fig.~\ref{sfig:message_passing_J}, $q(J_{jk}^{(1:N)})$ can be updated by collecting messages from nodes in its Markov blanket, that is,
\begin{align}
q(J_{jk}^{(1:N)}) \propto \exp\bigg[\sum_{t=1}^N\varphi_\sV^J(J_{jk}^{(t)}) + \sum_{t=2}^N\varphi_\sE^J(J_{jk}^{(t)}, J_{jk}^{(t-1)})\bigg],  \label{eq:q_J}
\end{align}
where the node potentials $\varphi_\sV^J(J_{jk}^{(t)})$ and edge potentials $\varphi_\sE^J(J_{jk}^{(t)}, J_{jk}^{(t-1)})$ are defined in~(S.16) and~(S.17) in the appendix. The above expression corresponds to a Gauss-Markov chain, and therefore, the mean and variance for each $J_{jk}^{(t)}$ and the pairwise covariance of $J_{jk}^{(t)}$ and $J_{jk}^{(t-1)}$ can be obtained via message passing (a.k.a. belief propagation) with complexity $\sO(N)$. We can then update the variational distribution of $\alpha_{jk}$ by receiving messages from $J_{jk}^{(1:N)}$ (see blue dotted arrows in Fig.~\ref{sfig:message_passing_J}) as:
\begin{align}
q(\alpha_{jk}) = \text{Ga}\bigg(\frac{N-1}{2},\frac{\sum_{t=2}^N\langle (J_{jk}^{(t)}-J_{jk}^{(t-1)})^2\rangle}{2}\bigg),
\end{align}
where $\text{Ga}(a,b)$ denotes a Gamma distribution with shape parameter $a$ and rate parameter $b$. 

Finally, let us turn our attention to $\bm\kappa^{(1:N)}$. The message updates for $\bm\kappa^{(1:N)}_j$ are indicated by the red dashed arrows in Fig.~\ref{sfig:message_passing_kappa}. Note that the prior and likelihood are not conjugate in this case. As mentioned in the Section A in the appendix, we need to specify the functional form of $q(\bm\kappa_j^{(1:N)})$, compute the natural gradients, and then update the natural parameters of $q(\bm\kappa_j^{(1:N)})$ following~\eqref{eq:nonconj_update_rule}. Here, we choose $q(\bm\kappa_j^{(1:N)})$ to be Gaussian. Owing to the thin-membrane priors on $\bm\kappa_j^{(1:N)}$, $q(\bm\kappa_j^{(1:N)})$ is also associated with a Gauss-Markov chain, in the same fashion as $q(J_{jk}^{(1:N)})$. Therefore, we can parameterize $q(\bm\kappa_j^{(1:N)})$ as:
\begin{align}
&q(\bm\kappa_j^{(1:N)}) \propto \exp\bigg[\sum_{t=1}^N\varphi_\sV^\kappa(\kappa_j^{(t)}) + \sum_{t=2}^N\varphi_\sE^\kappa(\kappa_j^{(t)}, \kappa_j^{(t-1)})\bigg] \notag \\
&\propto \exp\bigg[\sum_{t=1}^N \Big(-\frac{\Omega_{t,t}}{2} {\kappa_j^{(t)}}^2 + h_t \kappa_j^{(t)}\Big) - \sum_{t=2}^N \Omega_{t, t-1} \kappa_j^{(t)}\kappa_j^{(t-1)}\bigg], \label{eq:q_kappa}
\end{align}
where $\Omega$ denotes the $N\times N$ tri-diagonal precision matrix and $\bm h$ is the $N$-dimensional potential vector. In light of~\eqref{eq:nonconj_update_rule}, $\Omega$ and $\bm h$ can be updated as in~(S.19)-(S.21). Note that the step size $\rho$ in~(S.19)-(S.21) is chosen via line search. After obtaining the mean, the variance and the pairwise covariance of $\bm\kappa_j^{(1:N)}$ via message passing in the Gauss-Markov chain, we can update $q(\beta)$ as $q(\beta) = \text{Ga}((N - 1) P / 2, \sum_{j = 1}^P\sum_{t = 2}^N \langle(\kappa_j^{(t)} - \kappa_j^{(t-1)})^2\rangle / 2)$. Detailed derivation of the variational inference algorithm can be found in Section B in the appendix.

\subsection{Time Complexity}
We notice that the most expensive operations in the update rules in Table~S.1 are the products $\langle K_{j,-jk}^{(t)}\rangle \bm x_{-jk}^{(t)}$ and $\langle K_{k,-jk}^{(t)}\rangle \bm x_{-jk}^{(t)}$ in~(S.12) and $\langle K_{j,-j}^{(t)} \rangle\bm x_{-j}^{(t)}$ in~(S.18). The time complexity of these operations is $\sO(P)$. The last product is used for updating the diagonal element $\kappa_j^{(t)}$, and hence, the time complexity for updating all $NP$ diagonal elements in $K^{(1:N)}$ is $\sO(NP^2)$. On the other hand, the first two products are used for updating one off-diagonal element $K_{jk}^{(t)}$. Take into account all $\sO(NP^2)$ off-diagonal elements in $K^{(1:N)}$, and the overall time complexity of BASS should be $\sO(NP^3)$. However, instead of computing these products every time when updating an off-diagonal element $K_{jk}^{(t)}$, we can first keep a record of $\langle K_{j,-j}^{(t)} \rangle\bm x_{-j}^{(t)}$ for $j = 1, \cdots, P$ at the beginning of BASS. Next, for each off-diagonal element $K_{jk}^{(t)}$, the products can be computed as $\langle K_{j,-jk}^{(t)}\rangle \bm x_{-jk}^{(t)} = \langle K_{j,-j}^{(t)} \rangle\bm x_{-j}^{(t)} - \langle K_{jk}^{(t)} \rangle\bm x_k^{(t)}$ and likewise for $\langle K_{k,-jk}^{(t)}\rangle \bm x_{-jk}^{(t)}$. After updating the variational distribution of this off-diagonal element, we can then update the record as $\langle K_{j,-j}^{(t)} \rangle\bm x_{-j}^{(t)} = \langle K_{j,-jk}^{(t)}\rangle \bm x_{-jk}^{(t)} + \langle K_{jk}^{(t)} \rangle\bm x_k^{(t)}$ and likewise for $\langle K_{k,-k}^{(t)}\rangle \bm x_{-k}^{(t)}$. As a consequence, we can cycle through all off-diagonal elements in $K^{(1:N)}$ without recomputing the products every time. The resulting time complexity of BASS can be reduced to $\sO(NP^2)$.


\subsection{Simulated Annealing}
\label{ssec:sim_anneal}
As introduced in Section A in the appendix, the updated rules derived in Section~\ref{ssec:vi} can also be written in the manner of natural gradient ascent as:
\begin{align}
\bm\theta^{\{i+1\}} 
&= (1-\rho)\bm\theta^{\{i\}} + \rho \nabla_{\bm\eta}\sL_1\big(\bm\theta^{\{i\}}\big), \label{eq:nonconj_update_rule}
\end{align}
where $\bm\theta$ denotes the parameters to be updated, $\nabla_{\bm\eta}\sL_1\big(\bm\theta^{\{i\}}\big)$ denotes the natural gradient of the evidence lower bound (ELBO) w.r.t. $\bm\theta$, and $0 < \rho < 1$ is the step size. Since the ELBO is non-convex, the above variational inference algorithm suffers from the issue of local maxima. To counteract this problem, we employ simulated annealing and modify the update rule of the variational inference in~\eqref{eq:nonconj_update_rule} as~\cite{{GZP:2018}}:
\begin{align}
\tilde{{}\bm\theta} = (1-\rho)\bm\theta^{\{i\}} + \rho \Big[\frac{1}{T^{\{i\}}}\nabla_{\bm\eta}\sL_1\big(\bm\theta^{\{i\}}\big) + \big(1 - \frac{1}{T^{\{i\}}}\big)\bm\epsilon^{\{i\}}\Big], 
\end{align}
where $T^{\{i\}}$ denotes the annealing temperature in iteration $i$, and $\bm\epsilon^{\{i\}}$ denotes the annealing noise vector. 
Note that $T^{\{i\}}\to 1$ as $i\to\infty$. The update $\tilde{{}\bm\theta}$ is accepted with the probability:
\begin{align}
p(\bm\theta^{\{i+1\}} = \tilde{{}\bm\theta}) = \min\Big\{1,\exp\big[\frac{\sL(\tilde{{}\bm\theta}) - \sL(\bm\theta^{\{i\}})}{1 - 1 / T^{\{i\}}}\big]\Big\},
\end{align}
otherwise $\bm\theta^{\{i+1\}} = \bm\theta^{\{i\}}$. When the temperature $T^{\{i\}}$ is high, $\tilde{{}\bm\theta}$ is sufficiently volatile to avoid shallow local maxima. As $T^{\{i\}}$ decreases to $1$ , the algorithm mimics the original variational inference and converges.

Next, we discuss how to build the noise vector $\bm\epsilon^{\{i\}}$. Currently, there is no generic rule for specifying the noise distribution. In this work, we only add noise when updating the natural parameters of $q(\bm s_{jk}^{(1:N)})$, $q( J_{jk}^{(1:N)})$, and $q(\bm \kappa_j^{(1:N)})$, since the update rules for these parameters are more complicated than the others and so the resulting estimates are more likely to converge to local maxima. Let $\nabla_{\bm\eta}\sL_1(\bm\theta^{\{i\}})$ denote the natural gradient of $\sL_1$ in~(S.11) w.r.t. the natural parameters $\bm\theta$ of $q(\bm s_{jk}^{(1:N)})$, $q( J_{jk}^{(1:N)})$, and $q(\bm \kappa_j^{(1:N)})$. It can be observed from the corresponding update rules (cf.~Eqs.~(S.12)-(S.21) in Table~S.1) that $\nabla_{\bm\eta}\sL_1(\bm\theta^{\{i\}})$ can be decomposed into two terms: $\nabla_{\bm\eta}\sL_1(\bm\theta^{\{i\}}) = \nabla_{\bm\eta}\sL_1(\bm\theta^{\{i\}}, \bm x^{(1:N)}) + \nabla_{\bm\eta}\sL_1(\bm\theta^{\{i\}}, \bm \xi)$, where $\nabla_{\bm\eta}\sL_1(\bm\theta^{\{i\}}, \bm x^{(1:N)})$ is a function of the observations $\bm x^{(1:N)}$, and $\nabla_{\bm\eta}\sL_1(\bm\theta^{\{i\}}, \bm\xi)$ is a function of the hyperparameters $\bm \xi = \{\pi_1, A_{00}, A_{11}, \bm\alpha, \beta\}$. Correspondingly, we decompose the noise vector as $\bm\epsilon^{\{i\}} = \bm\epsilon^{\{i\}}(\bm x^{(1:N)}) + \bm\epsilon^{\{i\}}(\bm\xi)$.

For the hyperparameters $\bm\xi$, we utilize their variational distributions to obtain the noise $\bm\epsilon^{\{i\}}(\bm\xi)$. Concretely, we draw a random sample $\tilde{\bm\xi}$ of $\bm \xi$ from the variational distribution in each iteration, and compute the noise vector as $\bm\epsilon^{\{i\}}(\bm\xi) = \nabla_{\bm\eta}\sL_1(\bm\theta^{\{i\}}, \tilde{\bm\xi})$. On the other hand, for the function of the observations, let $\nabla_{\bm\eta}\tilde\sL_1(\bm\theta^{\{i\}}, \bm x^{(1:N)}) = \nabla_{\bm\eta}\sL_1(\bm\theta^{\{i\}}, \bm x^{(1:N)})/T^{\{i\}} + (1 - 1/T^{\{i\}})\bm\epsilon^{\{i\}}(\bm x^{(1:N)})$ denote the noisy gradient that is corrupted by the annealing noise. As $T^{\{i\}}\to 1$, $\nabla_{\bm\eta}\tilde\sL_1(\bm\theta^{\{i\}}, \bm x^{(1:N)})$ becomes less noisy and converges to the exact gradient $\nabla_{\bm\eta}\sL_1(\bm\theta^{\{i\}}, \bm x^{(1:N)})$. Here, instead of proposing a distribution for $\bm\epsilon^{\{i\}}(\bm x^{(1:N)})$, we specify the noisy gradient $\nabla_{\bm\eta}\tilde\sL_1(\bm\theta^{\{i\}}, \bm x^{(1:N)})$ directly by replacing the observations $\bm x^{(1:N)}$ in $\nabla_{\bm\eta}\sL_1(\bm\theta^{\{i\}}, \bm x^{(1:N)})$ with its bootstrapped sample set $\tilde{\bm x}^{(1:N)}$. Such bootstrapped sets are often used for time-invariant graphical model selection~\cite{LRW:2010,MB:2010,LHPW:2013} so as to find a network that is robust to bootstrapping. We borrow this idea and provide an empirical distribution for $\bm x^{(1:N)}$ by bootstrapping the original observations. More specifically, for each time point $t$, we set $\tilde{\bm x}^{(t)} = \bm x^{(\tau)}$ by sampling $\tau$ uniformly from a window around $t$ with width $w$, namely, $\{t - w, t - w + 1, \cdots, t + w\}$. The noisy gradient can then be computed as $\nabla_{\bm\eta}\tilde\sL_1(\bm\theta^{\{i\}}, \bm x^{(1:N)}) = \nabla_{\bm\eta}\sL_1(\bm\theta^{\{i\}}, \tilde{\bm x}^{(1:N)})$. Furthermore, we set $w = (1 - 1 / T^{\{i\}}) N / 2$ such that the variance of the noisy gradient $\nabla_{\bm\eta}\tilde\sL_1(\bm\theta^{\{i\}}, \bm x^{(1:N)})$ decreases with the decreases of $T$ and $\nabla_{\bm\eta}\tilde\sL_1(\bm\theta^{\{i\}}, \bm x^{(1:N)})$ converges to the exact gradient $\nabla_{\bm\eta}\sL_1(\bm\theta^{\{i\}}, \bm x^{(1:N)})$ as $T^{\{i\}}\to 1$.

In practice, we increase $R^{\{i\}} = 1 / T^{\{i\}}$ instead of decreasing $T^{\{i\}}$ as $i$ increases. Concretely, we first specify the number of iterations for annealing as $N_a$. We then begin the algorithm with $R^{\{1\}} = 0$ (i.e., $T^{\{1\}} = \infty$) and increases $R^{\{i\}}$ by $10 / N_a$ in every 10th iteration. After $N_a$ iterations, $R^{\{N_a\}} = 1$ (i.e., $T^{\{N_a\}} = 1$). In our experiments, we set $N_a = 500$ unless stated otherwise. A brief summary of the proposed BASS algorithm is presented in Algorithm 1. For the detailed implementation of the proposed simulated annealing technique, we refer the readers to Algorithm S.1 in the appendix.

\begin{algorithm}[t]
\footnotesize
  \caption{BASS} \label{alg:BASS}
  \begin{algorithmic}[0]
  \State \textbf{Input:} The observations $\bm x_{1:P}^{(1:N)}$ and the number of iterations for annealing $N_a$.
  \State \textbf{Output:} The estimated precision matrices $\langle K^{(1:N)}\rangle$.
  \For {i = 1 to $N_a$}
    \State - Compute $w = (1 - R^{\{i\}}) N / 2$.
    \State - Generate $\tilde{\bm x}_{1:P}^{(1:N)}$ by bootstrapping from $\bm x_{1:P}^{(1:N)}$ with window width $w$.
    \State - Sample $\tilde\pi_1$, $\tilde A_{00}$, $\tilde A_{11}$, $\tilde\alpha_{jk}$, and $\tilde\beta$ from their variational distributions.
    \State - Update $q(s_{jk}^{(1:N)})$, $q(J_{jk}^{(1:N)})$, and $q(\kappa_j^{(1:N)})$ following the proposed simulated annealing mechanism in Section~\ref{ssec:sim_anneal}.
    \State - Update $q(\pi_1)$, $q(A_{00})$, $q(A_{11})$, $q(\alpha_{jk})$, $q(\beta)$ following their update rules as in Section~\ref{ssec:vi}.
  \EndFor
  \Repeat
    \State - Update all variational distributions following their update rules as in Section~\ref{ssec:vi}.
  \Until convergence.
  \end{algorithmic}
\end{algorithm}

\begin{figure}[t]
\vspace{-2mm}
\centering
\subfloat[No annealing]{
\includegraphics[height = 3.45cm,origin=c]{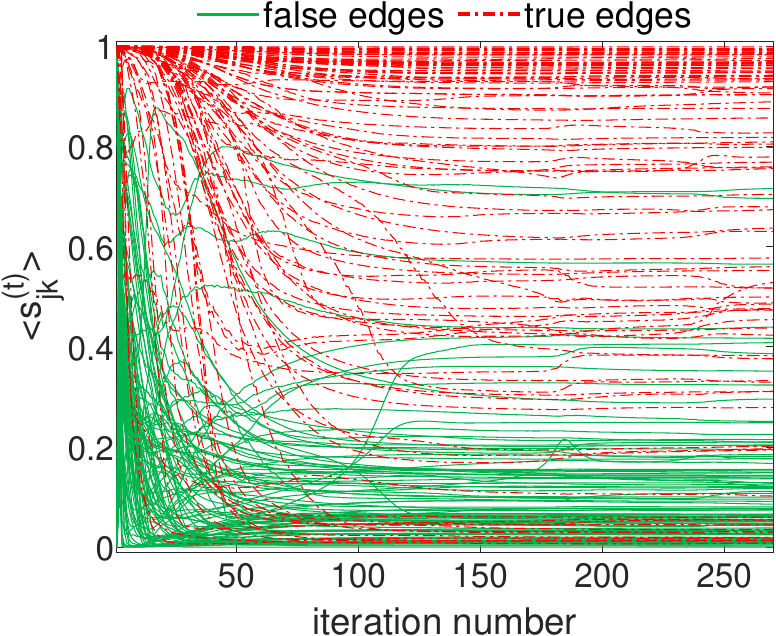}
\label{sfig:noannealing}}
\subfloat[Annealing]{
\includegraphics[height = 3.45cm,origin=c]{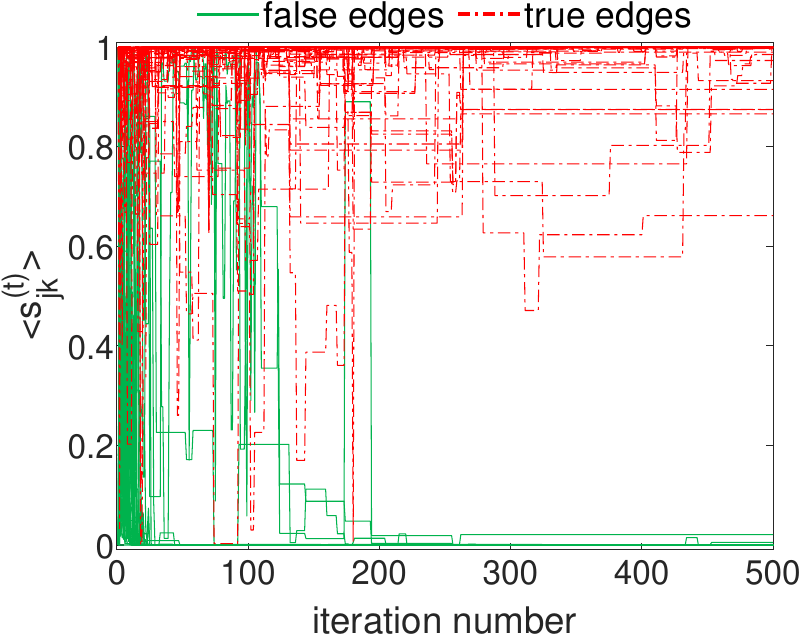}
\label{sfig:annealing}}
\vspace{-1mm}
\caption{Estimation of 200 randomly selected $\langle s_{jk}^{(t)}\rangle$ (i.e., zero pattern of the off-diagonal elements in $K^{(1:T)}$) as a function of iteration number $i$ for BASS when applying to synthetic data with $P = 20$ and $N = 1000$ without and with simulated annealing. Half of the selected off-diagonal elements correspond to the absent edges in the true graph and other half correspond to the true edges.}
\label{fig:Es_vs_iter}
\vspace{-4mm}
\end{figure}

In order to demonstrate the usefulness of the proposed simulated annealing approach, we depict the convergence results of BASS without and with simulated annealing in Fig.~\ref{fig:Es_vs_iter}. Here, we consider a synthetic data set. The true time-varying graph corresponding to this data set is given. We then randomly select 100 true edges (cf. red dashed lines) and 100 absent edges (cf. green solid lines) from the true graph and check how the corresponding $\langle s_{jk}^{(t)}\rangle$ changes as the algorithm proceeds. The initial values for all parameters are the same in both cases. We can tell from Fig.~\ref{sfig:noannealing} that the true and false edges are partially mixed with each other in the original variational inference algorithm without annealing. By contrast, in Fig.~\ref{sfig:annealing}, after using the proposed simulated annealing approach, the red dashed lines clearly stand above the green lines as BASS proceeds. Note that $\langle s_{jk}^{(t)}\rangle$ denotes the existence probability of edge $(j, k)$ at time $t$. The estimated $\langle s_{jk}^{(t)}\rangle$ given by BASS with annealing is close to $1$ for true edges and $0$ for the false ones, successfully separating the true and false edges in an automated manner.

\section{Graphical Models for Stationary Time Series}
\label{ssec:GMTS}

In this section, we discuss how to exploit BASS to learn interactions among $P$ univariate stationary Gaussian processes (i.e., time series) $\bm y_{1:P}^{(1:N)}$.  A graphical model $\mathcal G = (\mathcal V,\mathcal E)$ for $\bm y^{(t)}$ can be constructed by letting an abscent edge $(j,k)\not\in\mathcal E$ denote that the two entire time series $\bm y_j^{(1:N)}$ and $\bm y_k^{(1:N)}$ are conditionally independent given the remaining collection of time series $\bm y_{-jk}^{(1:N)}$~\cite{DYWVLJC:2012}, that is,
\begin{align}
\cov\big(\bm y_j^{(t)},\bm y_k^{(t+\tau)}|\bm y_{-jk}^{(1:N)}\big) = 0, \quad \forall\tau.
\end{align}
In other words, the lagged conditional covariance equals 0 for all time lags $\tau$. On the other hand, the conditional dependence can also be defined in the frequency domain of the time series. Concretely, we first define the spectral density matrix as the Fourier transform of the lagged covariance matrix $\cov(\bm y^{(t)},\bm y^{(t+\tau)})$:
\begin{align}
S^{(\omega)} = \sum_\tau \cov(\bm y^{(t)},\bm y^{(t+\tau)})\exp(-i\omega\tau),
\end{align}
for $\omega\in[0,2\pi]$. Let $K^{(\omega)} = [S^{(\omega)}]^{-1}$, the conditional independence between $\bm y_j^{(1:N)}$ and $\bm y_k^{(1:N)}$ holds if and only if~\cite{JHG:2015,DYWVLJC:2012}:
\begin{align}
K^{(\omega)}_{jk} = 0, \quad \forall \omega.
\end{align}
This suggests that one common zero entry in the inverse spectral density matrices across a certain frequency band is equivalent to the conditional independence between the corresponding two time series in this frequency band. Therefore, for a multivariate time series, we aim to infer the inverse spectral density matrices $K^{(\omega)}$.

Here, we follow the state-of-the-art Whittle approximation framework~\cite{Whittle:1953}: Suppose that $\bm f^{(\omega)}_{1:P}$ is the discrete Fourier transform of $\bm y_{1:P}^{(1:N)}$ at frequency $\omega$:
\begin{align}
f^{(\omega)}_j = \sum_t y_j^t \exp(-i\omega t),  
\end{align}
then $\bm f^{(\omega)}_{1:P}$ are independent complex Gaussian random variables with mean zero and precision matrix given by the inverse spectral density matrix $K^{(\omega)}$ at the same frequency:
\begin{align}
\bm f^{(\omega)}_{1:P} \sim \mathcal N_c\Big(\bm 0,{K^{(\omega)}}^{-1}\Big). \label{eq:freq_dist}
\end{align}
As a result, we can learn $K^{(\omega)}$ that changes smoothly with $\omega$ from $\bm f^{(\omega)}$ using BASS. In this scenario, the covariate is the frequency $\omega$. 
It should be stressed that the complex Gaussian distribution in~\eqref{eq:freq_dist} can be written as:
\begin{align}
p(\bm f^{(\omega)}|K^{(\omega)}) \propto \det(K^{(\omega)})\exp\big(-{\bm f^{(\omega)}}^*K^{(\omega)}\bm f^{(\omega)}\big), \label{eq:freq_exact_likelihood}
\end{align}
where ${\bm f^{(\omega)}}^*$ denotes the complex conjugate transpose of $\bm f^{(\omega)}$. The above density function does not have the operation of square root as in the density function of the Gaussian distribution for real numbers~\eqref{eq:exact_llhd}. The corresponding pseudo-likelihood of $K^{(\omega)}$ can then be expressed as:
\begin{align}
&p(f_j^{(\omega)} | K^{(\omega)}_{jj}, K^{(\omega)}_{j,-j}) \propto K^{(\omega)}_{jj} \exp\Big[- K^{(\omega)}_{jj}f_j^{(\omega)}\overline{f_j^{(\omega)}} \notag \\
&\quad - \overline{f_j^{(\omega)}}K^{(\omega)}_{j,-j}\bm f_{-j}^{(\omega)} - f_j^{(\omega)}\overline{K^{(\omega)}_{j,-j}{\bm f}_{-j}^{(\omega)}} \notag \\
&\quad - {K^{(\omega)}_{jj}}^{-1} K^{(\omega)}_{j,-j}\bm f_{-j}^{(\omega)}\overline{K^{(\omega)}_{j,-j}{\bm f}_{-j}^{(\omega)}} \Big],
\end{align}
where $\overline{f_j^{(\omega)}}$ is the complex conjugate of $f_j^{(\omega)}$. In addition, the prior distribution on $J_{jk}^{(\omega)}$ is also a complex Gaussian distribution. We therefore modify the ELBO $\sL$ and the corresponding update rules accordingly. The detailed update rules are summarized in Table~S.2 in the appendix.

\section{Experimental Results}
\label{sec:results}

In this section, we compare the proposed BASS algorithm with the state-of-the-art methods in the literature. Specifically, for the problem of learning time-varying graphical models, we consider three benchmark methods:
\begin{enumerate}
   \item KERNEL~\cite{ZLW:2010}-\cite{QH:2016}: Kernel-smoothed covariance matrices $S^{(t)}$ are first estimated, and then a graphical model is inferred at each time point by solving the graphical lasso problem:
   \begin{align}
   K^{(t)} = \argmin_{K^{(t)} \succeq 0} \tr(S^{(t)} K^{(t)}) - \log\det K^{(t)} + \lambda_1\|K^{(t)}\|_1, \notag 
   \end{align}
   where $\lambda_1$ controls the sparsity of $K^{(t)}$.
   \item SINGLE~\cite{MHSLAM:2014,GN:2014,YLSWY:2015}: It further controls the smoothness of $K^{(t)}$ over $t$ by imposing penalty on the difference between the precision matrices at every two consecutive time points:
   \begin{align}
   K^{(t)} =&\, \argmin_{K^{(t)}\succeq 0} \sum_{t=1}^N \Big[\tr(S^{(t)} K^{(t)}) - \log\det K^{(t)} + \notag \\
   &\,\lambda_1\|K^{(t)}\|_1\Big] + \lambda_2 \sum_{t=2}^N\|K^{(t)} - K^{(t-1)}\|_1. \label{eq:SINGLE}
   \end{align}
   \item LOGGLE~\cite{YP:2020}: It exploits the local group lasso penalty to promote the sparsity of $K^{(t)}$ and the smoothness across time simultaneously:
   \begin{align}
   K^{(t)} =&\, \argmin_{K^{(t)}\succeq 0}  \sum_t \tr(S^{(t)} K^{(t)}) - \log\det K^{(t)} + \notag \\
   &\,\lambda_1\sum_{j\neq k} \sqrt{\sum_{\tau \in \mathcal D(t, d)} {K_{jk}^{(\tau)}}^2}, \label{eq:LOGGLE}
   \end{align}
   where $\mathcal D(t, d) = \{\tau: |\tau - t|\leq d \}$ denotes neighborhood around $t$ with width $d$. The exact likelihood in the above expression is replaced by the pseudo-likelihood in the implementation to achieve better performance~\cite{YP:2020}.
\end{enumerate}
The time complexity of the above three methods is $\sO(NP^3)$. On the other hand, for the problem of learning graphical models for stationary time series in frequency domain, we compare BASS with the GMS approach proposed in~\cite{JHG:2015}, which can be regarded as the counterpart of KERNEL in the frequency domain.


\subsection{Time-Varying Graphical Models (Time Domain)}

\subsubsection{Synthetic Data}
Given the dimension $P$, the number of time points $N$, and the average number of edges $N_e$, we simulate synthetic Gaussian distributed data from time-varying graphical models as follows. We first generate the off-diagonal elements in the precision matrices $K^{(1:N)}$ as:
\begin{align}
K_{jk}^{(t)} =&\, A_{jk}\sin\bigg(\frac{\pi t}{2N}\bigg) + B_{jk}\cos\bigg(\frac{\pi t}{2N}\bigg) \notag \\
&\,+ C_{jk}\sin\bigg(\pi \Big(\frac{t}{N} + D_{jk}\Big) \bigg),  \label{eq:syn_data}
\end{align}
where for all $j$ and $k$ $A_{jk}$, $B_{jk}$, and $C_{jk}$ are drawn uniformly from $[-1, -0.5]\cup[0.5, 1]$, and $D_{jk}$ follows a uniform distribution in $[-0.25, 0.25]$. We then set a threshold and zero out the off-elements whose magnitude is smaller than the threshold such that the average number of edges in $K^{(1:N)}$ is $N_e$. Next, we compute the diagonal entries as $K^{(t)}_{jj} = \sum_{k\neq j} |K_{jk}^{(t)}| + 0.1$ to guarantee the positive definiteness of $K^{(1:N)}$. Finally, we draw a sample $\bm x^{(t)}$ at each time point $t$ from $\sN(\bm 0, {K^{(t)}}^{-1})$.

\begin{table*}[t!]
\renewcommand{\arraystretch}{1.4}
  \vspace{-1mm}
  \centering
  \caption{Graph recovery results from different methods for synthetic data with different dimensions ($P = 20, 100, 500$, $N = 1000$, $N_e = P$). The standard deviations are shown in the brackets.}
  \vspace{-1mm}
  \resizebox{\linewidth}{!}{
    \begin{tabular}{c|cccc|cccc|cccc}
    \hline
    \hline
    \multirow{2}{*}{Methods}               &\multicolumn{4}{c|}{$P = 20$}         &\multicolumn{4}{c|}{$P = 100$}        &\multicolumn{4}{c}{$P = 500$}\\
    \cline{2-13}
                                        &Precision      &Recall         &$F_1$-score    &Time(s)        &Precision  &Recall &$F_1$-score    &Time(s)    &Precision  &Recall &$F_1$-score    &Time(s)          \\
    \hline
    BASS                               &0.89 (5.06e-2) &0.86 (2.31e-2) &0.88 (3.12e-2) &1.76e2 (2.39)      &0.82 (4.06e-2) &0.77 (3.89e-2) &0.80 (2.68e-2) &4.73e3 (2.24e1)    &0.74 (1.02e-2) &0.69 (1.11e-2) &0.72 (1.01e-2) &1.33e5 (3.59e3)\\
    KERNEL (orcale)                     &0.83 (1.75e-2) &0.88 (4.37e-2) &0.85 (1.88e-2) &4.81e2 (7.24)      &0.71 (1.94e-2) &0.79 (2.02e-2) &0.75 (7.21e-3) &4.50e4 (8.20e2)    &0.66 (2.45e-3) &0.49 (6.39e-3) &0.58 (4.50e-3) &5.80e6 (1.38e5)\\
    KERNEL (CV)                         &0.63 (1.18e-2) &0.91 (1.91e-2) &0.74 (1.32e-2) &6.51e2 (2.58e1)    &0.71 (1.70e-2) &0.79 (2.23e-2) &0.75 (9.03e-3) &6.61e4 (1.07e3)    &0.86 (1.72e-1) &0.29 (1.94e-1) &0.38 (1.75e-1) &8.59e6 (1.47e5)\\
    SINGLE (oracle)                     &0.86 (2.58e-2) &0.90 (3.62e-2) &0.88 (2.56e-2) &1.04e4 (3.19e2)    &0.84 (1.26e-2) &0.74 (2.10e-2) &0.79 (1.01e-2) &1.58e6 (2.99e5)    &&&&\\
    SINGLE (CV)                         &0.86 (2.54e-2) &0.89 (3.84e-2) &0.88 (2.79e-2) &5.09e4 (2.17e3)    &0.68 (1.60e-2) &0.51 (1.41e-2) &0.58 (7.53e-3) &6.94e6 (1.04e5)    &&&&\\
    LOGGLE (oracle)                     &0.89 (1.17e-2) &0.86 (3.98e-2) &0.88 (1.89e-2) &1.75e4 (1.03e3)    &0.81 (1.51e-2) &0.78 (2.10e-2) &0.80 (9.59e-3) &1.51e6 (3.60e5)    &&&&\\
    LOGGLE (CV)                         &0.71 (2.18e-2) &0.93 (3.36e-2) &0.81 (1.78e-2) &5.08e4 (1.75e3)    &0.47 (1.15e-2) &0.80 (2.31e-2) &0.59 (1.41e-3) &5.07e6 (4.59e5)    &&&&\\
    \hline
    \hline
    \end{tabular}    }%
    \label{tab:syn_dim}
    \vspace{-1mm}
\renewcommand{\arraystretch}{1}
\end{table*}

\begin{table*}[t!]
\renewcommand{\arraystretch}{1.4}
  \centering
  \caption{Graph recovery results from different methods for synthetic data with different sample size ($P = 20$, $N = 500, 1000, 2000$, $N_e = P$). The standard deviations are shown in the brackets.}
  \vspace{-1mm}
  \resizebox{\linewidth}{!}{
    \begin{tabular}{c|cccc|cccc|cccc}
    \hline
    \hline
    \multirow{2}{*}{Methods}               &\multicolumn{4}{c|}{$N = 500$}         &\multicolumn{4}{c|}{$N = 1000$}        &\multicolumn{4}{c}{$N = 2000$}\\
    \cline{2-13}
                                        &Precision  &Recall &$F_1$-score    &Time(s)    &Precision  &Recall &$F_1$-score    &Time(s)    &Precision  &Recall &$F_1$-score    &Time(s)          \\
    \hline
    BASS                               &0.83 (7.02e-2) &0.71 (2.52e-2) &0.77 (2.86e-2) &9.75e1 (1.85)      &0.89 (5.06e-2) &0.86 (2.31e-2) &0.88 (3.12e-2) &1.76e2 (2.39)      &0.94 (4.30e-2) &0.90 (4.98e-2) &0.92 (2.20e-2) &3.44e2 (6.24)\\
    KERNEL (orcale)                     &0.62 (4.25e-2) &0.84 (3.21e-2) &0.71 (1.69e-2) &2.33e2 (5.56)      &0.83 (1.75e-2) &0.88 (4.37e-2) &0.85 (1.88e-2) &4.81e2 (7.24)      &0.88 (2.02e-2) &0.91 (2.32e-2) &0.89 (1.34e-2) &1.11e3 (4.48e1)\\
    KERNEL (CV)                         &0.53 (1.78e-2) &0.79 (1.21e-1) &0.61 (1.50e-1) &3.12e2 (5.05)      &0.63 (1.18e-2) &0.91 (1.91e-2) &0.74 (1.32e-2) &6.51e2 (2.58e1)    &0.84 (1.82e-2) &0.93 (1.48e-2) &0.88 (1.15e-2) &1.37e3 (3.10e1)\\
    SINGLE (oracle)                     &0.69 (5.37e-2) &0.82 (4.23e-2) &0.75 (1.67e-2) &4.17e3 (5.81e1)    &0.86 (2.58e-2) &0.90 (3.62e-2) &0.88 (2.56e-2) &1.04e4 (3.19e2)    &0.92 (1.23e-2) &0.92 (2.31e-2) &0.92 (9.80e-3) &2.85e4 (4.99e2)\\
    SINGLE (CV)                         &0.70 (4.81e-2) &0.77 (6.95e-2) &0.73 (2.55e-2) &2.39e4 (1.49e3)    &0.86 (2.54e-2) &0.89 (3.84e-2) &0.88 (2.79e-2) &5.09e4 (2.17e3)    &0.92 (1.23e-2) &0.92 (2.31e-2) &0.92 (9.80e-3) &1.44e5 (3.81e3)\\
    LOGGLE (oracle)                     &0.71 (3.24e-2) &0.83 (3.80e-2) &0.77 (4.77e-3) &8.28e3 (3.20e2)    &0.89 (1.17e-2) &0.86 (3.98e-2) &0.88 (1.89e-2) &1.75e4 (1.03e3)    &0.93 (2.50e-2) &0.92 (3.02e-2) &0.92 (1.93e-2) &3.82e4 (3.25e3)\\
    LOGGLE (CV)                         &0.59 (8.98e-2) &0.79 (5.29e-2) &0.67 (7.40e-2) &2.01e4 (7.29e2)    &0.71 (2.18e-2) &0.93 (3.36e-2) &0.81 (1.78e-2) &5.08e4 (1.75e3)    &0.80 (9.23e-2) &0.94 (3.49e-2) &0.86 (4.60e-2) &1.22e5 (1.05e4)\\
    \hline
    \hline
    \end{tabular}    }%
    \label{tab:syn_sample}
    \vspace{-1mm}
\renewcommand{\arraystretch}{1}
\end{table*}

\begin{table*}[t!]
\renewcommand{\arraystretch}{1.4}
  \centering
  \caption{Graph recovery results from different methods for synthetic data with different density ($P = 20$, $N = 1000$, $N_e = 10, 20, 40$). The standard deviations are shown in the brackets.}
  \vspace{-1mm}
  \resizebox{\linewidth}{!}{
    \begin{tabular}{c|cccc|cccc|cccc}
    \hline
    \hline
    \multirow{2}{*}{Methods}               &\multicolumn{4}{c|}{$N_e = 10$}         &\multicolumn{4}{c|}{$N_e = 20$}        &\multicolumn{4}{c}{$N_e = 40$}\\
    \cline{2-13}
                                        &Precision  &Recall &$F_1$-score    &Time(s)    &Precision  &Recall &$F_1$-score    &Time(s)    &Precision  &Recall &$F_1$-score    &Time(s)          \\
    \hline
    BASS                               &0.90 (8.89e-2) &0.94 (2.16e-2) &0.92 (4.63e-2) &1.68e2 (3.18)      &0.89 (5.06e-2) &0.86 (2.31e-2) &0.88 (3.12e-2) &1.76e2 (2.39)      &0.85 (4.57e-2) &0.66 (5.84e-2) &0.74 (2.36e-2) &1.63e2 (1.80)\\
    KERNEL (orcale)                     &0.86 (1.33e-1) &0.88 (1.29e-1) &0.86 (2.52e-2) &4.51e2 (7.54)      &0.83 (1.75e-2) &0.88 (4.37e-2) &0.85 (1.88e-2) &4.81e2 (7.24)      &0.90 (2.93e-2) &0.59 (6.16e-2) &0.71 (4.27e-2) &4.39e2 (7.26)\\
    KERNEL (CV)                         &0.77 (2.90e-1) &0.80 (1.72e-1) &0.73 (1.16e-1) &6.02e2 (2.92e1)    &0.63 (1.18e-2) &0.91 (1.91e-2) &0.74 (1.32e-2) &6.51e2 (2.58e1)    &0.58 (2.00e-1) &0.82 (1.36e-1) &0.65 (4.88e-2) &5.57e2 (2.23e1)\\
    SINGLE (oracle)                     &0.87 (2.58e-2) &0.90 (3.62e-2) &0.89 (2.56e-2) &1.04e4 (3.19e2)    &0.86 (2.58e-2) &0.90 (3.62e-2) &0.88 (2.56e-2) &1.04e4 (3.19e2)    &0.87 (3.55e-2) &0.61 (5.38e-2) &0.72 (3.82e-2) &9.83e3 (1.90e2)\\
    SINGLE (CV)                         &0.67 (7.38e-2) &0.93 (2.42e-2) &0.78 (4.51e-2) &5.43e4 (5.52e3)    &0.86 (2.54e-2) &0.89 (3.84e-2) &0.88 (2.79e-2) &5.09e4 (2.17e3)    &0.65 (2.26e-1) &0.75 (1.20e-1) &0.66 (7.79e-2) &4.79e4 (1.60e3)\\
    LOGGLE (oracle)                     &0.86 (7.43e-2) &0.89 (8.43e-2) &0.88 (3.11e-2) &1.77e4 (1.36e3)    &0.89 (1.17e-2) &0.86 (3.98e-2) &0.88 (1.89e-2) &1.75e4 (1.03e3)    &0.86 (8.88e-3) &6.53 (5.24e-2) &0.74 (3.60e-2) &1.62e4 (9.66e2)\\
    LOGGLE (CV)                         &0.55 (7.16e-2) &0.98 (2.18e-2) &0.70 (5.83e-2) &5.16e4 (2.52e3)    &0.71 (2.18e-2) &0.93 (3.36e-2) &0.81 (1.78e-2) &5.08e4 (1.75e3)    &0.76 (1.58e-1) &0.70 (7.45e-2) &0.72 (6.97e-2) &5.01e4 (1.89e3)\\
    \hline
    \hline
    \end{tabular}    }%
    \label{tab:syn_density}
    \vspace{-4mm}
\renewcommand{\arraystretch}{1}
\end{table*}

\begin{figure}[t]
\vspace{-1mm}
\centering
\includegraphics[width = 0.8\columnwidth]{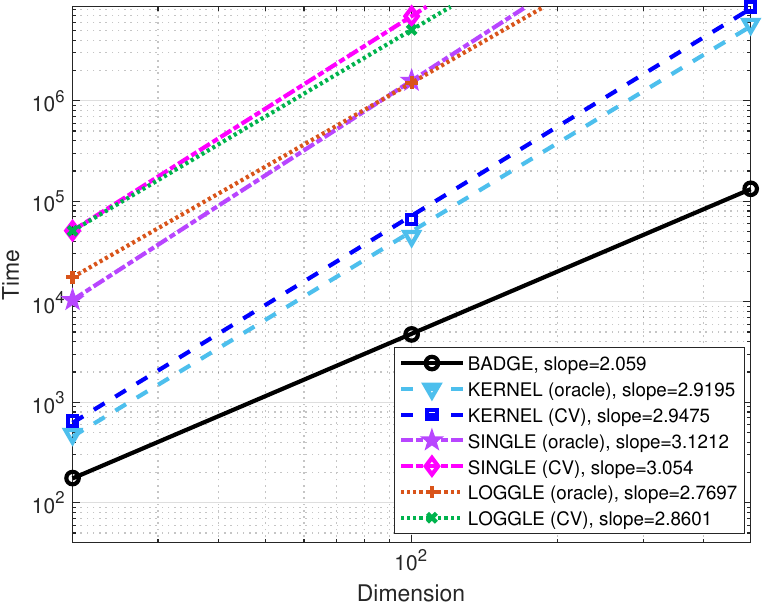}
\vspace{-1mm}
\caption{Computational time as a function of dimension $P$. We fit a straight line to the logarithm of average run time in Table~\ref{tab:syn_dim} vs. the logarithm of $P$, and compute the slope. The slope provides an empirical measure of the time complexity.}
\label{fig:t_vs_p}
\vspace{-5mm}
\end{figure}

We compare all methods in terms of precision, recall, $F_1$-score, and computational time. Precision is defined as the proportion of correctly estimated edges to all the edges in the estimated graph; recall is defined as the proportion of correctly estimated edges to all the edges in the true graph; $F_1$-score is defined as 2$\cdot$precision$\cdot$recall/(precision+recall), which is a weighted average of the precision and recall. For the benchmark methods, since the ground truth is given, we first select the tuning parameters that maximize the $F_1$-score and refer to the results as oracle results (a.k.a. optimal results). We also show the results when the tuning parameters are selected via cross validation (CV), which is commonly used in existing works~\cite{KSAX:2010,MHSLAM:2014,GN:2017}-\cite{YP:2020}. More specifically, for all three frequentist methods, we choose the kernel bandwidth $h$ from 5 candidates $\{\exp(-4)N, \exp(-3)N, \cdots, N\}$ and $\lambda_1$ from 5 candidates $\{\exp(-4), \exp(-3), \cdots, 1\}$. We further select $\lambda_2$ from $\{\exp(-4), \exp(-2), \cdots, \exp(4)\}$ for SINGLE and $d$ from $\{\exp(-4)N, \exp(-3)N, \cdots, N\}$ for LOGGLE. We show the results for graphs with different dimensions $P$, different sample sizes $N$, and different graph densities (characterized by the average number of edges $N_e$) respectively in Table~\ref{tab:syn_dim}, Table~\ref{tab:syn_sample}, and Table~\ref{tab:syn_density}. The results are averaged over 5 trials and the standard deviation is presented in the brackets. All methods are implemented using R and Rcpp.

We observe that BASS typically obtains the highest $F_1$-score with the shortest period of computational time, regardless of the dimension, the sample size, and the graph density. In other words, BASS can well recover the time-varying graph structure in an automated fashion. On the other hand, the oracle results of LOGGLE and SINGLE are comparable to that of BASS. However, in practice, we have no information on the true graphs and therefore cannot choose the tuning parameters that maximize the $F_1$-score (i.e., minimize the difference between the true and estimated graphs). As mentioned before, one practical approach to choosing the tuning parameters is CV. Unfortunately, as can be observed from Table~\ref{tab:syn_dim}-\ref{tab:syn_density}, the CV results are typically worse than the oracle results, indicating that CV cannot always find the optimal tuning parameters. Since we only consider 5 candidates for each tuning parameter and so the distance between every two combinations of tuning parameters is relatively large, the wrong choice of the optimal parameters may lead to a large gap between the CV and the oracle results. Additionally, we can see that LOGGLE and SINGLE outperform KERNEL in terms of $F_1$-score after capturing the temporal dependence across $K^{(t)}$. Nevertheless, they become prohibitively slow when $P \geq 100$, severely hindering their application to large-scale problems. By contrast, BASS improves the estimation accuracy while being more efficient. Indeed, as shown in Fig.~\ref{fig:t_vs_p}, the computational time of BASS approximately increases \emph{quadratically} with $P$, whereas that of the other methods is approximately a \emph{cubic} function of $P$. These observations are in agreement with the theoretical time complexity. On the other hand, the computational time of all methods increases linearly with $N$ (see Table~\ref{tab:syn_sample}), while the graph density does not have much effect on the computational time (see Table~\ref{tab:syn_density}). Finally, we find that the performance of BASS and all oracle results in terms of $F_1$-score improves as the dimension $P$ decreases, the graph density $N_e$ decreases or the sample size $N$ increases. In general, more information is required to reliably estimate the true graphs when $P$ or $N_e$ increases, but the sample size $N$ remains invariant in these cases, thus degrading the performance of these methods. On the other hand, according to Eq.~\eqref{eq:syn_data}, the precision matrix $K^{(t)}$ changes more slowly with $t$ as the sample size $N$ increases, and so it becomes easier to estimate the time-varying graphs. This explains the improvement of the $F_1$-score with increasing $N$.

\subsubsection{Stock Return Data of 78 Banks}

\begin{figure}[t]
\vspace{-1mm}
\centering
\subfloat[]{
\includegraphics[width = \columnwidth, origin=c]{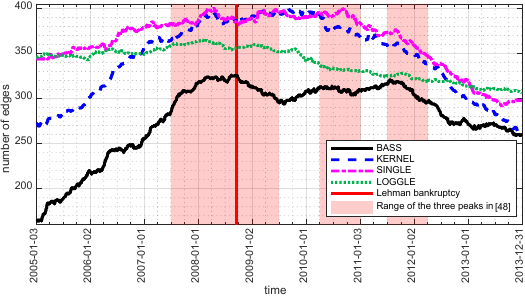}
\label{sfig:bank_allmethods}}
\\
\subfloat[]{
\includegraphics[width = \columnwidth, origin=c]{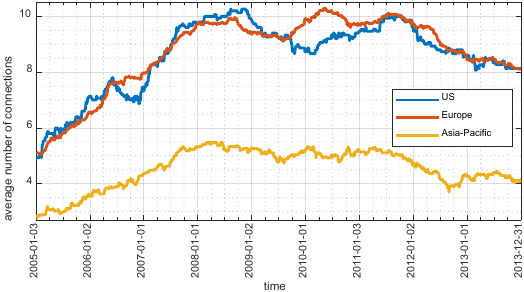}
\label{sfig:bank_BADGE}}
\vspace{-1mm}
\caption{Results of the stock return data of 78 banks: (a) the number of edges as a function of time resulting from different methods; (b) the average number of connections for banks in US, Europe, and Asia-Pacific given by BASS.}
\label{fig:bank78}
\vspace{-3mm}
\end{figure}

\begin{figure*}[t]
\centering
\begin{minipage}[b]{0.92\textwidth}
\subfloat[2005-01-03]{
\includegraphics[width = 0.19\linewidth]{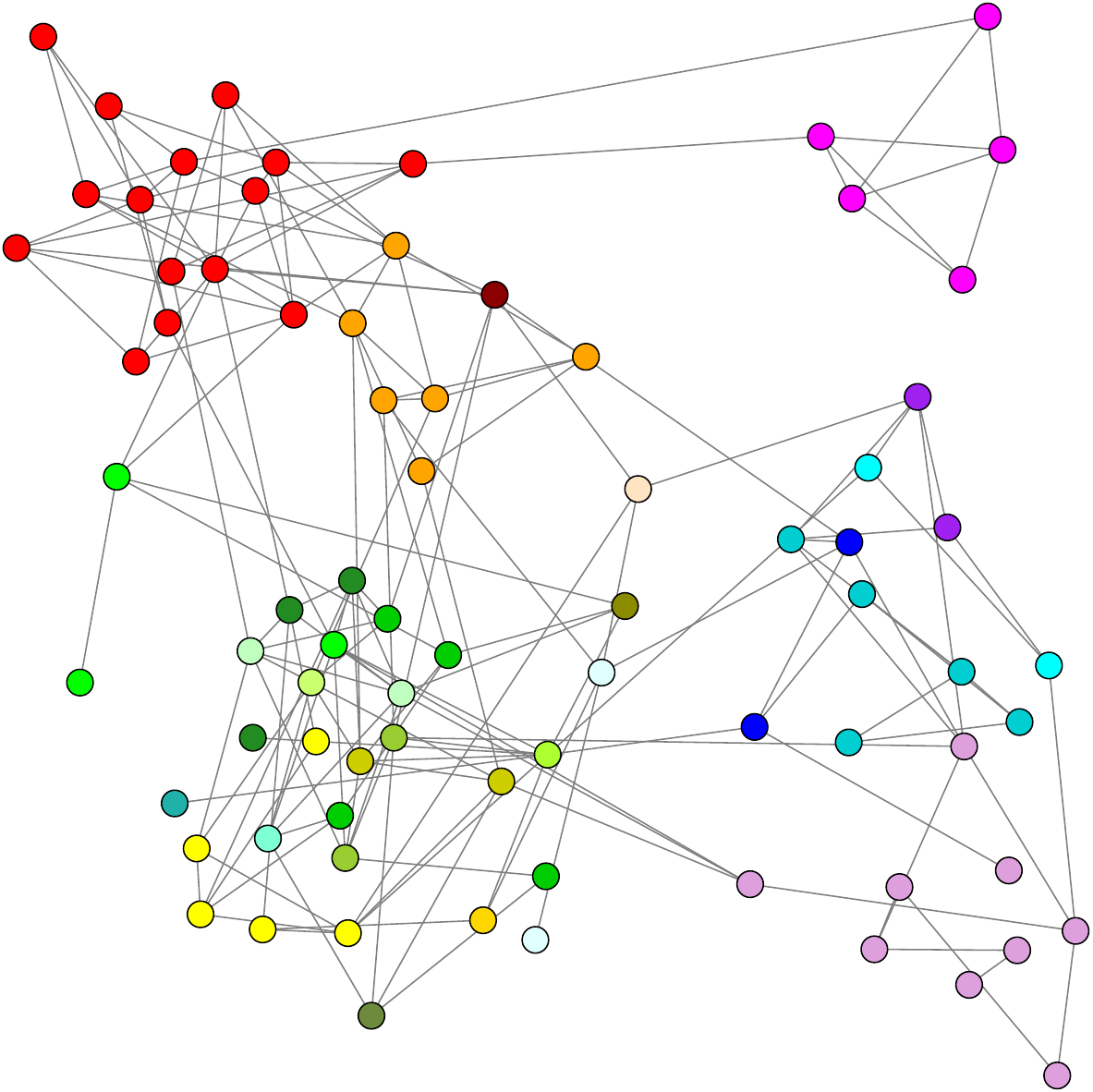}
\label{sfig:05_01}}
\subfloat[2006-01-02]{
\includegraphics[width = 0.19\linewidth]{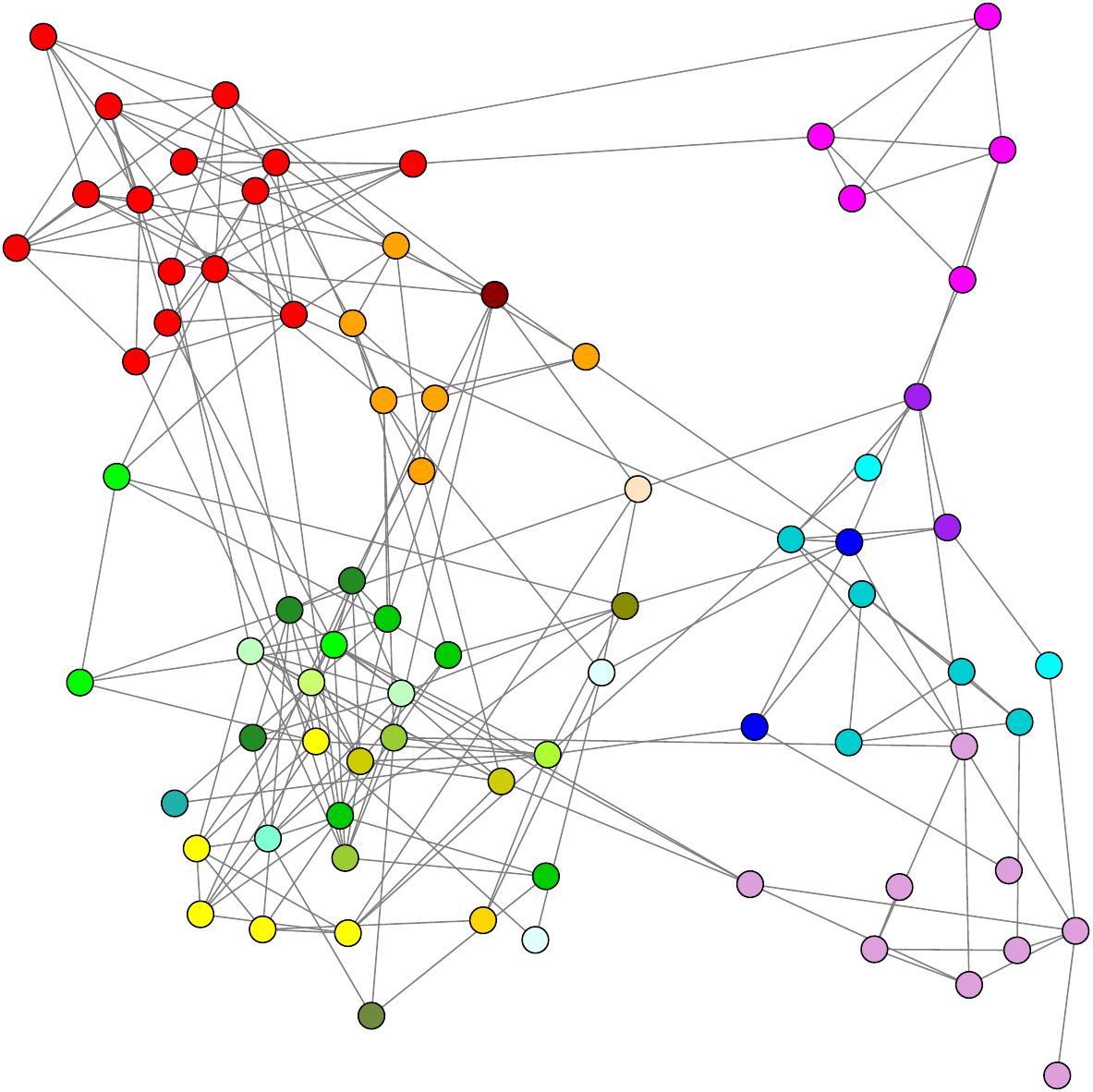}
\label{sfig:06_01}}
\subfloat[2007-01-01]{
\includegraphics[width = 0.19\linewidth]{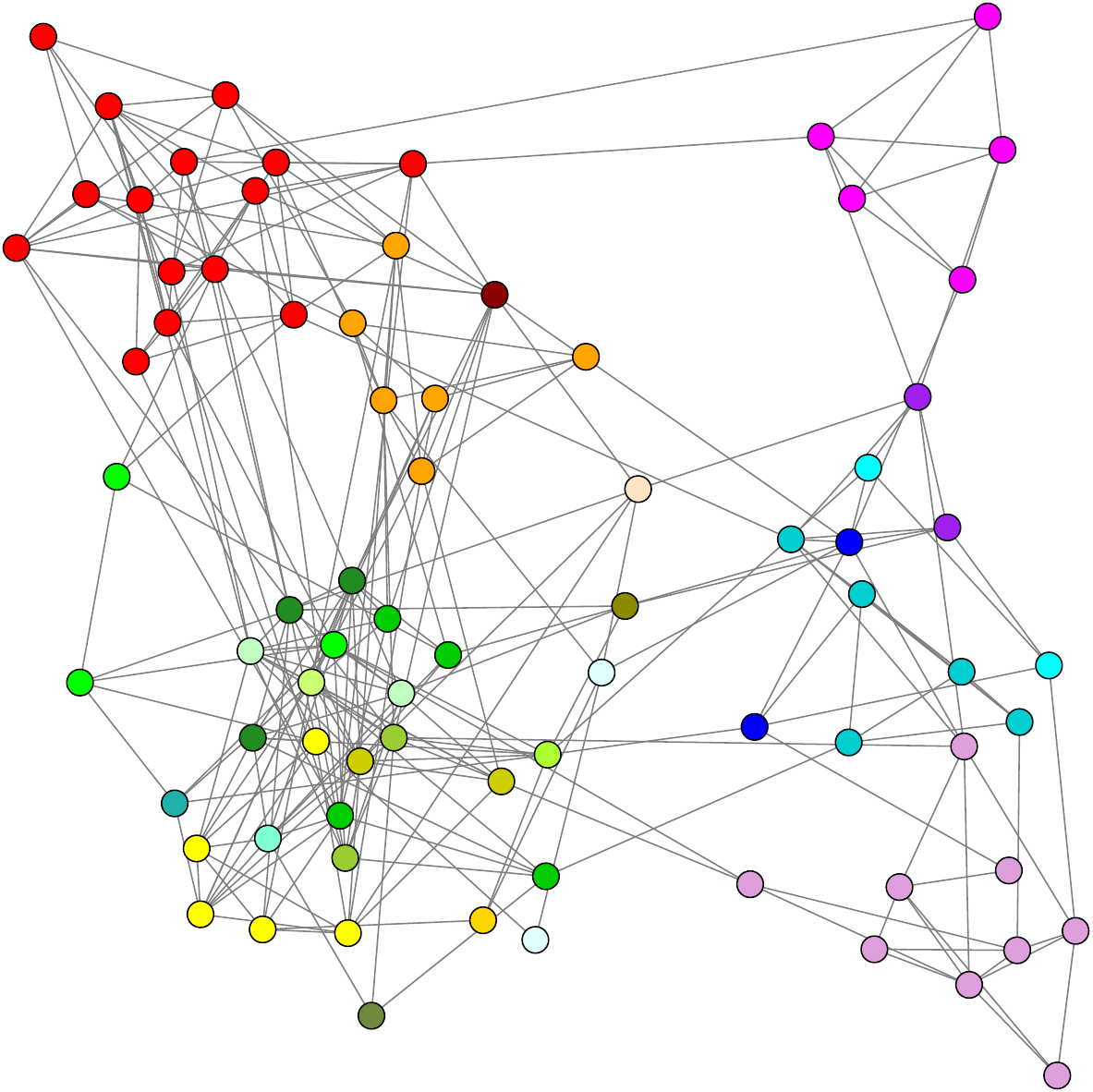}
\label{sfig:07_01}}
\subfloat[2008-01-01]{
\includegraphics[width = 0.19\linewidth]{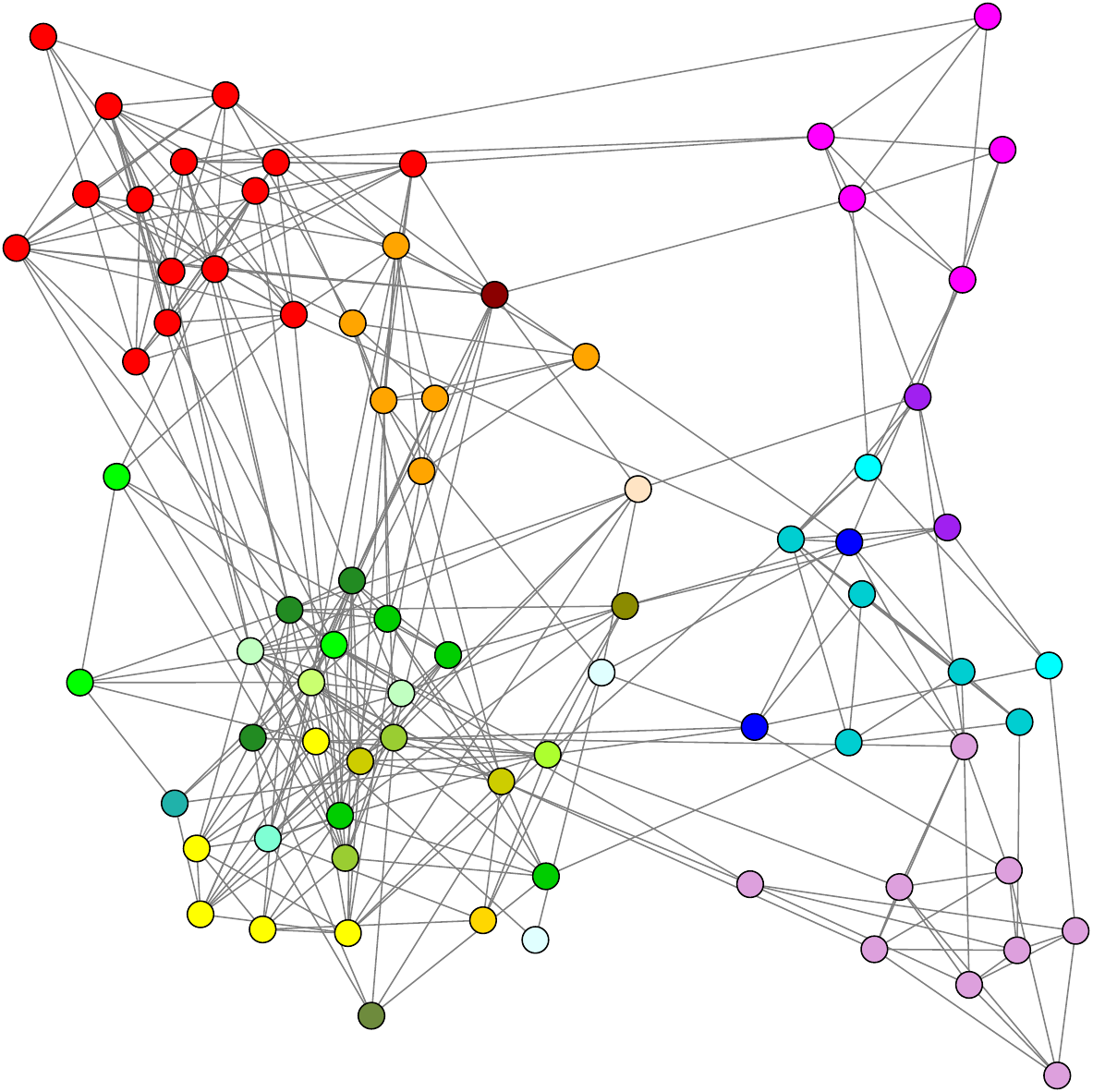}
\label{sfig:08_01}}
\subfloat[2009-01-02]{
\includegraphics[width = 0.19\linewidth]{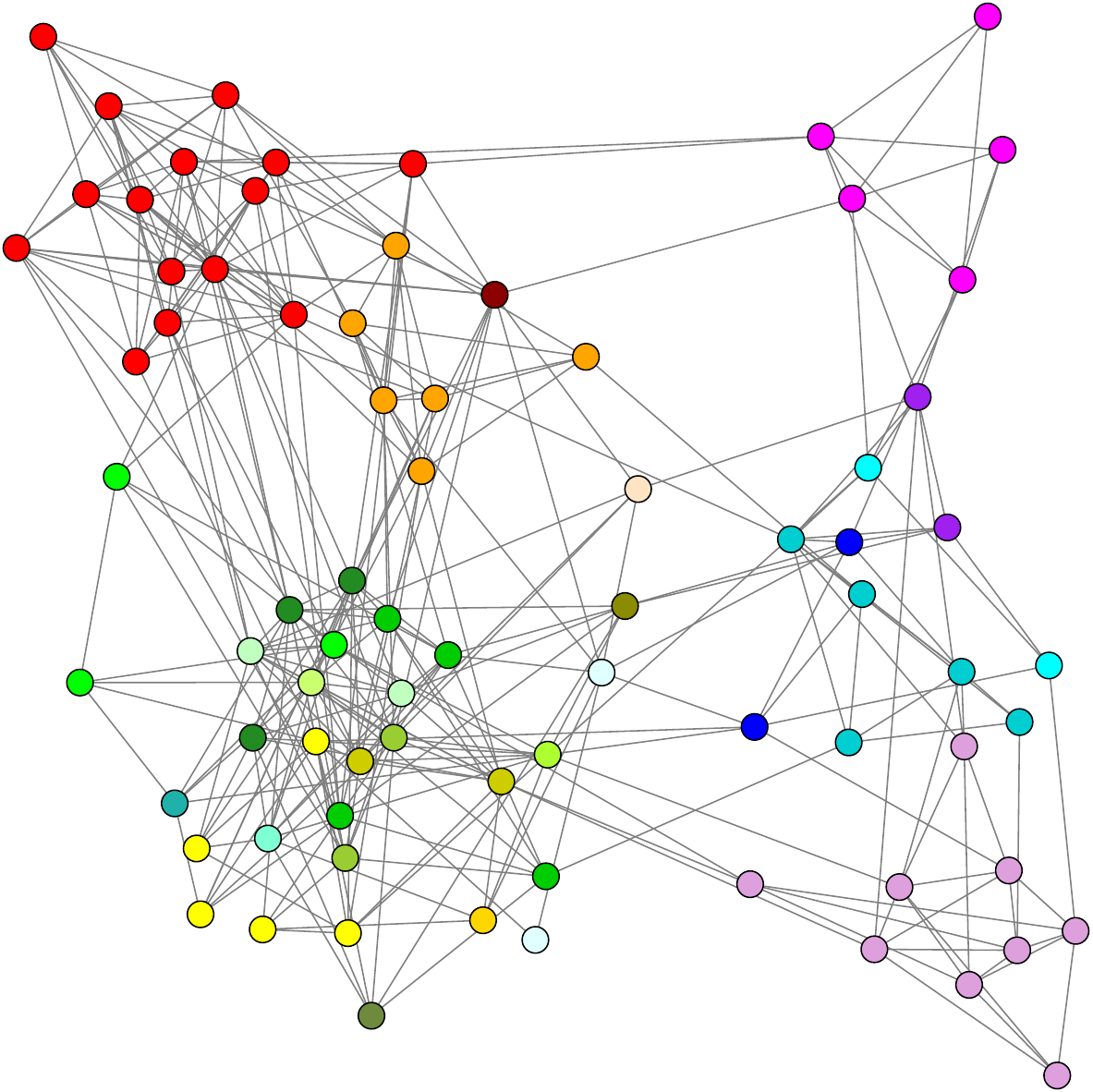}
\label{sfig:09_01}}\\
\subfloat[2010-01-01]{
\includegraphics[width = 0.19\linewidth]{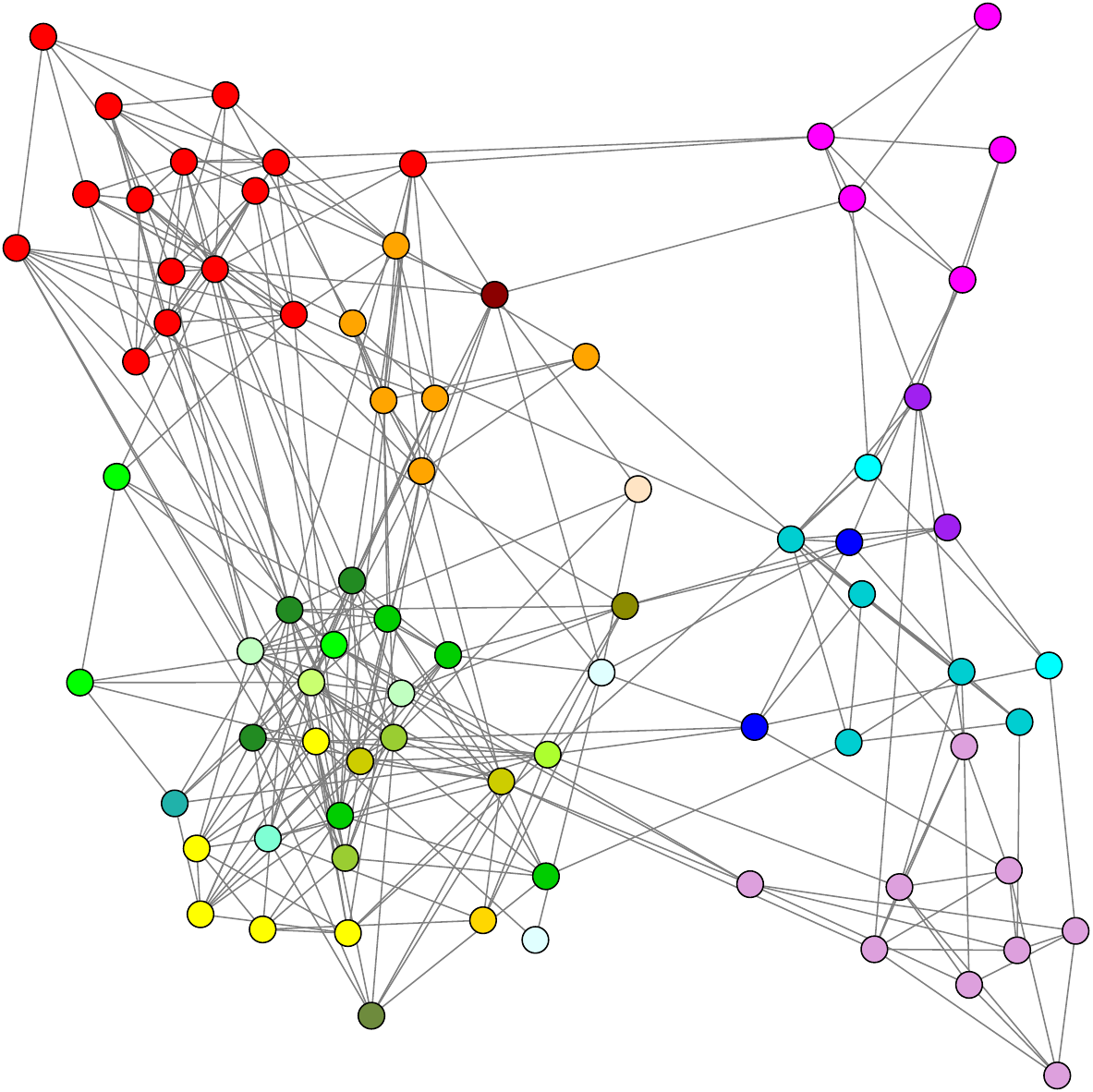}
\label{sfig:10_01}}
\subfloat[2011-01-03]{
\includegraphics[width = 0.19\linewidth]{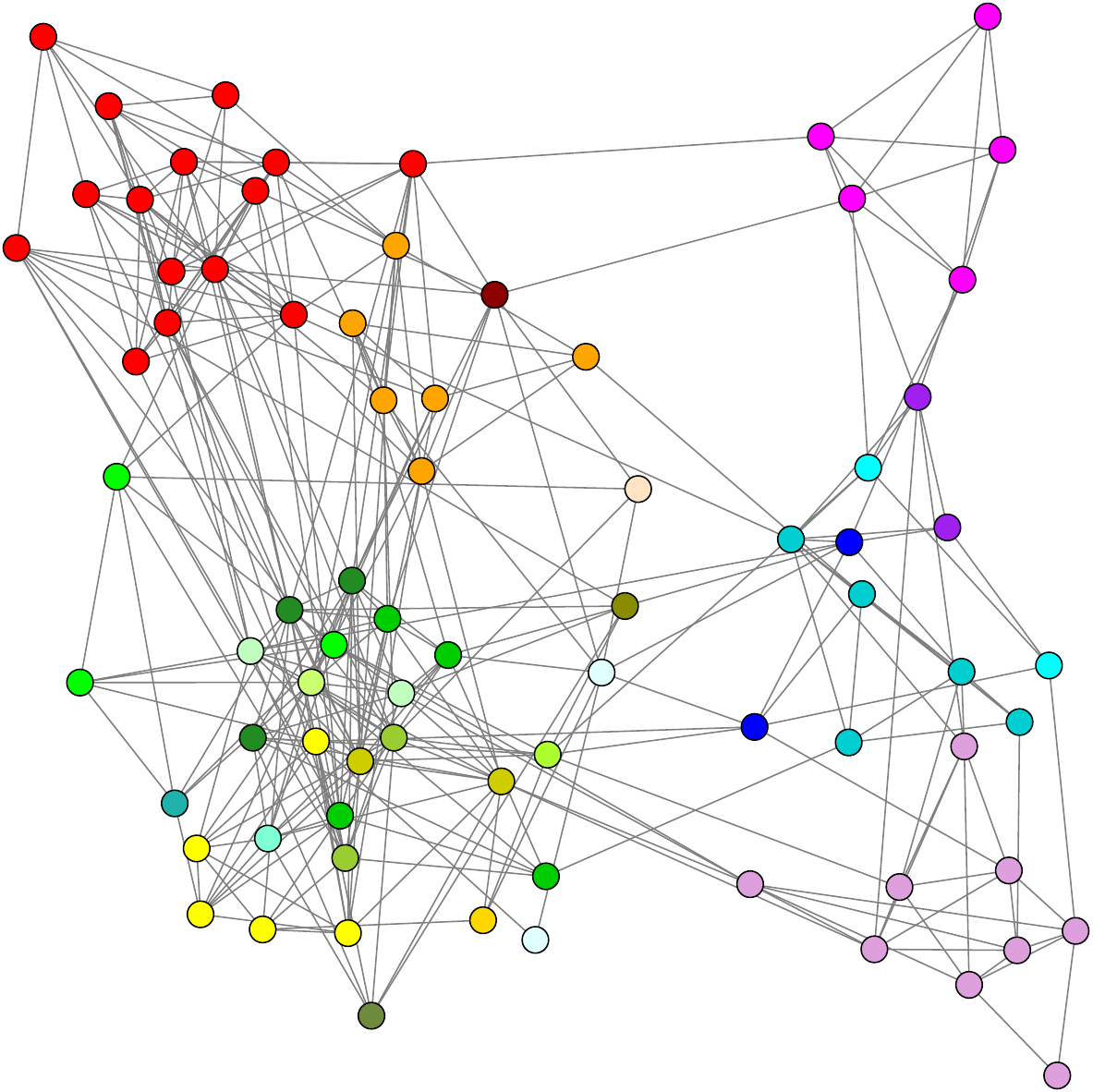}
\label{sfig:11_01}}
\subfloat[2012-01-02]{
\includegraphics[width = 0.19\linewidth]{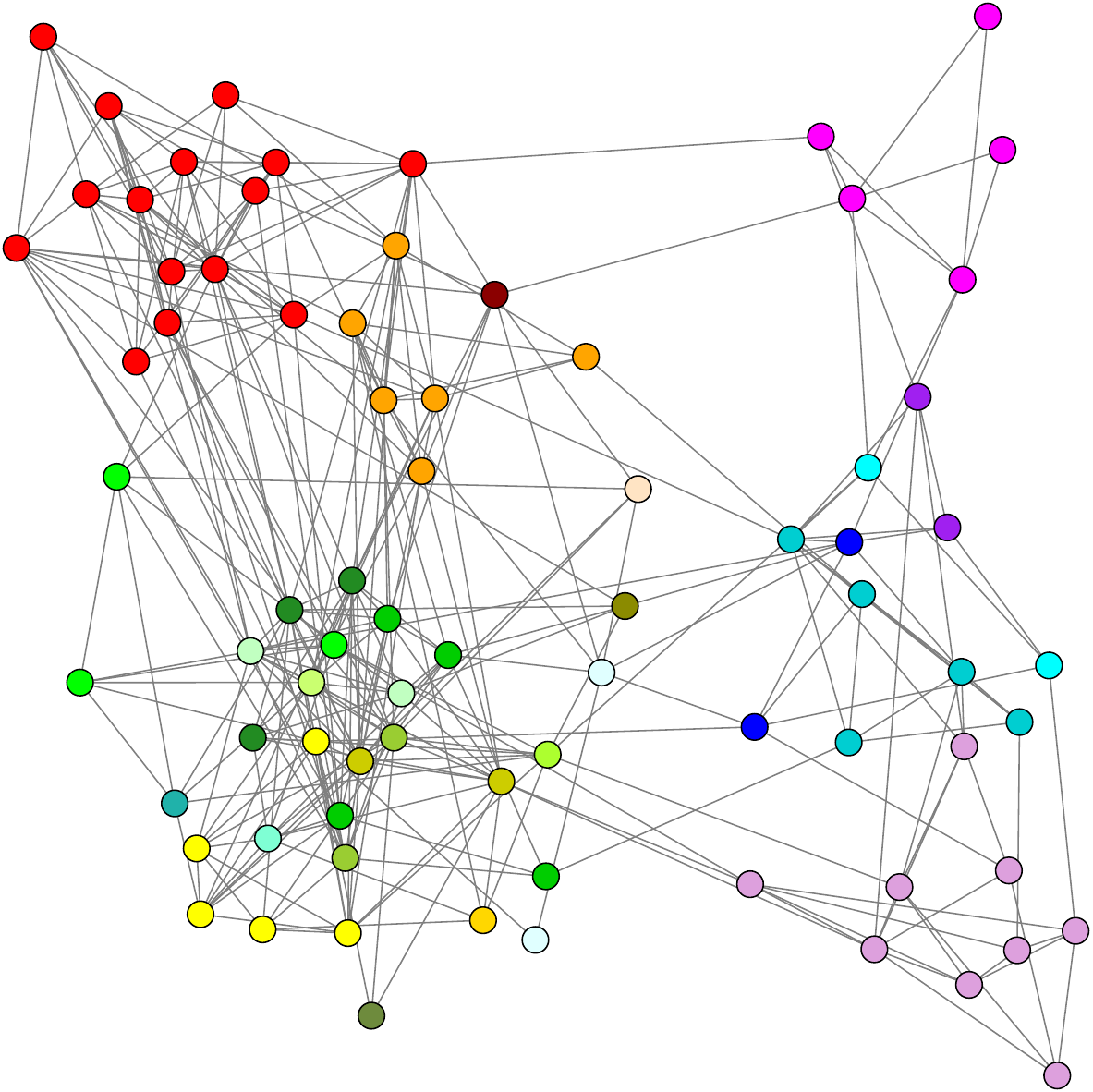}
\label{sfig:12_01}}
\subfloat[2013-01-01]{
\includegraphics[width = 0.19\linewidth]{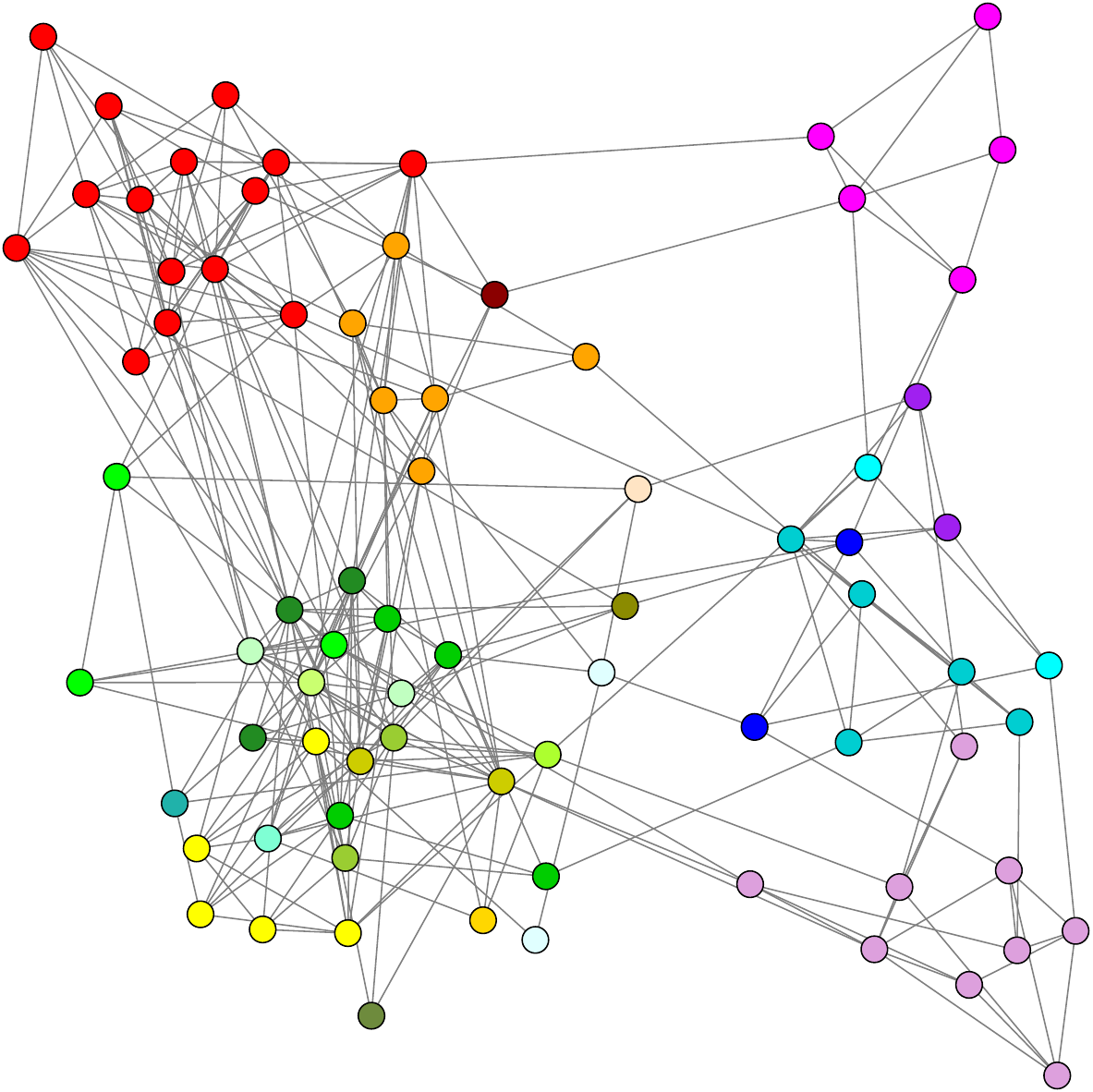}
\label{sfig:13_01}}
\subfloat[2013-12-31]{
\includegraphics[width = 0.19\linewidth]{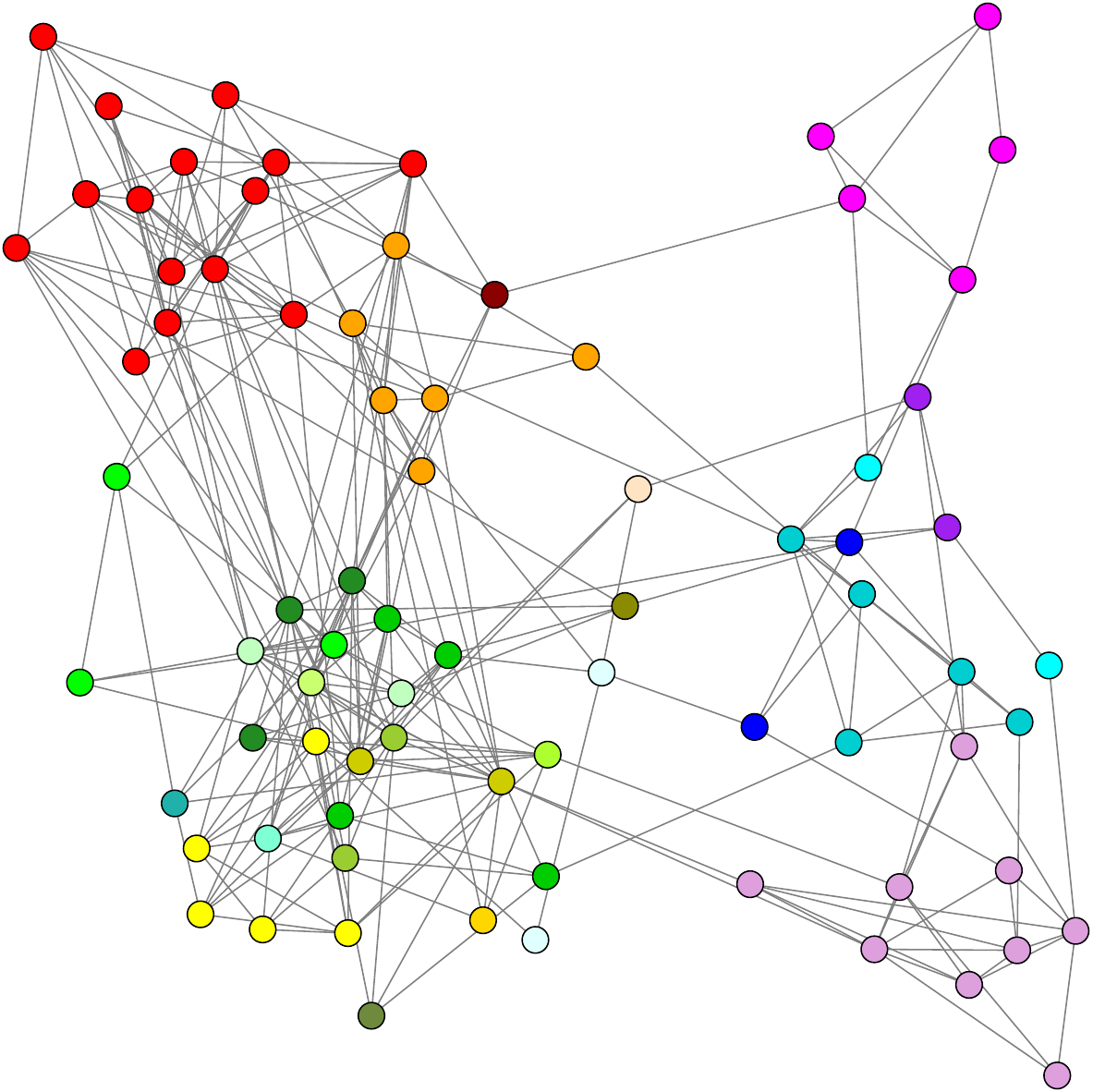}
\label{sfig:13_12}}
\end{minipage}
\hfill
\subfloat{
\includegraphics[width = 0.06\textwidth]{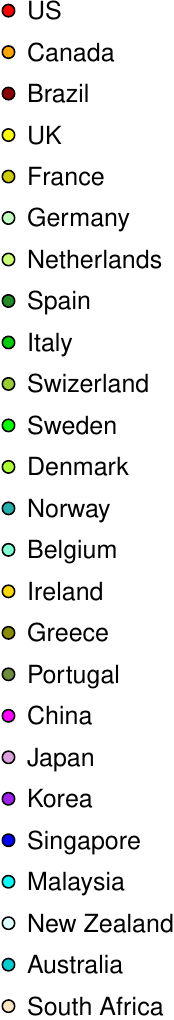}}
\vspace{-1mm}
\caption{Financial networks resulting from BASS from 2005 to 2013. The networks are time-varying. The banks are clustered according to their regions in an automatic fashion.}
\label{fig:finance_net}
\vspace{-5mm}
\end{figure*}

We infer time-varying financial networks from the daily stock return data during the 2008 Great Recession (from 2005 to 2013) of 78 banks in 25 countries around the world. Learning financial networks from data greatly helps to model the system risk in a financial market~\cite{BGLP:2012}; such networks can be utilized to find the interactions between financial institutions for risk contagion that impair the system stability as well as to determine which institutions are more contagious or subject to contagion. The 78 banks are those in the world's top 150 (by assets) whose data are publicly available at Yahoo Finance (https://finance.yahoo.com/) from 2005 to 2013. In total, we have 2348 observations for each of the 78 variables. $6.49\%$ of the data are missing at random. The proposed Bayesian method, BASS, can deal with the missing data, by inferring their variational distributions along with the distribution of the precision matrices, whereas the frequentist methods cannot. As such, we use the data imputed by BASS as the input to KERNEL, SINGLE, and LOGGLE. The tuning parameters in these methods are selected by CV. Before applying all methods, we normalize the data to have unit variance. The time-varying variance for data normalization is inferred via kernel smoothing, and the kernel bandwidth is chosen via CV. 

The estimated number of edges in the time-varying networks as a function of time resulting from different methods is summarized in Fig.~\ref{sfig:bank_allmethods}. The critical point during the 2008 global financial crisis is the Lehman bankruptcy on Sept. 15, 2008 when the US government started to implement a huge stimulus package so as to alleviate the influence of the financial crisis~\cite{Williams:2010}. As demonstrated in Fig.~\ref{sfig:bank_allmethods}, the number of edges resulting from all methods tends to increase before the Lehman bankruptcy and then decreases gradually after it. In the viewpoint of system risk, all banks in the financial system would progressively suffer from the financial crisis due to the risk contagion, leading to similar stock price movement and more connections in the networks during the financial crisis~\cite{GY:2016}. Apart from the major peak in 2008, BASS further yields another two peaks in 2010 and 2011. These two peaks correspond to the Greek debt crisis of 2009-2010 and the subsequent sovereign debt crisis of 2011 in Europe. Interestingly, Demirer~\eall~\cite{DDLY:2018} analyzed the volatilities of 96 banks in the world's top 150 and further constructed time-varying networks by inferring sparse vector autoregressive approximation models in sliding windows of the volatility data. The three peaks were also observed in their experiment; the range of the three peaks in~\cite{DDLY:2018} is shown in Fig.~\ref{sfig:bank_allmethods}. Here we learn networks from the stock return data by means of BASS, instead of extracting them from volatilities, yet we obtain similar results. As opposed to BASS, the three peaks are not very obvious for KERNEL, SINGLE, and LOGGLE. The results of these methods may become better by selecting the tuning parameters from a larger set of candidates, at the expense of increased computational time. Unfortunately, the computational time for KERNEL, SINGLE, and LOGGLE is already 1.48e5, 4.90e7, and 8.06e6 seconds respectively, and testing more candidates of the tuning parameters will make these methods distressingly slow. On the other hand, BASS only takes 3.08e4 seconds to converge. In summary, BASS can better capture the changes in the financial networks in less amount of run time.

Next, let us delve into the results given by BASS. We plot the estimated financial networks at the beginning of each year from 2005 to 2013 in Fig.~\ref{fig:finance_net}. In this figure, the banks are clustered according to their regions automatically, as banks in the same region are expected to have more interactions. We further choose the cluster of US, Europe (including UK), and Asia-Pacific and depict the average number of connections for banks in each cluster as a function of time in Fig.~\ref{sfig:bank_BADGE}. Both Fig.~\ref{sfig:bank_BADGE} and Fig.~\ref{fig:finance_net} tell us that number of connections for US and European banks is larger than that of banks in Asia-Pacific. In light of the theory of system risk, the financial institutions with more connections are called central institutions (i.e., sit in the center of the financial system). Such institutions are more sensitive to financial crises, and conversely, their failure can lead to the breakdown of the entire system with a larger probability~\cite{BGLP:2012}. With this theory in mind, we can now analyze how the networks changed during the financial crisis. Before the Lehman bankruptcy, the average number of connections for US banks first increased in 2006, due to the Government's unexpected decision to tighten monetary policy in May and June that year~\cite{DDLY:2018}. There was no other major shock in 2006 though, and therefore, the average number of connections for US banks decreased later in 2006~\cite{DDLY:2018}. In early 2007, the collapse of several mortgage originators led to the sharp increase of connections in US~\cite{Williams:2010}. Due to the large number of connections between banks in US and Europe, the European banks also lost a tremendous amount on toxic assets and from bad loans. Consequently, the connections for European banks also increased. In late 2007 and 2008, the 2007 subprime mortgage crisis in US finally led to the global financial crisis~\cite{Williams:2010}, since US sat in the center of the network. We can see that the entire financial network in 2008 (see Fig.~\ref{sfig:08_01}-\ref{sfig:09_01}) is much denser than the one before and after the crisis (see Fig.~\ref{sfig:05_01} and Fig.~\ref{sfig:13_12}). In the post-Lehman period, the US market calmed after the government injected a massive amount of capital into major US banks~\cite{Williams:2010}. Thus, the number of connections for US banks decreased correspondingly. On the other hand, the European debt crisis occurred in Greece in late 2009 and further spread to Ireland and Portugal~\cite{Constancio:2012}. The delay of the rescue package for Greece caused the second peak of connections for European banks in May 2010. Later in 2011, the debt crisis further affected Spain and Italy~\cite{Constancio:2012}, leading to the third peak for Europe from June to August in 2011 in Fig.~\ref{sfig:bank_BADGE}. Note that the number of connections for US banks was low when the debt crisis first happened in 2009. However, due to risk contagion in the financial network, the number of connections for US banks also reached another peak in 2011. On the other hand, since there is a fewer number of connections between Asia-Pacific countries and US and Europe (see Fig.~\ref{fig:finance_net}), the financial crisis did not impact Asia-Pacific countries as severely as US and Europe.

\begin{table*}[t!]
\renewcommand{\arraystretch}{1.4}
  \vspace{-1mm}
  \centering
  \caption{Graph recovery results from BASS and GMS for synthetic time series with different dimensions ($P = 20, 100$, $N = 1000$, $N_e = P$). The standard deviations are shown in the brackets.}
  \vspace{-1mm}
  \resizebox{0.7\linewidth}{!}{
    \begin{tabular}{c|cccc|cccc}
    \hline
    \hline
    \multirow{2}{*}{Methods}               &\multicolumn{4}{c|}{$P = 20$}         &\multicolumn{4}{c}{$P = 100$}\\
    \cline{2-9}
                                        &Precision      &Recall         &$F_1$-score    &Time(s)        &Precision  &Recall &$F_1$-score    &Time(s)\\
    \hline
    BASS                               &0.90 (3.14e-2) &0.88 (4.23e-2) &0.89 (3.09e-2) &2.18e2 (2.10e1)    &0.88 (4.90e-2) &0.71 (3.92e-2) &0.78 (4.20e-2) &4.75e3 (2.85e2)\\
    GMS (orcale)                        &0.83 (1.05e-1) &0.72 (8.27e-2) &0.76 (7.17e-2) &3.36e3 (5.13e1)    &0.71 (1.94e-2) &0.79 (2.02e-2) &0.75 (7.21e-3) &4.16e5 (6.20e3)\\
    GMS (CV)                            &0.63 (1.02e-1) &0.78 (1.20e-1) &0.69 (8.18e-2) &4.32e3 (7.93e2)    &0.71 (7.69e-2) &0.68 (2.53e-2) &0.69 (3.69e-2) &5.05e5 (9.34e3)\\
    \hline
    \hline
    \end{tabular}    }%
    \label{tab:syn_freq_dim}
    \vspace{-1mm}
\renewcommand{\arraystretch}{1}
\end{table*}
\begin{table*}[t!]
\renewcommand{\arraystretch}{1.4}
  \centering
  \caption{Graph recovery results from BASS and GMS for synthetic time series with different lengths ($P = 20$, $N = 500, 1000, 2000$, $N_e = P$). The standard deviations are shown in the brackets.}
  \vspace{-1mm}
  \resizebox{\linewidth}{!}{
    \begin{tabular}{c|cccc|cccc|cccc}
    \hline
    \hline
    \multirow{2}{*}{Methods}               &\multicolumn{4}{c|}{$N = 500$}         &\multicolumn{4}{c|}{$N = 1000$}        &\multicolumn{4}{c}{$N = 2000$}\\
    \cline{2-13}
                                        &Precision  &Recall &$F_1$-score    &Time(s)    &Precision  &Recall &$F_1$-score    &Time(s)    &Precision  &Recall &$F_1$-score    &Time(s)          \\
    \hline
    BASS                               &0.90 (6.51e-2) &0.81 (9.32e-2) &0.85 (6.74e-2) &1.11e2 (8.26)      &0.90 (3.14e-2) &0.88 (4.23e-2) &0.89 (3.09e-2) &2.18e2 (2.10e1)    &0.89 (8.40e-2) &0.94 (3.60e-2) &0.91 (5.74e-2) &4.44e2 (5.56e1)\\
    GMS (orcale)                        &0.90 (1.15e-1) &0.70 (1.09e-1) &0.78 (9.26e-2) &1.70e3 (9.68)      &0.83 (1.05e-1) &0.72 (8.27e-2) &0.76 (7.17e-2) &3.36e3 (5.13e1)    &0.80 (2.74e-2) &0.73 (1.12e-1) &0.76 (6.04e-2) &7.28e3 (5.12e2)\\
    GMS (CV)                            &0.50 (3.17e-1) &0.76 (1.61e-1) &0.53 (1.94e-1) &2.76e3 (1.55e2)    &0.63 (1.02e-1) &0.78 (1.20e-1) &0.69 (8.18e-2) &4.32e3 (7.93e2)    &0.75 (7.56e-2) &0.73 (1.03e-1) &0.74 (7.36e-2) &9.01e3 (9.79e2)\\
    \hline
    \hline
    \end{tabular}    }%
    \label{tab:syn_freq_sample}
    \vspace{-1mm}
\renewcommand{\arraystretch}{1}
\end{table*}
\begin{table*}[t!]
\renewcommand{\arraystretch}{1.4}
  \centering
  \caption{Graph recovery results from BASS and GMS for synthetic time series with different graph density ($P = 20$, $N = 1000$, $N_e = 10, 20, 40$). The standard deviations are shown in the brackets.}
  \vspace{-1mm}
  \resizebox{\linewidth}{!}{
    \begin{tabular}{c|cccc|cccc|cccc}
    \hline
    \hline
    \multirow{2}{*}{Methods}               &\multicolumn{4}{c|}{$N_e = 10$}         &\multicolumn{4}{c|}{$N_e = 20$}        &\multicolumn{4}{c}{$N_e = 40$}\\
    \cline{2-13}
                                        &Precision  &Recall &$F_1$-score    &Time(s)    &Precision  &Recall &$F_1$-score    &Time(s)    &Precision  &Recall &$F_1$-score    &Time(s)          \\
    \hline
    BASS                               &0.93 (1.16e-1) &0.97 (6.88e-2) &0.95 (6.49e-2) &2.28e2 (1.18e1)    &0.90 (3.14e-2) &0.88 (4.23e-2) &0.89 (3.09e-2) &2.18e2 (2.10e1)    &0.93 (4.20e-2) &0.72 (2.89e-2) &0.81 (2.37e-2) &2.04e2 (7.90)\\
    GMS (orcale)                        &0.90 (6.87e-2) &0.81 (2.64e-2) &0.85 (3.77e-2) &3.46e3 (9.44e1)    &0.83 (1.05e-1) &0.72 (8.27e-2) &0.76 (7.17e-2) &3.36e3 (5.13e1)    &0.74 (7.61e-2) &0.68 (2.10e-2) &0.71 (4.12e-2) &3.68e3 (3.32e1)\\
    GMS (CV)                            &0.79 (1.62e-1) &0.80 (3.60e-2) &0.79 (8.35e-2) &4.70e3 (7.87e2)    &0.63 (1.02e-1) &0.78 (1.20e-1) &0.69 (8.18e-2) &4.32e3 (7.93e2)    &0.74 (7.61e-2) &0.68 (2.10e-2) &0.71 (4.12e-2) &5.03e3 (7.12e2)\\
    \hline
    \hline
    \end{tabular}    }%
    \label{tab:syn_freq_density}
    \vspace{-3mm}
\renewcommand{\arraystretch}{1}
\end{table*}

\subsection{Graphical Models for Time Series (Frequency Domain)}

\subsubsection{Synthetic Data}
To test the proposed method for inferring graphical models in the frequency domain, we consider simulated time series with length $N$ generated from a first order vector autoregressive process for $P$ variables. Specifically, we simulate data from the model
\begin{align}
\bm y^{(t)} = A \bm y^{(t-1)} + \bm\epsilon^{(t)},
\end{align}
where $\bm y^{(t)}\in\mathbb R^P$, $A\in\mathbb R^{P\times P}$, and $\bm\epsilon^{(t)}\sim\mathcal N(\bm 0,I)$. The inverse spectral density of the process is then given by~\cite{SV:2010}:
\begin{align}
K^{(\omega)} = I + A'A + \exp(-i\omega)A+\exp(i\omega)A'.
\end{align}
We consider time series with different dimensions, sample sizes, and graph density, and compare the results given by BASS with those of GMS~\cite{JHG:2015}. The graph density is characterized by the number of non-zero elements in $A$, which is denoted as $N_e$ in the sequel. For GMS, the kernel bandwidth $h$ is selected from $\{\exp(-4)N, \exp(-2)N, \cdots, \exp(4)N\}$ and $\lambda_1$ is chosen from $\{\exp(-4), \exp(-3), \cdots, 1\}$. Again, we select the tuning parameters by maximizing the $F_1$-score between the estimated and true graphs (i.e., the oracle results) as well as using CV. Before applying these methods, we first normalize the data $\bm y_{1:P}^{(1:N)}$ to have unit variance. We then apply the normalized Fourier transform to obtain the Fourier coefficients $\bm f_{1:P}^{(1:N)}$ such that the variance of $\bm f_{1:P}^{(1:N)}$ does not increase with $N$. The results averaged over 5 trials are summarized in Table~\ref{tab:syn_freq_dim}-\ref{tab:syn_freq_density}.

The results are similar to those in the time domain. Hence, we only provide a brief summary of the results here. BASS yields the best performance in the terms of the $F_1$-score with the least amount of computational time in all cases. The oracle results are the second best. CV cannot always select the optimal tuning parameters, and so the CV results are worse than the oracle results. We can also observe that the increase of dimension $P$ or the graph density $N_e$ deteriorates the performance, whereas the increase of the sample size $N$ improves the performance, as expected. The computational time of BASS and GMS is approximately a linear function of $NP^2$ and $NP^3$ respectively.

\subsubsection{Scalp EEG of AD Patients}

\begin{figure}[t]
\vspace{-1mm}
\centering
\subfloat[BASS]{
\includegraphics[height = 3.2cm,origin=c]{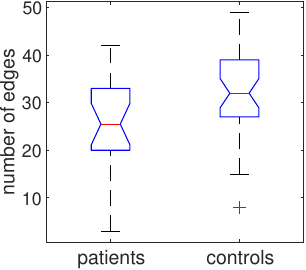}
\label{sfig:MCI_BADGE}}
\hspace{2pt}
\subfloat[GMS]{
\includegraphics[height = 3.2cm,origin=c]{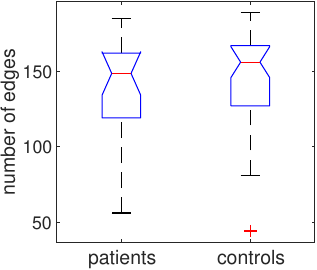}
\label{sfig:MCI_GSM}}
\vspace{-1mm}
\caption{Boxplots of the number of edges given by BASS and GMS for the first data set with MCI patients.}
\label{fig:MCI}
\vspace{-4mm}
\end{figure}
\begin{figure}[t]
\vspace{-1mm}
\centering
\subfloat[BASS]{
\includegraphics[height = 3.2cm,origin=c]{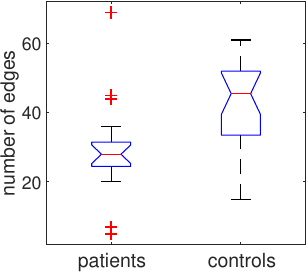}
\label{sfig:MAD_BADGE}}
\hspace{2pt}
\subfloat[GMS]{
\includegraphics[height = 3.2cm,origin=c]{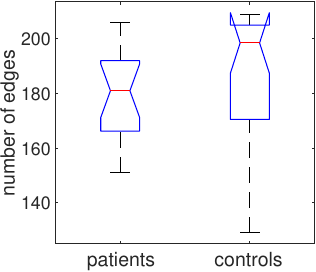}
\label{sfig:MAD_GSM}}
\vspace{-1mm}
\caption{Boxplots of the number of edges given by BASS and GMS for the second data set with Mild AD patients.}
\label{fig:MildAD}
\vspace{-4mm}
\end{figure}

In this section, we consider the problem of inferring functional brain networks from scalp EEG recordings. Specifically, we analyze two data sets. The first one contains 22 patients with mild cognitive impairment (MCI, a.k.a. pre-dementia) and 38 healthy control subjects~\cite{DVMC:2010}. The patients complained of the memory problem, and later on, they all developed mild AD (i.e., the first stage of AD). The second data set consists of 17 patients with mild AD and 24 control subjects~\cite{HIHGOWDV:2006}. For more details, please refer to Section C in the appendix. Although AD cannot be cured at present, existing symptom-delaying medications are proven to be more effective at early stages of AD, such as MCI and mild AD~\cite{DVMC:2010}. On the other hand, scalp EEG recording systems are inexpensive and potentially mobile, thus making it a useful tool to screen a large population for the risk of AD. As a result, it is crucial to identify the patients from scalp EEG signals at early stages of AD. Learning functional brain networks can help to distinguish between patients and healthy people~\cite{YD:2018}.

We first perform the normalized Fourier transform on all channels of EEG signals to obtain $\bm f^{(\omega)}$. We only consider $\bm f^{(\omega)}$ in the frequency band $4-30$Hz in order to filter out the noise in the signal. $K^{(\omega)}$ is then inferred from $\bm f^{(\omega)}$ by applying the proposed BASS algorithm.  We further split $K^{(\omega)}$ into three frequency ranges: $4-8$Hz, $8-12$Hz, and $12-30$Hz, as suggested by previous works on the same data sets~\cite{DVMC:2010,HIHGOWDV:2006}. For each frequency band, we infer the corresponding graphical models by finding the common zero patterns of all $K^{(\omega)}$ for $\omega$ in this band. We compare BASS with GMS (CV) in this experiment to check which method can yield graphs that can better distinguish between the patients and the control subjects. The candidate set of the kernel bandwidth $h$ and the penalty parameter $\lambda_1$ for GMS (CV) is respectively $\{\exp(-3)N, \exp(-2.5)N, \cdots, \exp(-1)N\}$ and $\{\exp(-5), \exp(-4.5), \cdots, \exp(-0.5)\}$

First, we count the number of edges in the graphical models, which can be regarded as a measure of synchrony between different EEG channels. We observe that graphical models in $4-8$Hz can best distinguish between patients and controls for both data sets and both methods. We depict in Fig.~\ref{fig:MCI}-\ref{fig:MildAD} the boxplots of the number of edges in the graphical models. Clearly, the graphical models for patients are more sparse than those for healthy people, and this phenomenon becomes more pronounced for Mild AD patients. Such findings are consistent with the loss of synchrony within the EEG signals for AD patients as reported in the literature~\cite{DVMC:2010,HIHGOWDV:2006}. We further conduct the Mann-Whitney test on the number of edges in the two sets of graphical models, respectively for the patients and the controls. The resulting p-value given by BASS for the two data sets is respectively $7.55\times 10^{-3}$ and $1.49\times 10^{-3}$, which are statistically significant. As a comparison, the p-value resulting from GMS is $3.99\times 10^{-1}$ for the MCI data and $9.26\times 10^{-2}$ for the Mild AD data. Next, we further train a random forest classifier based on the estimated brain networks to differentiate between the patients and the controls. The input to the classifier is the adjacency matrices associated with the networks. We then use the leave-one-out CV to evaluate the performance of the classifier. The accuracy yielded by BASS for the two data sets is $80.00\%$ and $85.37\%$ respectively, whereas that of GMS is $65.85\%$ and $70\%$. On the other hand, the computational time averaged over all subjects is $3.50\times 10^2$ seconds for BASS and $1.40\times 10^4$ seconds for GMS. Apparently, BASS can better describe the perturbations in the EEG synchrony for MCI and mild AD patients, while being more efficient.


\section{Conclusions}
\label{sec:conclusion}

In this paper, we propose a novel Bayesian model BASS to solve the problem of estimating dynamic graphical models. In contrast to the existing methods that have a high time complexity of $\sO(NP^3)$ and require extensive parameter tuning, the time complexity of BASS is only $\sO(NP^2)$ and it is free of tuning. Specifically, we develop a natural gradient based variational inference algorithm to learn the Bayesian model. To deal with the problem of local maxima, we resort to simulated annealing and propose to use bootstrapping to generate the annealing noise. Compared with the existing methods, BASS can better recover the true graphs with the least amount of computational time. We then apply BASS to analyze the stock return data of 78 banks worldwide, and observe that the resulting financial network becomes denser during the 2008 global financial crisis and the subsequent European debt crises. On the other hand, we find the resemblance between inferring time-varying inverse covariance matrices and frequency-varying inverse spectral density matrices, and extend BASS to learn graphical models among a multitude of stationary time series in the frequency domain. Results from EEG data of MCI and mild AD patients show that the proposed model may help to diagnose AD from scalp EEG at an early stage.

As BASS can only tackle Gaussian distributed data at present, we intend to extend it to non-Gaussian data by means of Gaussian copulas~\cite{YDW:2012} in future work. Additionally, it is interesting to extend BASS to deal with piece-wise constant graphical models~\cite{YD:2018}. In this case, the data can be partitioned into a certain number of time segments. The graph within each segment remains unchanged, but the graphs for every two consecutive segments can be completely different. Such piece-wise constant graphical models also find wide applications in practice~\cite{KEKZEC:2010, YD:2018}.



\begin{IEEEbiography}[{\includegraphics[width=1in,height=1.25in,clip,keepaspectratio]{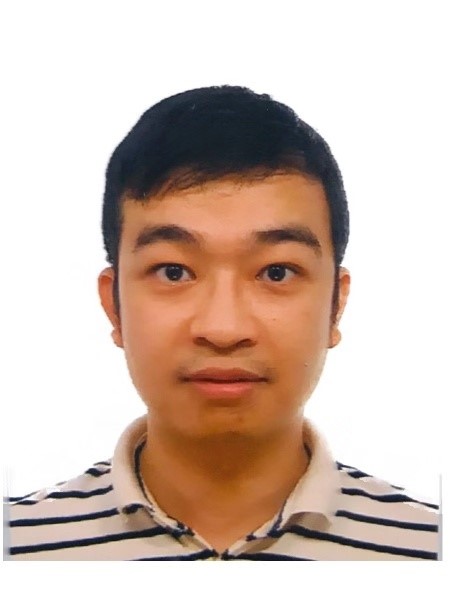}}]{Hang Yu}
(S'12-M'15) is a Senior Algorithm Expert at Ant Group, China. Before joining Ant Group, he was a Senior Research Fellow at Nanyang Technological University (NTU), Singapore. He received the Ph.D. degree in electrical and electronics engineering from NTU in 2015. His research interests include graphical models, Bayesian inference, stochastic optimization, extreme value statistics, deep learning, and machine learning. 
\end{IEEEbiography}

\vspace{-15pt}

\begin{IEEEbiography}[{\includegraphics[width=1in,height=1.25in,clip,keepaspectratio]{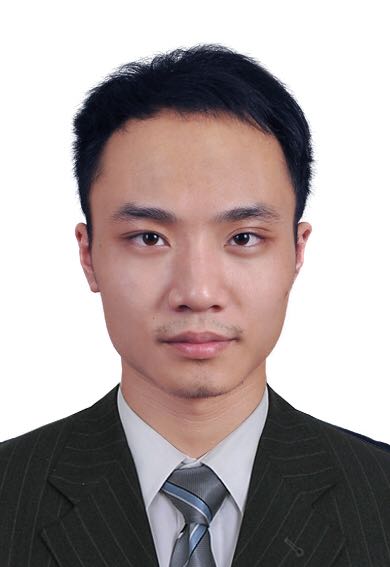}}]{Songwei Wu}
received his B.S. degree in statistics from Sun Yat-Sen University (SYSU), China in 2011, and the Ph.D. degree in electrical and electronic engineering from Nanyang Technological University(NTU), Singapore in 2020. He now works as a quantitative researcher. His research interest lies in convex optimization, machine learning, and quantitative finance.
\end{IEEEbiography}

\vspace{-15pt}

\begin{IEEEbiography}[{\includegraphics[width=1in,height=1.25in,clip,keepaspectratio]{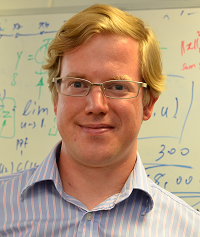}}]{Justin Dauwels}
(S'02-M'05-SM'12) is an Associate Professor at TU Delft since January 2021. Prior to this, he was an Associate Professor with the School of Electrical and Electronic Engineering at Nanyang Technological University (NTU), Singapore. His research interests are in Bayesian statistics, iterative signal processing, and computational neuroscience. He obtained the PhD degree in electrical engineering at the Swiss Polytechnical Institute of Technology (ETH) in Zurich in December 2005. He was a postdoctoral fellow at the RIKEN Brain Science Institute (2006-2007) and a research scientist at the Massachusetts Institute of Technology (2008-2010). He has been a JSPS postdoctoral fellow (2007), a BAEF fellow (2008), a Henri-Benedictus Fellow of the King Baudouin Foundation (2008), and a JSPS invited fellow (2010, 2011). His research team has won several best paper awards at international conferences. He has filed 5 US patents related to data analytics. 
\end{IEEEbiography}

\onecolumn
\renewcommand{\theequation}{S.\arabic{equation}}
\renewcommand{\thetable}{S.\arabic{table}}
\renewcommand{\thefigure}{S.\arabic{figure}}
\renewcommand{\thealgorithm}{S.\arabic{algorithm}}
\setcounter{equation}{0}
\setcounter{table}{0}
\setcounter{figure}{0}
\setcounter{algorithm}{0}
\appendices

	%
	\title{Supplementary Material for BASS}
	%
	%
	%
	
	\author{Hang Yu,~\IEEEmembership{Member,~IEEE,}, Songwei Wu,
		and~Justin Dauwels,~\IEEEmembership{Senior Member,~IEEE}}

	\maketitle

\section{Varational Inference and Natural Gradients}
Suppose that variables $\bm x$, $\bm z_1$, and $\bm z_2$ form a hierarchical Bayesian model, in which $\bm x$ is observed whereas $\bm z_1$ and $\bm z_2$ are the latent variables. The joint distribution can be factorized as: $p(\bm x, \bm z_1, \bm z_2) = p(\bm x|\bm z_1)p(\bm z_1|\bm z_2)p(\bm z_2)$. The ultimate goal of the variational inference is to approximate the exact but intractable posterior $p(\bm z_1, \bm z_2|\bm x)$ by a tractable variational distribution $q(\bm z_1, \bm z_2)$ that is closest in Kullback-Leibler (KL) divergence to $p(\bm z_1, \bm z_2|\bm x)$. Minimizing the KL divergence is equivalent to maximizing a lower bound of the evidence $\log p(\bm x)$, that is~\cite{Bishop:2006},
\begin{align}
\sL =  \E_{q(\bm z_1, \bm z_2)}[\log p(\bm x, \bm z_1,\bm z_2)]-\E_{q(\bm z_1, \bm z_2)}[\log q(\bm z_1, \bm z_2)], \label{eq:ELBO}
\end{align}
where $\E_{q(\bm z_1, \bm z_2)}$ is the expectation over the distribution $q(\bm z_1, \bm z_2)$ and $\sL$ is often referred to as evidence lower bound (ELBO)~\cite{Bishop:2006,HBWP:2013}. The inequality stems from Jensen's inequality and the equality holds when $q(\bm z_1, \bm z_2) = p(\bm z_1, \bm z_2|\bm x)$.

Typically, we apply the mean-field approximation and factorize the variational distribution as $q(\bm z_1, \bm z_2) = q(\bm z_1)q(\bm z_2)$. Next, we choose $q(\bm z_1)$ and $q(\bm z_2)$ that maximize the ELBO $\sL$. In the case where all distributions in the Bayesian model are from the exponential family and are conditionally conjugate, the classical expectation-maximization variational Bayes algorithm~\cite{Bishop:2006,HBWP:2013} provides an efficient tool to update the variational distributions. Suppose that $p(\bm z_2)$ takes the following exponential form:
\begin{align}
p(\bm z_2) \propto \exp\big(\bm\gamma' \bm\phi(\bm z_2)\big), \label{eq:p_z}
\end{align}
where $\bm\gamma$ is a vector of natural parameters (a.k.a. canonical parameters) and $\bm\phi(\bm z_2)=[\phi_1(\bm z_2),\cdots,\phi_m(\bm z_2)]$ denotes the vector of sufficient statistics. Since the prior $p(\bm z_2)$ is conjugate to the likelihood $p(\bm z_1|\bm z_2)$, we can express the likelihood in the same functional form as the prior w.r.t. $\bm z_2$:
\begin{align}
p(\bm z_1|\bm z_2) \propto \exp\big(\bm\phi(\bm z_2)'\bm\psi(\bm z_1)\big), \label{eq:p_y_z}
\end{align}
where $\bm\phi(\bm z_2)$ denotes the natural parameters, and $\bm\psi(\bm z_1)$ depending on $\bm z_1$ only denotes the sufficient statistics. The variational distribution $q(\bm z_i)$ for $i\in \{1,2\}$ that maximizes $\sL$ can be derived as~\cite{Bishop:2006,HBWP:2013}:
\begin{align}
q(\bm z_i) \propto&\ \exp\big\{\E_{q(\bm z_{-i})}[\log p(\bm x, \bm z_1, \bm z_2)]\big\} \label{eq:conj_update_rule} 
\end{align}
Note that the expectation inside the exponential is taken over all latent variables except the one whose variational distribution is to be updated. The expectation-maximization variational Bayes algorithm then cycles through the update rules for $q(\bm z_1)$ and $q(\bm z_2)$ until convergence. This algorithm fully exploits the geometry of the posterior and implicitly adopts natural gradients, resulting in simple close-form update rules and faster convergence than standard gradients~\cite{HBWP:2013}-\cite{KBLSS:2016}. However, \eqref{eq:conj_update_rule} yields a distribution with a close-form expression only under the conjugate scenario.

In the case where the pair of prior and the likelihood is not conjugate, we first specify the functional form of the variational distribution, compute the natural gradient w.r.t. the natural parameters of the distribution, and then follow the direction of the natural gradients to update these parameters. Concretely, we still specify the variational distribution to be the exponential family distributions, due to their generality and many useful algebraic properties. As such, $q(\bm z_2)$ can be expressed as:
\begin{align}
q(\bm z_2|\bm\theta) = \exp\big[\bm\theta'\bm\phi(\bm z_2)-A(\bm\theta)\big],
\end{align}
where $\bm\theta$ is a vector of natural parameters and $A(\bm\theta) = \log \int \exp[\bm\theta'\bm\phi(\bm z_2)]d\bm z_2$ is the log-partition function. We call the above representation minimal if all components of the vector of sufficient statistics $\bm\phi(\bm z_2) = [\phi_1(\bm z_2),\cdots,\phi_m(\bm z_2)]$ are linearly independent for all $\bm z_2$. Minimal representation suggests that every distribution $q(\bm z_2|\bm\theta)$ has a unique natural parameterization $\bm\theta$. We further define the mean parameter vector as $\bm\eta = \E[\bm\phi(\bm z_2)]$. It is easy to show that:
\begin{align}
\bm\eta = \nabla_{\bm\theta} A(\bm\theta). \label{eq:mapping_mu_theta}
\end{align}
Note that this mapping is one-to-one if and only if the representation is minimal. Next, we consider optimizing the natural parameters $\bm\theta$ of the variational distribution following the direction of the natural gradient. The natural gradient pre-multiplies the standard gradient by the inverse of the Fisher information matrix $\sI$. In particular for distributions in the minimal exponential family (i.e., $q(\bm z_2|\bm\theta)$), the resulting Fisher information matrix is given by:
\begin{align}
\sI(\bm\theta) &= -\E_{q(\bm z_2)}\big[\nabla_{\bm\theta}^2\log q(\bm z_2|\bm\theta)\big]= \nabla_{\bm\theta}^2 A(\bm\theta) = \frac{\partial\bm\eta}{\partial\bm\theta}.
\end{align}
The last equality follows directly from~\eqref{eq:mapping_mu_theta}. Thus, the natural gradient of $\sL$ w.r.t. $\bm\theta$ can be simplified as: 
\begin{align}
\sI(\bm\theta)^{-1}\nabla_{\bm\theta}\sL = \sI(\bm\theta)^{-1} \frac{\partial\bm\eta}{\partial\bm\theta}\nabla_{\bm\eta}\sL = \nabla_{\bm\eta}\sL.
\end{align}
In other words, for variational distributions in the minimal exponential family, the natural gradient of the ELBO $\sL$ w.r.t. to the natural parameters $\bm\theta$ is equivalent to the standard gradient of $\sL$ w.r.t. the corresponding mean parameters $\bm\eta$. We further notice that the second term in $\sL$~\eqref{eq:ELBO} is the entropy of $q(\bm z_2|\bm\theta)$ and its gradient w.r.t. $\bm\eta$ is $-\bm\theta$. As a result, let $\sL_1 = \E_{q(\bm z_1, \bm z_2)}[\log p(\bm x, \bm z_1,\bm z_2)]$ denote the first term in $\sL$~\eqref{eq:ELBO} and $0<\rho\leq 1$ be the step size, the update rule of $\bm\theta$ for the natural gradient algorithm is:
\begin{align}
\bm\theta^{\{i+1\}} 
&= (1-\rho)\bm\theta^{\{i\}} + \rho \nabla_{\bm\eta}\sL_1\big(\bm\theta^{\{i\}}\big). \label{eq:nonconj_update_rule}
\end{align}
Note that the above natural gradient update rule amounts to the expectation-maximization update rule in~\eqref{eq:conj_update_rule} when $p(\bm z_2)$ and $p(\bm z_1|\bm z_2)$ are conjugate as in~\eqref{eq:p_z} and~\eqref{eq:p_y_z} and $\rho = 1$. For instance, the update rule for $q(\bm z_2)$ resulting from both algorithms is the same and can be written as:
\begin{align}
q(\bm z_2) \propto \exp\Big\{\big[\bm\gamma + \langle\bm\psi(\bm z_1)\rangle\big]'\bm\phi(\bm z_2)\Big\}.
\end{align}
In other words, the expectation-maximization algorithm is a special case of the natural gradient algorithm when the Bayesian model is conditionally conjugate. The natural gradient variational inference algorithm is guaranteed to achieve linear convergence with a constant step size $\rho$ under mild conditions~\cite{KBLSS:2016}. In practice, to further accelerate the convergence, we follow~\cite{HRKTK:2010} to set $\rho = 1$ for the conjugate pairs (i.e., applying the expectation-maximization variational Bayes update rule in~\eqref{eq:conj_update_rule}) and to use the line search method to determine the step size $\rho$ for the non-conjugate pairs. The convergence with this step size scheme can be easily proven as in~\cite{SRG:2002}, since the ELBO is guaranteed to increase in every iteration.

\section{Derivation of the ELBO and the Variational Inference Algorithm}

\begingroup\abovedisplayskip=-3pt \belowdisplayskip=-3pt \abovedisplayshortskip = -3pt \belowdisplayshortskip = -3pt
\begin{table*}[t]
\caption{ELBO and detailed update rules of BASS.}
\label{tab:update_rule_BADGE}
 \fontsize{6.5}{7}\selectfont 
\begin{tabular}{m{0.976\linewidth}}
\hline
{\begin{align}
&\sL_1 = \sum_{t=1}^N \sum_{j=1}^P\Big[\frac{1}{2}\langle\kappa_j^{(t)}\rangle - \frac{1}{2}\langle\exp\big(\kappa_j^{(t)}\big)\rangle {x_j^{(t)}}^2 - x_j^{(t)} \langle K_{j,-j}^{(t)}\rangle\bm x_{-j}^{(t)} -\frac{1}{2}\langle\exp\big(-\kappa_j^{(t)}\big)\rangle{\bm x_{-j}^{(t)}}'\langle K_{-j,j} K_{j,-j} \rangle\bm x_{-j}^{(t)}\Big] + \sum_{j=1}^P\sum_{k = j + 1}^P\Big[\langle\delta(s_{jk}^{(1)} = 1)\rangle\langle\log\pi_1\rangle+\notag \\
&\quad\quad \langle\delta(s_{jk}^{(1)} = 0)\rangle\langle\log(1-\pi_1)\rangle\Big] + \sum_{t=2}^N \Bigg\{\sum_{j=1}^P\bigg\{\sum_{k = j+1}^P \Big[\langle\delta(s_{jk}^{(t-1)} = 0, s_{jk}^{(t)} = 0)\rangle\langle\log A_{00}\rangle + \langle\delta(s_{jk}^{(t-1)} = 0, s_{jk}^{(t)} = 1)\rangle\langle\log(1 - A_{00})\rangle +\notag \\
&\quad\quad \langle\delta(s_{jk}^{(t-1)} = 1, s_{jk}^{(t)} = 0)\rangle\langle\log(1- A_{11})\rangle+ \langle\delta(s_{jk}^{(t-1)} = 1, s_{jk}^{(t)} = 1)\rangle\langle\log A_{11}\rangle -\frac{1}{2}\langle\alpha_{jk}\rangle\langle\big(J_{jk}^{(t)} -J_{jk}^{(t-1)}\big)^2\rangle\Big]- \frac{1}{2}\langle\beta\rangle\langle\big(\kappa_j^{(t)}-\kappa_j^{(t-1)}\big)^2\rangle\bigg\}\Bigg\}\notag \\
&\quad\quad + \Big(\frac{N-1}{2}-1\Big)\sum_{j=1}^P\sum_{k=j+1}^P\langle\log\alpha_{jk}\rangle + \Big(\frac{P(N-1)}{2}-1\Big)\langle\log\beta\rangle. \label{eq:sL_1} \\
&\frac{\partial \sL_1}{\partial \langle K_{jk}^{(t)}\rangle} =  - 2 x_j^{(t)}x_k^{(t)} -\langle\exp(-\kappa_j^{(t)})\rangle x_{k}^{(t)}\langle K_{j,-jk}^{(t)}\rangle \bm x_{-jk}^{(t)}
-\langle\exp(-\kappa_k^{(t)})\rangle x_{j}^{(t)}\langle K_{k,-jk}^{(t)}\rangle \bm x_{-jk}^{(t)}. \label{eq:d_L1_d_K}\\
&\frac{\partial \sL_1}{\partial \langle {K_{jk}^{(t)}}^2\rangle} = -\frac{1}{2}\Big(\langle\exp(-\kappa_j^{(t)})\rangle{x_k^{(t)}}^2 + \langle\exp(-\kappa_k^{(t)})\rangle{x_j^{(t)}}^2)\Big).\label{eq:d_L1_d_K2} \\
&\varphi_\sV^s(s_{jk}^{(t)}) = \begin{cases}
\delta(s_{jk}^{(t)} = 1)\langle\log\pi_1\rangle+ \delta(s_{jk}^{(t)} = 0)\langle\log(1-\pi_1)\rangle- \bigg(\dfrac{\partial \sL_1}{\partial \langle K_{jk}^{(t)}\rangle}\langle J_{jk}^{(t)}\rangle + 2 \dfrac{\partial \sL_1}{\partial \langle {K_{jk}^{(t)}}^2\rangle}\langle {J_{jk}^{(t)}}^2\rangle\bigg)s_{jk}^{(t)}, & t = 1. \\[10pt]
-\bigg(\dfrac{\partial \sL_1}{\partial \langle K_{jk}^{(t)}\rangle}\langle J_{jk}^{(t)}\rangle + 2 \dfrac{\partial \sL_1}{\partial \langle {K_{jk}^{(t)}}^2\rangle}\langle {J_{jk}^{(t)}}^2\rangle\bigg)s_{jk}^{(t)}, & t > 1.
\end{cases} \label{eq:node_potential_s}\\
&\varphi_\sE^s(s_{jk}^{(t)}, s_{jk}^{(t-1)}) = \delta(s_{jk}^{(t-1)} = 0, s_{jk}^{(t)} = 1)\langle\log A_{01}\rangle + \delta(s_{jk}^{(t-1)} = 0, s_{jk}^{(t)} = 0)\langle\log(1 - A_{01})\rangle + \delta(s_{jk}^{(t-1)} = 1, s_{jk}^{(t)} = 1)\langle\log A_{11}\rangle \notag \\
&\quad\quad + \delta(s_{jk}^{(t-1)} = 1, s_{jk}^{(t)} = 0)\langle\log(1 - A_{11}).\label{eq:edge_potential_s} \\
&\varphi_\sV^J(J_{jk}^{(t)}) = \begin{cases}
\bigg(\dfrac{\partial \sL_1}{\partial \langle {K_{jk}^{(t)}}^2\rangle}\langle s_{jk}^{(t)} \rangle -\dfrac{1}{2}\langle\alpha_{jk}\rangle\bigg){J_{jk}^{(t)}}^2 + \dfrac{\partial \sL_1}{\partial \langle K_{jk}^{(t)}\rangle} \langle s_{jk}^{(t)}\rangle J_{jk}^{(t)}, & t\in\{1, N\}. \\[10pt]
\bigg(\dfrac{\partial \sL_1}{\partial \langle {K_{jk}^{(t)}}^2\rangle}\langle s_{jk}^{(t)} \rangle -\langle\alpha_{jk}\rangle\bigg){J_{jk}^{(t)}}^2 + \dfrac{\partial \sL_1}{\partial \langle K_{jk}^{(t)}\rangle} \langle s_{jk}^{(t)}\rangle J_{jk}^{(t)}, & 1 < t < N.
\end{cases} \label{eq:node_potential_J}\\
&\varphi_\sE^J(J_{jk}^{(t)}, J_{jk}^{(t-1)}) = -\langle\alpha_{jk}\rangle J_{jk}^{(t)}J_{jk}^{(t-1)}. \label{eq:edge_potential_J} \\
&\frac{\partial\sL_1}{\partial\langle\exp(-\kappa_j^{(t)})\rangle} = -\frac{1}{2}{\bm x_{-j}^{(t)}}'\langle K_{-j,j}^{(t)} K_{j,-j}^{(t)} \rangle\bm x_{-j}^{(t)} = -\frac{1}{2}\Big({\bm x_{-j}^{(t)}}'\langle K_{-j,j}^{(t)}\rangle\langle K_{j,-j}^{(t)} \rangle\bm x_{-j}^{(t)} + \V[K_{j,-j}^{(t)}] {\bm x_{-j}^{(t)}}^2\Big). \label{eq:d_L1_d_Kd_inv}\\
&h_t^{\{i+1\}} = (1-\rho) h_t^{\{i\}} + \frac{\rho}{2}\Big[1 - {x_j^{(t)}}^2 \langle\exp(\kappa_j^{(t)})\rangle(1 - \langle\kappa_j^{(t)}\rangle) -2\frac{\partial\sL_1}{\partial\langle\exp(-\kappa_j^{(t)})\rangle}\langle\exp(-\kappa_j^{(t)})\rangle(1 + \langle\kappa_j^{(t)}\rangle)\Big]. \label{eq:node_potential_kappa}\\
&\Omega_{t,t}^{\{i+1\}} = \begin{cases}
(1-\rho) \Omega_{t,t}^{\{i+1\}} + \rho\bigg\{\langle\beta\rangle + \dfrac{1}{2}\Big[{x_j^{(t)}}^2\langle\exp(\kappa_j^{(t)})\rangle   -2\dfrac{\partial\sL_1}{\partial\langle\exp(-\kappa_j^{(t)})\rangle}\langle\exp(-\kappa_j^{(t)})\rangle\Big]\bigg\},& t\in\{1,N\}. \\[10pt] 
(1-\rho) \Omega_{t,t}^{\{i+1\}} + \rho\bigg\{2\langle\beta\rangle + \dfrac{1}{2}\Big[{x_j^{(t)}}^2\langle\exp(\kappa_j^{(t)})\rangle  -2\dfrac{\partial\sL_1}{\partial\langle\exp(-\kappa_j^{(t)})\rangle}\langle\exp(-\kappa_j^{(t)})\rangle\Big]\bigg\}, & 1<t<N.
\end{cases}\label{eq:node_potential_kappa}\\
&\Omega_{t,t-1}^{\{i+1\}} = (1-\rho)\Omega_{t,t-1}^{\{i+1\}} - \rho \langle\beta\rangle. \label{eq:edge_potential_kappa}
\end{align}}
\end{tabular}
\begin{tabular}{m{0.4765\textwidth}m{0.4765\textwidth}}
\begin{align}
a = 1 + \sum_j\sum_k q(s_{jk}^{(1)} = 1). \label{eq:update_rule_pi}
\end{align} &
\begin{align}
b = 1 + \sum_j\sum_k q(s_{jk}^{(1)} = 0).
\end{align} \\
\begin{align}
c_0 = 1 +\sum_j\sum_k\sum_{t=2}^N q(s_{jk}^{(t)} = 0, s_{jk}^{(t-1)} = 0).
\end{align} &
\begin{align}
d_0 = 1 + \sum_j\sum_k\sum_{t=2}^N q(s_{jk}^{(t)} = 1, s_{jk}^{(t-1)} = 0).
\end{align} \\
\begin{align}
c_1 = 1 + \sum_j\sum_k\sum_{t=2}^N q(s_{jk}^{(t)} = 1, s_{jk}^{(t-1)} = 1).
\end{align} &
\begin{align}
d_1 = 1 +\sum_j\sum_k\sum_{t=2}^N q(s_{jk}^{(t)} = 0, s_{jk}^{(t-1)} = 1). \label{eq:update_rule_A11}
\end{align} \\
\hline
\end{tabular}
\vspace{-3mm}
\end{table*}
\endgroup

In this section, we derive the closed-form expressions of the first term $\sL_1$ in the ELBO and the variational inference algorithm for BASS.

Recall that the joint of distribution can be factorized as:
\begin{align}
&\, p(\bm x^{(1:N)}, \bm s^{(1:N)}, J^{(1:N)}, \bm\kappa^{(1:N)}, \pi_1, A_{00}, A_{11}, \bm\alpha, \beta) \notag \\
=&\, p(\bm x^{(1:N)}|\bm s^{(1:N)}, J^{(1:N)}, \bm\kappa^{(1:N)}) p(\bm s^{(1:N)}|\pi_1, A_{00}, A_{11}) p(J^{(1:N)}|\bm\alpha)p(\bm\kappa^{(1:N)}|\beta)p(\pi_1)p(A_{00})p(A_{11})p(\bm\alpha)p(\beta) \notag \\
=&\, \prod_{j = 1}^P \prod_{t=1}^N p(x_j^{(t)}|\kappa_j^{(t)}, J_{j,-j}^{(t)}, \bm s_{j, -j}^{(t)})\prod_{j = 1}^P \prod_{k = j+1}^P \Big[p(\bm s_{jk}^{(1:N)}|\pi_1, A_{00}, A_{11}) p(J_{jk}^{(1:N)}|\alpha_{jk}) p(\alpha_{jk})\Big]\prod_{j = 1}^P p(\bm\kappa_j^{(1:N)}|\beta) \notag \\
&\, \cdot p(\pi_1) p(A_{00}) p(A_{11}) p(\beta). \label{eq:joint_dist}
\end{align}
Correspondingly, the first term $\sL_1$ in the ELBO can be expanded as:
\begin{align}
\sL_1 =&\, \E_q\big[\log p(\bm x^{(1:N)}, \bm s^{(1:N)}, J^{(1:N)}, \bm\kappa^{(1:N)}, \pi_1, A_{00}, A_{11}, \bm\alpha, \beta)\big] \notag \\
=&\, \sum_{j = 1}^P \sum_{t=1}^N \E_q\big[\log p(x_j^{(t)}|\kappa_j^{(t)}, J_{j,-j}^{(t)}, \bm s_{j, -j}^{(t)})\big] + \sum_{j = 1}^P \sum_{k = j+1}^P \Big\{\E_q\big[\log p(\bm s_{jk}^{(1:N)}|\pi_1, A_{00}, A_{11})\big] + \E_q\big[\log p(J_{jk}^{(1:N)}|\alpha_{jk})\big] + \notag \\
&\,\E_q\big[\log  p(\alpha_{jk})\big]\Big\} + \sum_{j = 1}^P \E_q\big[\log p(\bm\kappa_j^{(1:N)}|\beta)\big] +\E_q\big[\log p(\pi_1)\big] + \E_q\big[\log p(A_{00})\big] + \E_q\big[\log p(A_{11})\big] + \E_q\big[\log p(\beta)\big].  \label{eq:first_term}
\end{align}
Note the that expectations are over the variational distribution $q(\bm s^{(1:N)}, J^{(1:N)}, \bm\kappa^{(1:N)}, \pi_1, A_{00}, A_{11}, \bm\alpha, \beta)$. Next, we focus on each expectation individually:
\begin{align}
\E_q\big[\log p(x_j^{(t)}|\kappa_j^{(t)}, J_{j,-j}^{(t)}, \bm s_{j, -j}^{(t)})\big] =&\, \frac{1}{2}\langle\kappa_j^{(t)}\rangle - \frac{1}{2}\langle\exp\big(\kappa_j^{(t)}\big)\rangle {x_j^{(t)}}^2 - x_j^{(t)} \langle K_{j,-j}^{(t)}\rangle\bm x_{-j}^{(t)} \notag \\
&\, -\frac{1}{2}\langle\exp\big(-\kappa_j^{(t)}\big)\rangle{\bm x_{-j}^{(t)}}'\langle K_{-j,j} K_{j,-j} \rangle\bm x_{-j}^{(t)}, \label{eq:log_p_x}\\
\E_q\big[\log p(\bm s_{jk}^{(1:N)}|\pi_1, A_{00}, A_{11})\big] =&\, \langle\delta(s_{jk}^{(1)} = 1)\rangle\langle\log\pi_1\rangle + \langle\delta(s_{jk}^{(1)} = 0)\rangle\langle\log(1-\pi_1)\rangle + \langle\delta(s_{jk}^{(t-1)} = 0, s_{jk}^{(t)} = 0)\rangle\langle\log A_{00}\rangle\notag \\
&\, + \langle\delta(s_{jk}^{(t-1)} = 0, s_{jk}^{(t)} = 1)\rangle\langle\log(1 - A_{00})\rangle + \langle\delta(s_{jk}^{(t-1)} = 1, s_{jk}^{(t)} = 0)\rangle\langle\log(1- A_{11})\rangle \notag \\
&\, + \langle\delta(s_{jk}^{(t-1)} = 1, s_{jk}^{(t)} = 1)\rangle\langle\log A_{11}\rangle, \\
\E_q\big[\log p(J_{jk}^{(1:N)}|\alpha_{jk}) =&\, \frac{N-1}{2}\langle\log\alpha_{jk}\rangle-\frac{1}{2}\langle\alpha_{jk}\rangle\sum_{t=2}^N\langle\big(J_{jk}^{(t)} -J_{jk}^{(t-1)}\big)^2\rangle, \\
\E_q\big[\log  p(\alpha_{jk})\big] =&\, - \langle\log\alpha_{jk}\rangle,\\
\E_q\big[\log p(\bm\kappa_j^{(1:N)}|\beta)\big] =&\, \frac{P(N-1)}{2}\langle\log\beta\rangle - \frac{1}{2}\langle\beta\rangle\sum_{t=2}^N\langle\big(\kappa_j^{(t)}-\kappa_j^{(t-1)}\big)^2\rangle, \\
\E_q\big[\log  p(\beta)\big] =&\, - \langle\log\beta\rangle,\\
\E_q\big[\log p(\pi_1)\big] =&\, \E_q\big[\log p(A_{00})\big] = \E_q\big[\log p(A_{11})\big] = 0 \label{eq:log_p_pi1}.
\end{align}
By substituting~\eqref{eq:log_p_x}-\eqref{eq:log_p_pi1} into~\eqref{eq:first_term}, we can obtain the expression of $\sL_1$ in~\eqref{eq:sL_1} in Table~\ref{tab:update_rule_BADGE}.

On the other hand, we factorize the variational distribution $q(\bm s^{(1:N)}, J^{(1:N)}, \bm\kappa^{(1:N)}, \pi_1, A_{00}, A_{11}, \bm\alpha, \beta)$ as:
\begin{align}
&\, q(\bm s^{(1:N)}, J^{(1:N)}, \bm\kappa^{(1:N)}, \pi_1, A_{00}, A_{11}, \bm\alpha, \beta) \notag \\
=&\, \prod_{j=1}^P\prod_{k = j+1}^P \Big[q(\bm s_{jk}^{(1:N)}) q(J_{jk}^{(1:N)})q(\alpha_{jk})\Big] \prod_{j=1}^P q(\bm\kappa_j^{(1:N)})q(\pi_1)q(A_{00})q(A_{11})q(\beta). \label{eq:var_factor}
\end{align}
Recall that the priors and likelihoods of all variables except $\bm\kappa_j^{(1:N)}$ are conjugate to each other. For the conjugate pairs, the variational distributions can be updated following the expectation maximization variational Bayes algorithm in~\eqref{eq:conj_update_rule}. Specifically,
\begin{align}
q(\bm s_{jk}^{(1:N)}) \propto&\, \exp\Big\{\E_{q(\bm s_{-jk}^{(1:N)}, J^{(1:N)}, \bm\kappa^{(1:N)}, \pi_1, A_{00}, A_{11}, \bm\alpha, \beta)}\big[ \log p(\bm x^{(1:N)}, \bm s^{(1:N)}, J^{(1:N)}, \bm\kappa^{(1:N)}, \pi_1, A_{00}, A_{11}, \bm\alpha, \beta)\big]\Big\}, \\
q(J_{jk}^{(1:N)}) \propto&\, \exp\Big\{\E_{q(\bm s^{(1:N)}, J_{-jk}^{(1:N)}, \bm\kappa^{(1:N)}, \pi_1, A_{00}, A_{11}, \bm\alpha, \beta)}\big[ \log p(\bm x^{(1:N)}, \bm s^{(1:N)}, J^{(1:N)}, \bm\kappa^{(1:N)}, \pi_1, A_{00}, A_{11}, \bm\alpha, \beta)\big]\Big\}, \\
q(\alpha_{jk}) \propto&\, \exp\Big\{\E_{q(\bm s^{(1:N)}, J^{(1:N)}, \bm\kappa^{(1:N)}, \pi_1, A_{00}, A_{11}, \bm\alpha_{-jk}, \beta)}\big[ \log p(\bm x^{(1:N)}, \bm s^{(1:N)}, J^{(1:N)}, \bm\kappa^{(1:N)}, \pi_1, A_{00}, A_{11}, \bm\alpha, \beta)\big]\Big\},\\
q(\pi_1) \propto&\, \exp\Big\{\E_{q(\bm s^{(1:N)}, J^{(1:N)}, \bm\kappa^{(1:N)}, A_{00}, A_{11}, \bm\alpha, \beta)}\big[ \log p(\bm x^{(1:N)}, \bm s^{(1:N)}, J^{(1:N)}, \bm\kappa^{(1:N)}, \pi_1, A_{00}, A_{11}, \bm\alpha, \beta)\big]\Big\},\\
q(A_{00}) \propto&\, \exp\Big\{\E_{q(\bm s^{(1:N)}, J^{(1:N)}, \bm\kappa^{(1:N)}, \pi_1, A_{11}, \bm\alpha, \beta)}\big[ \log p(\bm x^{(1:N)}, \bm s^{(1:N)}, J^{(1:N)}, \bm\kappa^{(1:N)}, \pi_1, A_{00}, A_{11}, \bm\alpha, \beta)\big]\Big\},\\
q(A_{11}) \propto&\, \exp\Big\{\E_{q(\bm s^{(1:N)}, J^{(1:N)}, \bm\kappa^{(1:N)}, \pi_1, A_{00}, \bm\alpha, \beta)}\big[ \log p(\bm x^{(1:N)}, \bm s^{(1:N)}, J^{(1:N)}, \bm\kappa^{(1:N)}, \pi_1, A_{00}, A_{11}, \bm\alpha, \beta)\big]\Big\},\\
q(\beta) \propto&\, \exp\Big\{\E_{q(\bm s^{(1:N)}, J^{(1:N)}, \bm\kappa^{(1:N)}, \pi_1, A_{00}, A_{11}, \bm\alpha)}\big[ \log p(\bm x^{(1:N)}, \bm s^{(1:N)}, J^{(1:N)}, \bm\kappa^{(1:N)}, \pi_1, A_{00}, A_{11}, \bm\alpha, \beta)\big]\Big\}.
\end{align}
The expectations inside the exponential functions are taken over the variational distributions of all variables except the one to be updated in the expressions. By substituting~\eqref{eq:log_p_x}-\eqref{eq:log_p_pi1} into the above expressions and then removing the irrelevant constant terms, we can obtain the expressions of the variational distributions in (14)-(19) in the main body of the paper.

Next, let us focus on $q(\bm\kappa_j^{(1:N)})$ whose prior and likelihood are not conjugate. The expectation maximization variational Bayes algorithm will not work in this case, since the log-partition function is intractable for the following expression
\begin{align}
\exp\Big\{\E_{q(\bm s^{(1:N)}, J^{(1:N)}, \bm\kappa_{-j}^{(1:N)}, \pi_1, A_{00}, A_{11}, \bm\alpha, \beta)}\big[ \log p(\bm x^{(1:N)}, \bm s^{(1:N)}, J^{(1:N)}, \bm\kappa^{(1:N)}, \pi_1, A_{00}, A_{11}, \bm\alpha, \beta)\big]\Big\}. \label{eq:kappa_emvb}
\end{align}
Instead, we follow the natural gradient variational inference algorithm. Under this scenario, we need to specify the functional of $q(\bm\kappa_j^{(1:N)})$ first. More specifically, we specify $q(\bm\kappa_j^{(1:N)})$ to be a Gaussian distribution with precision matrix $\Omega$ and potential vector $h$. Moreover, we can tell from~\eqref{eq:kappa_emvb} that $q(\bm\kappa_j^{(1:N)})$ corresponds to Markov chain, in which $\kappa_j^{(t)}$ only have conditional dependence with $\kappa_j^{(t-1)}$ and $\kappa_j^{(t+1)}$. As a result, $\Omega$ should be a tri-diagonal matrix. On the other hand, $\sL_1$ in terms of $q(\bm\kappa_j^{(1:N)})$ can be expressed as:
\begin{align}
\sL_1(q(\bm\kappa_j^{(1:N)})) =&\, \frac{1}{2}\sum_{t=1}^N \Big[\langle\kappa_j^{(t)}\rangle -\langle\exp(\kappa_j^{(t)})\rangle {x_j^{(t)}}^2 -\langle\exp(-\kappa_j^{(t)})\rangle{\bm x_{-j}^{(t)}}'\langle K_{-j,j} K_{j,-j} \rangle\bm x_{-j}^{(t)}\Big] - \frac{\langle\beta\rangle}{2}\sum_{t=2}^N\langle(\kappa_j^{(t)}- \kappa_j^{(t-1)})^2\rangle, \notag \\
\end{align}
We then compute the gradient of $\sL_1$ w.r.t. the mean parameters $\langle\kappa_j^{(t)}\rangle$, $-\langle{\kappa_j^{(t)}}^2\rangle/ 2$, and $-\langle\kappa_j^{(t-1)}\kappa_j^{(t)}\rangle$ as:
\begin{align}
\frac{\partial\sL_1}{\langle\kappa_j^{(t)}\rangle} =&\, \frac{1}{2}\Big[1 - {x_j^{(t)}}^2 \langle\exp(\kappa_j^{(t)})\rangle(1 - \langle\kappa_j^{(t)}\rangle) -2\frac{\partial\sL_1}{\partial\langle\exp(-\kappa_j^{(t)})\rangle}\langle\exp(-\kappa_j^{(t)})\rangle(1 + \langle\kappa_j^{(t)}\rangle)\Big], \label{eq:grad_L1_h} \\
\frac{\partial\sL_1}{-\langle{\kappa_j^{(t)}}^2\rangle / 2} =&\,\begin{cases}
\langle\beta\rangle + \dfrac{1}{2}\Big[{x_j^{(t)}}^2\langle\exp(\kappa_j^{(t)})\rangle   -2\dfrac{\partial\sL_1}{\partial\langle\exp(-\kappa_j^{(t)})\rangle}\langle\exp(-\kappa_j^{(t)})\rangle\Big],& t\in\{1,N\}, \\[12pt]
2\langle\beta\rangle + \dfrac{1}{2}\Big[{x_j^{(t)}}^2\langle\exp(\kappa_j^{(t)})\rangle  -2\dfrac{\partial\sL_1}{\partial\langle\exp(-\kappa_j^{(t)})\rangle}\langle\exp(-\kappa_j^{(t)})\rangle\Big], & 1<t<N,
\end{cases}\\
\frac{\partial\sL_1}{-\langle\kappa_j^{(t-1)}\kappa_j^{(t)}\rangle} =&\, -\langle\beta\rangle. \label{eq:grad_L1_Omega}
\end{align}
where
\begin{align}
\langle\exp(\kappa_j^{(t)})\rangle =&\,\exp\Big[\langle\kappa_j^{(t)}\rangle + \frac{1}{2}\big(\langle{\kappa_j^{(t)}}^2\rangle - \langle\kappa_j^{(t)}\rangle^2\big) \Big], \\
\langle\exp(-\kappa_j^{(t)})\rangle =&\,\exp\Big[-\langle\kappa_j^{(t)}\rangle + \frac{1}{2}\big(\langle{\kappa_j^{(t)}}^2\rangle - \langle\kappa_j^{(t)}\rangle^2\big) \Big], \\
\frac{\partial\sL_1}{\partial\langle\exp(-\kappa_j^{(t)})\rangle} =&\, -\frac{1}{2}{\bm x_{-j}^{(t)}}'\langle K_{-j,j}^{(t)} K_{j,-j}^{(t)} \rangle\bm x_{-j}^{(t)}.
\end{align}
It follows from the natural gradient variational inference algorithm~~\eqref{eq:nonconj_update_rule} that
\begin{align}
h_t^{\{i+1\}} =&\, (1-\rho) h_t^{\{i\}} + \rho \frac{\partial\sL_1}{\langle\kappa_j^{(t)}\rangle}, \label{eq:update_h}\\
\Omega_{t,t}^{\{i+1\}} =&\, (1-\rho) \Omega_{t,t}^{\{i+1\}} + \rho + \rho \frac{\partial\sL_1}{-\langle{\kappa_j^{(t)}}^2\rangle / 2}, \\
\Omega_{t,t-1}^{\{i+1\}} =&\, (1-\rho)\Omega_{t,t-1}^{\{i+1\}} + \rho \frac{\partial\sL_1}{-\langle\kappa_j^{(t-1)}\kappa_j^{(t)}\rangle}. \label{eq:update_Omega}\\
\end{align}
Substitute~\eqref{eq:grad_L1_h}-\eqref{eq:grad_L1_Omega} into~\eqref{eq:update_h}-\eqref{eq:update_Omega}, and we can get the update rules for $q(\bm\kappa_j^{(1:N)})$ as in~\eqref{eq:node_potential_kappa}-\eqref{eq:edge_potential_kappa} in Table~\ref{tab:update_rule_BADGE}.

\begin{algorithm}
\footnotesize
  \caption{BASS} \label{alg:BASS}
  \begin{algorithmic}[0]
  \State \textbf{Input:} The observations $\bm x_{1:P}^{(1:N)}$ and the number of iterations for annealing $N_a$.
  \State \textbf{Output:} The estimated precision matrices $\langle K^{(1:N)}\rangle$.
  \State Initialize all variational distributions and set $R^{\{1\}} = 0$.
  \State - Compute $\langle K_{j,-j}^{(t)} \rangle\bm x_{-j}^{(t)}$ for $j = 1, \cdots, P$ and $t = 1, \cdots, N$.
  \For {i = 1 to $N_a$}
    \State - Compute $w = (1 - R^{\{i\}}) N / 2$.
    \State - Generate $\tilde{\bm x}_{1:P}^{(1:N)}$ by bootstrapping from $\bm x_{1:P}^{(1:N)}$ with window width $w$.
    \State - Compute $\langle K_{j,-j}^{(t)} \rangle\tilde{\bm x}_{-j}^{(t)}$ for $j = 1, \cdots, P$ and $t = 1, \cdots, N$.
    \For {j = 1 to $P$}
        \For {k = j + 1 to $P$}
            \State -\ Compute $\langle K_{j,-jk}^{(t)}\rangle \bm x_{-jk}^{(t)} = \langle K_{j,-j}^{(t)} \rangle\bm x_{-j}^{(t)} - \langle K_{jk}^{(t)} \rangle\bm x_k^{(t)}$ and $\langle K_{k,-jk}^{(t)}\rangle \bm x_{-jk}^{(t)} = \langle K_{k,-k}^{(t)} \rangle\bm x_{-k}^{(t)} - \langle K_{jk}^{(t)} \rangle\bm x_k^{(t)}$.
            \State -\ Compute $\langle K_{j,-jk}^{(t)}\rangle\tilde{\bm x}_{-jk}^{(t)} = \langle K_{j,-j}^{(t)} \rangle\tilde{\bm x}_{-j}^{(t)} - \langle K_{jk}^{(t)} \rangle\tilde{\bm x}_k^{(t)}$ and $\langle K_{k,-jk}^{(t)}\rangle\tilde{\bm x}_{-jk}^{(t)} = \langle K_{k,-k}^{(t)} \rangle\tilde{\bm x}_{-k}^{(t)} - \langle K_{jk}^{(t)} \rangle\tilde{\bm x}_k^{(t)}$.
            \State -\ Compute $\partial \sL_1/\partial \langle K_{jk}^{(t)}\rangle$ and $\partial \sL_1/\partial \langle {K_{jk}^{(t)}}^2\rangle$ as in~\eqref{eq:d_L1_d_K} and~\eqref{eq:d_L1_d_K2}.
            \State -\ Compute $\partial \tilde\sL_1/\partial \langle K_{jk}^{(t)}\rangle$ and $\partial \tilde\sL_1/\partial \langle {K_{jk}^{(t)}}^2\rangle$ by replacing $\bm x_{1:P}^{(1:N)}$ in\eqref{eq:d_L1_d_K} and~\eqref{eq:d_L1_d_K2} with $\tilde{\bm x}_{1:P}^{(1:N)}$.
            \State -\ Compute the ELBO w.r.t. $q(s_{jk}^{(1:N)})$ as
            \State \ \ $\sL(q(s_{jk}^{(1:N)})) = \langle s_{jk}^{(1)}\rangle\langle\log\pi_1\rangle+ (1-\langle s_{jk}^{(1)}\rangle)\langle\log(1-\pi_1)\rangle+ \sum_{t=1}^N \langle s_{jk}^{(t)}\rangle (\langle J_{jk}^{(t)}\rangle \partial \sL_1/\partial \langle K_{jk}^{(t)}\rangle + \langle {J_{jk}^{(t)}}^2\rangle \partial \sL_1/\partial \langle {K_{jk}^{(t)}}^2\rangle) + $
            \State \ \ \qquad\qquad\qquad\quad\ $\sum_{t=2}^N [q(s_{jk}^{(t-1)} = 0, s_{jk}^{(t)} = 0)\langle\log A_{00}\rangle + q(s_{jk}^{(t-1)} = 0, s_{jk}^{(t)} = 1)\langle\log(1 - A_{00})\rangle +  $
            \State \ \ \qquad\qquad\qquad\quad\ $q(s_{jk}^{(t-1)} = 1, s_{jk}^{(t)} = 0)\langle\log(1- A_{11})\rangle+q(s_{jk}^{(t-1)} = 1, s_{jk}^{(t)} = 1)\langle\log A_{11}\rangle] +\sH(q(s_{jk}^{(1:N)}))$.
            \State -\ Sample $\tilde\pi_1 \sim q(\pi_1)$, $\tilde A_{00} \sim q(A_{00})$, and $\tilde A_{11} \sim q(A_{11})$.
            \State -\ Compute $q(\tilde s_{jk}^{(1:N)})$ by replacing in (14) $\bm x_{1:P}^{(1:N)}$ with $\tilde{\bm x}_{1:P}^{(1:N)}$, $\langle\log\pi_1\rangle$ with $R^{\{i\}}\langle\log\pi_1\rangle + (1 - R^{\{i\}})\log\tilde\pi_1$ and likewise for the remaining
            \State \ \ terms regarding $\pi_1$, $A_{00}$ and $A_{11}$.
            \State -\ Compute $q(\tilde s_{jk}^{(t)})$ and $q(\tilde s_{jk}^{(t)}, \tilde s_{jk}^{(t - 1)})$ for all $t$ via message passing and further evaluate $\sL(q(\tilde s_{jk}^{(1:N)}))$.
            \State -\ Set $q(s_{jk}^{(1:N)}) = q(\tilde s_{jk}^{(1:N)})$ with probability $\min(1, \exp((\sL(q(\tilde s_{jk}^{(1:N)})) - \sL(q(s_{jk}^{(1:N)}))) / (1 - R^{\{i\}})))$.
            \State -\ Compute the ELBO w.r.t. $q(J_{jk}^{(1:N)})$ as
            \State \ \ $\sL(q(J_{jk}^{(1:N)})) = \sum_{t=1}^N \langle s_{jk}^{(t)}\rangle (\langle J_{jk}^{(t)}\rangle \partial \sL_1/\partial \langle K_{jk}^{(t)}\rangle + \langle {J_{jk}^{(t)}}^2\rangle \partial \sL_1/\partial \langle {K_{jk}^{(t)}}^2\rangle) - \langle\alpha_{jk}\rangle / 2\sum_{t=2}^N\langle(J_{jk}^{(t)} - J_{jk}^{(t-1)})^2\rangle + \sH(q(J_{jk}^{(1:N)}))$.
            \State -\ Sample $\tilde\alpha_{jk}\sim q(\alpha_{jk})$.
            \State -\ Compute $q(\tilde J_{jk}^{(1:N)})$ by replacing in (18) $\bm x_{1:P}^{(1:N)}$ with $\tilde{\bm x}_{1:P}^{(1:N)}$ and $\langle\alpha_{jk}\rangle$ with $R^{\{i\}}\langle\alpha_{jk}\rangle + (1 - R^{\{i\}})\tilde\alpha_{jk}$.
            \State -\ Compute $q(\tilde J_{jk}^{(t)})$ and $q(\tilde J_{jk}^{(t)}, \tilde J_{jk}^{(t - 1)})$ for all $t$ via message passing and further evaluate $\sL(q(\tilde J_{jk}^{(1:N)}))$.
            \State -\ Set $q(J_{jk}^{(1:N)}) = q(\tilde J_{jk}^{(1:N)})$ with probability $\min(1, \exp((\sL(q(\tilde J_{jk}^{(1:N)})) - \sL(q(J_{jk}^{(1:N)}))) / (1 - R^{\{i\}})))$.
            \State -\ Update $q(\alpha_{jk})$ following~(19) if $q(J_{jk}^{(1:N)})$ is updated.
            \State -\ Update $\langle K_{j,-j}^{(t)} \rangle\bm x_{-j}^{(t)} = \langle K_{j,-jk}^{(t)}\rangle \bm x_{-jk}^{(t)} + \langle K_{jk}^{(t)} \rangle\bm x_k^{(t)}$ and likewise for $\langle K_{k,-k}^{(t)} \rangle\bm x_{-k}^{(t)}$, $\langle K_{j,-j}^{(t)} \rangle\tilde{\bm x}_{-j}^{(t)}$, and $\langle K_{k,-k}^{(t)} \rangle\tilde{\bm x}_{-k}^{(t)}$.
        \EndFor
    \EndFor
    \State -\ Update $q(\pi_1)$, $q(A_{00})$, and $q(A_{11})$ following~(15)-(17).
    \For {j = 1 to $P$}
        \State -\ Compute the ELBO w.r.t. $q(\kappa_j^{(1:N)})$ as
        \State \ \ $\sL(q(\kappa_j^{(1:N)})) = \sum_{t=1}^N [\langle\kappa_j^{(t)}\rangle / 2 - \langle\exp(\kappa_j^{(t)})\rangle {x_j^{(t)}}^2 / 2 - x_j^{(t)} \langle K_{j,-j}^{(t)}\rangle\bm x_{-j}^{(t)} -\langle\exp(-\kappa_j^{(t)})\rangle{\bm x_{-j}^{(t)}}'\langle K_{-j,j} K_{j,-j} \rangle\bm x_{-j}^{(t)} / 2] - \langle\beta\rangle / 2\sum_{t=2}^N\langle(\kappa_j^{(t)}-$
        \State \ \ \qquad\qquad\qquad\quad\ $\kappa_j^{(t-1)})^2\rangle + \sH(q(\kappa_j^{(1:N)}))$.
        \State -\ Determine the step size $\rho$ via line search along the direction of the exact gradient.
        \State -\ Sample $\tilde\beta \sim q(\beta)$.
        \State -\ Compute $q(\tilde \kappa_j^{(1:N)})$ by replacing in (20) $\bm x_{1:P}^{(1:N)}$ with $\tilde{\bm x}_{1:P}^{(1:N)}$ and $\langle\beta\rangle$ with $R^{\{i\}}\langle\beta\rangle + (1 - R^{\{i\}})\tilde\beta$.
        \State -\ Set $q(\kappa_j^{(1:N)}) = q(\tilde \kappa_j^{(1:N)})$ with probability $\min(1, \exp((\sL(q(\tilde \kappa_j^{(1:N)})) - \sL(q(\kappa_j^{(1:N)}))) / (1 - R^{\{i\}})))$.
    \EndFor
    \State -\ Update $q(\beta)$.
    \State -\ $R^{\{i + 1\}} = R^{\{i\}} - 10 / N_a$ if $i$ is divisible by $10$. Otherwise, $R^{\{i + 1\}} = R^{\{i\}}$.
  \EndFor
  \Repeat
    \For {j = 1 to $P$}
        \For {k = j + 1 to $P$}
            \State -\ Compute $\langle K_{j,-jk}^{(t)}\rangle \bm x_{-jk}^{(t)} = \langle K_{j,-j}^{(t)} \rangle\bm x_{-j}^{(t)} - \langle K_{jk}^{(t)} \rangle\bm x_k^{(t)}$ and $\langle K_{k,-jk}^{(t)}\rangle \bm x_{-jk}^{(t)} = \langle K_{k,-k}^{(t)} \rangle\bm x_{-k}^{(t)} - \langle K_{jk}^{(t)} \rangle\bm x_k^{(t)}$.
            \State -\ Update $q(s_{jk}^{(1:N)})$ following~(14) and compute $q(s_{jk}^{(t)})$ and $q(s_{jk}^{(t)}, s_{jk}^{(t - 1)})$ for all $t$ via message passing.
            \State -\ Update $q(J_{jk}^{(1:N)})$ following~(18) and compute $q(J_{jk}^{(t)})$ and $q(J_{jk}^{(t)}, J_{jk}^{(t - 1)})$ for all $t$ via message passing.
            \State -\ Update $q(\alpha_{jk})$ following~(19).
            \State -\ Update $\langle K_{j,-j}^{(t)} \rangle\bm x_{-j}^{(t)} = \langle K_{j,-jk}^{(t)}\rangle \bm x_{-jk}^{(t)} + \langle K_{jk}^{(t)} \rangle\bm x_k^{(t)}$ and likewise for $\langle K_{k,-k}^{(t)} \rangle\bm x_{-k}^{(t)}$.
        \EndFor
    \EndFor
    \State -\ Update $q(\pi_1)$, $q(A_{00})$, and $q(A_{11})$ following~(15)-(17).
    \For {j = 1 to $P$}
        \State -\ Determine the step size $\rho$ via line search along the direction of the exact gradient such that $\sL(q(\kappa_j^{(1:N)}))$ is maximized.
        \State -\ Update $q(\kappa_j^{(1:N)})$ following (20) and compute $q(\kappa_j^{(t)})$ and $q(\kappa_j^{(t)}, \kappa_j^{(t - 1)})$ for all $t$ via message passing.
    \EndFor
    \State -\ Update $q(\beta)$.
  \Until convergence.
  \State -\ Compute $\langle K_{jk}^{(t)}\rangle = \langle s_{jk}^{(t)}\rangle \langle J_{jk}^{(t)}\rangle$ for $j\neq k$ and for all $t$, and $\langle K_{jj}^{(t)}\rangle = \langle\exp(\kappa_j^{(t)})\rangle$ for all $j$ and $t$.
  \end{algorithmic}
\end{algorithm}

\newpage
\newpage

\begingroup\abovedisplayskip=-0.1pt \belowdisplayskip=-0.1pt 
\begin{table}
\caption{Detailed update rules of BASS for inferring graphical models for stationary time series in the frequency domain.}
\label{tab:update_rule_BADGE}
 \fontsize{6.5}{7}\selectfont 
\begin{tabular}{m{0.976\linewidth}}
\hline
{\begin{align}
&\frac{\partial \sL_1}{\partial \langle K_{jk}^{(\omega)}\rangle} =  - 2 \overline{f_j^{(\omega)}}f_k^{(\omega)} -\langle\exp(-\kappa_j^{(\omega)})\rangle  f_{k}^{(\omega)}\big(\langle \overline{K_{j,-jk}^{(\omega)}}\rangle \overline{\bm f_{-jk}^{(\omega)}}\big)
-\langle\exp(-\kappa_k^{(\omega)})\rangle \overline{f_{j}^{(\omega)}}\big(\langle K_{k,-jk}^{(\omega)}\rangle \bm f_{-jk}^{(\omega)}\big). \label{eq:d_L1_d_K}\\
&\frac{\partial \sL_1}{\partial \langle K_{jk}^{(\omega)}\overline{K_{jk}^{(\omega)}}\rangle} = -\langle\exp(-\kappa_j^{(\omega)})\rangle f_k^{(\omega)}\overline{f_k^{(\omega)}} - \langle\exp(-\kappa_k^{(\omega)})\rangle f_j^{(\omega)}\overline{f_j^{(\omega)}}.\label{eq:d_L1_d_K2} \\
&\varphi_\sV^s(s_{jk}^{(1)}) = \delta(s_{jk}^{(1)} = 1)\langle\log\pi_1\rangle+ \delta(s_{jk}^{(1)} = 0)\langle\log(1-\pi_1)\rangle- \bigg(\frac{\partial \sL_1}{\partial \langle K_{jk}^{(1)}\rangle}\langle J_{jk}^{(1)}\rangle +
\overline{\frac{\partial \sL_1}{\partial \langle K_{jk}^{(1)}\rangle}\langle J_{jk}^{(1)}\rangle}+ 2 \frac{\partial \sL_1}{\partial \langle K_{jk}^{(1)} \overline{K_{jk}^{(1)}}\rangle}\langle J_{jk}^{(1)}\overline{J_{jk}^{(1)}}\rangle\bigg)s_{jk}^{(1)}. \label{eq:node_potential_s1}\\
&\varphi_\sV^s(s_{jk}^{(\omega)}) = -\bigg(\frac{\partial \sL_1}{\partial \langle K_{jk}^{(\omega)}\rangle}\langle J_{jk}^{(\omega)}\rangle + \overline{\frac{\partial \sL_1}{\partial \langle K_{jk}^{(\omega)}\rangle}\langle J_{jk}^{(\omega)}\rangle} + 2 \frac{\partial \sL_1}{\partial \langle K_{jk}^{(\omega)}\overline{K_{jk}^{(\omega)}}\rangle}\langle J_{jk}^{(\omega)}\overline{J_{jk}^{(\omega)}}\rangle\bigg)s_{jk}^{(\omega)}, \qquad \forall \omega > 1. \\
&\varphi_\sE^s(s_{jk}^{(\omega)}, s_{jk}^{(\omega - 1)}) = \delta(s_{jk}^{(\omega - 1)} = 0, s_{jk}^{(\omega)} = 1)\langle\log A_{01}\rangle + \delta(s_{jk}^{(\omega - 1)} = 0, s_{jk}^{(\omega)} = 0)\langle\log(1 - A_{01})\rangle + \delta(s_{jk}^{(\omega - 1)} = 1, s_{jk}^{(\omega)} = 1)\langle\log A_{11}\rangle \notag \\
&\quad\quad + \delta(s_{jk}^{(\omega - 1)} = 1, s_{jk}^{(\omega)} = 0)\langle\log(1 - A_{11}).\label{eq:edge_potential_s} \\
&\varphi_\sV^J(J_{jk}^{(\omega)}) = \bigg(\frac{\partial \sL_1}{\partial \langle K_{jk}^{(\omega)}\overline{K_{jk}^{(\omega)}}\rangle}\langle s_{jk}^{(\omega)} \rangle -\langle\alpha_{jk}\rangle\bigg)J_{jk}^{(\omega)}\overline{J_{jk}^{(\omega)}} + \frac{\partial \sL_1}{\partial \langle K_{jk}^{(\omega)}\rangle} \langle s_{jk}^{(\omega)}\rangle J_{jk}^{(\omega)} + \overline{\frac{\partial \sL_1}{\partial \langle K_{jk}^{(\omega)}\rangle}} \langle s_{jk}^{(\omega)}\rangle \overline{J_{jk}^{(\omega)}}, \qquad\forall \omega\in\{1, N\}.  \label{eq:node_potential_J1}\\
&\varphi_\sV^J(J_{jk}^{(\omega)}) = \bigg(\frac{\partial \sL_1}{\partial \langle K_{jk}^{(\omega)}\overline{K_{jk}^{(\omega)}}\rangle}\langle s_{jk}^{(\omega)} \rangle -2\langle\alpha_{jk}\rangle\bigg)J_{jk}^{(\omega)}\overline{J_{jk}^{(\omega)}} + \frac{\partial \sL_1}{\partial \langle K_{jk}^{(\omega)}\rangle} \langle s_{jk}^{(\omega)}\rangle J_{jk}^{(\omega)} + \overline{\frac{\partial \sL_1}{\partial \langle K_{jk}^{(\omega)}\rangle}} \langle s_{jk}^{(\omega)}\rangle \overline{J_{jk}^{(\omega)}}, \qquad\forall 1 < \omega < N. \\
&\varphi_\sE^J(J_{jk}^{(\omega)}, J_{jk}^{(\omega - 1)}) = -\langle\alpha_{jk}\rangle\big( J_{jk}^{(\omega)}\overline{J_{jk}^{(\omega - 1)}} + \overline{J_{jk}^{(\omega)}}J_{jk}^{(\omega - 1)}\big). \label{eq:edge_potential_J} \\
&\frac{\sL_1}{\langle\exp(-\kappa_j^{(\omega)})\rangle} = -{\bm f_{-j}^{(\omega)}}^*\langle K_{-j,j}^{(\omega)} K_{j,-j}^{(\omega)} \rangle\bm f_{-j}^{(\omega)}. \\
&h_t^{\{i+1\}} = (1-\rho) h_t^{\{i\}} + \rho\Big[1 - f_j^{(\omega)}\overline{f_j^{(\omega)}} \langle\exp(\kappa_j^{(\omega)})\rangle(1 - \langle\kappa_j^{(\omega)}\rangle) -\frac{\sL_1}{\langle\exp(-\kappa_j^{(\omega)})\rangle}\langle\exp(-\kappa_j^{(\omega)})\rangle(1 + \langle\kappa_j^{(\omega)}\rangle)\Big]. \\
&\Omega_{t,t}^{\{i+1\}} = (1-\rho) \Omega_{t,t}^{\{i+1\}} + \rho\bigg[\langle\beta\rangle + f_j^{(\omega)}\overline{f_j^{(\omega)}}\langle\exp(\kappa_j^{(\omega)})\rangle   -\frac{\sL_1}{\langle\exp(-\kappa_j^{(\omega)})\rangle}\langle\exp(-\kappa_j^{(\omega)})\rangle\bigg],\qquad \omega\in\{1,N\}. \label{eq:node_potential_kappa1}\\
&\Omega_{t,t}^{\{i+1\}} = (1-\rho) \Omega_{t,t}^{\{i+1\}} + \rho\bigg[2\langle\beta\rangle + f_j^{(\omega)}\overline{f_j^{(\omega)}}\langle\exp(\kappa_j^{(\omega)})\rangle   -\frac{\sL_1}{\langle\exp(-\kappa_j^{(\omega)})\rangle}\langle\exp(-\kappa_j^{(\omega)})\rangle\bigg], \qquad 1<\omega<N.\\
&\Omega_{t,t-1}^{\{i+1\}} = (1-\rho)\Omega_{t,t-1}^{\{i+1\}} - \rho \langle\beta\rangle. \label{eq:edge_potential_kappa}
\end{align}}
\end{tabular}
\begin{tabular}{m{0.477\columnwidth}m{0.477\columnwidth}}
\begin{align}
q(\alpha_{jk}) = \text{Ga}\Big(N-1,\sum_{t=2}^N\langle (J_{jk}^{(t)}-J_{jk}^{(t-1)})^2\rangle\Big).
\end{align} &
\begin{align}
q(\beta) = \text{Ga}\bigg(\frac{(N - 1) P}{2}, \frac{\sum_{j = 1}^P\sum_{t = 2}^N \langle(\kappa_j^{(t)} - \kappa_j^{(t-1)})^2\rangle}{2}\bigg).
\end{align} \\
\begin{align}
a = 1 + \sum_j\sum_k q(s_{jk}^{(1)} = 1). \label{eq:update_rule_pi}
\end{align} &
\begin{align}
b = 1 + \sum_j\sum_k q(s_{jk}^{(1)} = 0).
\end{align} \\
\begin{align}
c_0 = 1 +\sum_j\sum_k\sum_{t=2}^N q(s_{jk}^{(t)} = 0, s_{jk}^{(t-1)} = 0).
\end{align} &
\begin{align}
d_0 = 1 + \sum_j\sum_k\sum_{t=2}^N q(s_{jk}^{(t)} = 1, s_{jk}^{(t-1)} = 0).
\end{align} \\
\begin{align}
c_1 = 1 + \sum_j\sum_k\sum_{t=2}^N q(s_{jk}^{(t)} = 1, s_{jk}^{(t-1)} = 1).
\end{align} &
\begin{align}
d_1 = 1 +\sum_j\sum_k\sum_{t=2}^N q(s_{jk}^{(t)} = 0, s_{jk}^{(t-1)} = 1). \label{eq:update_rule_A11}
\end{align} \\
\hline
\end{tabular}
\end{table}
\endgroup

\section{Description of the AD Data}

The first data set contains 22 patients with mild cognitive impairment (MCI, a.k.a. pre-dementia) and 38 healthy control subjects~\cite{DVMC:2010}. The patients complained of the memory problem, and later on, they all developed mild AD (i.e., the first stage of AD). The ages of the two groups are $71.9\pm 10.2$ and $71.7\pm 8.3$, respectively. We provide some more details in the recording setup. Ag/AgCl electrodes (disks of diameter 8mm) were placed on 21 sites according to the 10-20 international system, with the reference electrode on the right ear-lobe. EEG was recorded with Biotop 6R12 (NEC San-ei, Tokyo, Japan) at a sampling rate of 200Hz.

The second data set consists of 17 patients with mild AD and 24 control subjects~\cite{HIHGOWDV:2006}. The ages of the two groups are $77.6\pm 10.0$ and $69.4\pm 11.5$, respectively. The patient group underwent full battery of cognitive tests (Mini Mental State Examination, Rey Auditory Verbal Learning Test, Benton Visual Retention Test, and memory recall tests). The EEG time series were recorded using 21 electrodes positioned according to Maudsley system, similar to the 10-20 international system, at a sampling frequency of 128 Hz.


\begin{thebibliography}{1}
\bibitem{YD:2016}
H. Yu, and J. Dauwels, ``Variational Bayes Learning of Time-Varying Graphcial Models,'' \emph{Proc. MLSP}, 2016.

\bibitem{KF:2009}
D. Koller and N. Friedman, ``Probabilistic Graphical Models,'' \emph{The MIT Press}, 2009.

\bibitem{MB:2006}
N. Meinshausen, and B. Bühlmann, ``High-dimensional graphs and variable selection with the lasso,'' \emph{The annals of statistics}, vol. 34, no. 3, pp. 1436-1462, 2006.

\bibitem{Yuan:2010}
M. Yuan, ``High dimensional inverse covariance matrix estimation via linear programming,'' \emph{J. Mach. Learn. Res.}, vol. 11, pp. 2261-2286, 2010.

\bibitem{SZ:2013}
T. Sun, and C. H. Zhang, ``Sparse matrix inversion with scaled lasso,'' \emph{J. Mach. Learn. Res.}, vol. 14, no. 1, pp. 3385-3418, 2013.

\bibitem{ODKB:2014}
S. Oh, O. Dalal, K. Khare, and B. Rajaratnam, ``Optimization methods for sparse pseudo-likelihood graphical model selection,'' In \emph{Proc. NIPS}, pp. 667-675, 2014.

\bibitem{LW:2017}
H. Liu, and L. Wang, ``Tiger: A tuning-insensitive approach for optimally estimating gaussian graphical models,'' \emph{Electronic Journal of Statistics}, vol. 11, no. 1, pp. 241-294, 2017.

\bibitem{FHT:2008}
J. Friedman, T. Hastie, and R. Tibshirani, ``Sparse inverse covariance estimation with the graphical lasso,'' \emph{Biostatistics}, \textbf{9} (3): 432-441, 2008. 

\bibitem{HSDR:2014}
C. J. Hsieh, M. A. Sustik, I. S. Dhillon, and P. Ravikumar, ``QUIC: quadratic approximation for sparse inverse covariance estimation,'' \emph{J. Mach. Learn. Res.}, 15(1), 2911-2947, 2014.

\bibitem{YXD:2019}
H. Yu, L. Xin, and J. Dauwels, ``Variational Wishart Approximation for Graphical Model Selection: Monoscale and Multiscale Models,'' \emph{IEEE Trans. Signal Process.}, 67(24), 6468-6482, 2019.

\bibitem{YWXD:2020}
H. Yu, S. Wu, L. Xin, and J. Dauwels, ``Fast Bayesian inference of sparse networks with automatic sparsity determination,'' \emph{J. Mach. Learn. Res.}, 21(124), 1-54, 2020.

\bibitem{KEKZEC:2010}
M. A.\ Kramer, U. T.\ Eden, E. D.\ Kolaczyk, R.\ Zepeda, E. N.\ Eskandar, and S. S.\ Cash, ``Coalescence and Fragmentation of Cortical Networks during Focal Seizures'', \emph{J. Neurosci.}, \textbf{30} (30): 10076-10085, 2010. 

\bibitem{ZLW:2010}
S. Zhou, J. Lafferty, and L. Wasserman, ``Time varying undirected graphs,'' \emph{Mach. Learn.}, \textbf{80}: 295-319, 2010. 

\bibitem{KSAX:2010}
M. Kolar, L. Song, A. Ahmed, and E. P. Xing, ``Estimating Time-Varying Networks,'' \emph{Ann. Appl. Math.}, \textbf{4} (1): 94-123, 2010. 

\bibitem{KX:2011}
M. Kolar and E. P. Xing, ``On Time Varying Undirected Graphs,'' \emph{J. Mach. Learn. Res.}, \textbf{15} W\&CP, 2011. 

\bibitem{QH:2016}
H. Qiu and F. Han, ``Joint estimation of multiple graphical models from high dimensional time series,'' \emph{J. R. Statist. Soc.} B, \textbf{78} (2): 487-507, 2016. 

\bibitem{MHSLAM:2014}
R. P. Monti, P. Hellyer, D. Sharp, R. Leech, C. Anagnostopoulos, G. Montana, ``Estimating time-varying brain connectivity networks from functional MRI time series,'' \emph{NeuroImage}, \textbf{103}: 427-443, 2014. 

\bibitem{AX:2009}
A. Ahmed and E. P. Xing, ``Recovering time-varying networks of dependencies in social and biological studies,'' \emph{Proc. Natl. Acad. Sci. USA}, \textbf{106} (29): 11878-11883, 2009. 

\bibitem{GN:2014}
A. J. Gibberd and J. D. B. Nelson, ``High Dimensional Changepoint Detection with a Dynamic Graphical Lasso,'' \emph{Proc. ICASSP}, 2014.

\bibitem{YLSWY:2015}
S. Yang, Z. Lu, X. Shen, P. Wonka, and J. Ye, ``Fused Multiple Graphical Lasso,'' \emph{SIAM J. Optim.}, \textbf{25} (2): 916-943, 2015. 

\bibitem{GN:2015}
A. J. Gibberd and J. D. B. Nelson, ``Estimating Dynamic Graphical Models from Multivariate Time-Series Data,'' \emph{Proc. AALTD}, 2015.

\bibitem{WA:2015}
E. C. Wit and A. Abbruzzo, ``Inferring slowly-changing dynamic gene-regulatory networks,'' \emph{BMC bioinformatics}, vol. 16(S6), no. S5, 2015.

\bibitem{GN:2017}
A. J. Gibberd and J. D. Nelson, ``Regularized estimation of piecewise constant gaussian graphical models: The group-fused graphical lasso,'' \emph{Journal of Computational and Graphical Statistics}, vol. 26, no. 3, pp. 623-634, 2017.

\bibitem{HPBL:2017}
D. Hallac, Y. Park, S. Boyd, and J. Leskovec, ``Network inference via the time-varying graphical lasso,'' In \emph{Proc. SIGKDD}, pp. 205-213, 2017.

\bibitem{YP:2020}
J. Yang and J. Peng, ``Estimating time-varying graphical models,'' \emph{Journal of Computational and Graphical Statistics}, vol. 29, no. 1, pp. 191-202, 2020.

\bibitem{LRW:2010}
H.\ Liu, K.\ Roeder, and L.\ Wasserman, ``Stability Approach to Regularization Selection (StARS) for High Dimensional Graphical Models,''  \emph{Proc. NIPS}, 2010.

\bibitem{ZS:2013}
J. Ziniel and P. Schniter, ``Dynamic Compressive Sensing of Time-Varying Signals via Approximate Message Passing,'' \emph{IEEE Trans. Signal Process.}, vol. 61, no. 21, pp. 5270-5284, 2013.

\bibitem{AWH:2014}
M. R. Andersen, O. Winther, and L. K. Hansen, ``Bayesian Inference for Structured Spike and Slab Priors,'' \emph{Proc. NIPS}, 2014.

\bibitem{YDJ:2014}
H. Yu, J. Dauwels, and P. Johnathan, ``Extreme-Value Graphical Models with Multiple Covariates,'' \emph{IEEE Trans. Signal Process.}, \textbf{62} (21): 5734-5747, 2014. 

\bibitem{GZP:2018}
S. Gultekin, A. Zhang, and J. Paisley, ``Asymptotic Simulated Annealing for Variational Inference,'' In \emph{2018 IEEE Global Communications Conference (GLOBECOM)}, pp. 1-7, 2018.

\bibitem{Dahlhaus:2000}
R. Dahlhaus, ''Graphical interaction models for multivariate time series,'' \emph{Metrika}, vol. 51, no. 2, pp. 157-172, 2000.

\bibitem{Matsuda:2006}
Y. Matsuda, ``A test statistic for graphical modelling of multivariate time series,'' \emph{Biometrika}, vol. 93 no. 2, pp. 399–409, 2006.

\bibitem{WW:2015}
R. J. Wolstenholme and A. T. Walden, ''An efficient approach to graphical modeling of time series,'' \emph{IEEE Trans. Signal Process.}, vol. 63, no. 12, pp. 3266-3276, 2015.

\bibitem{Tugnait:2019}
J. K. Tugnait, ``Edge Exclusion Tests for Graphical Model Selection: Complex Gaussian Vectors and Time Series,'' \emph{IEEE Trans. Signal Process.}, 67(19), 5062-5077, 2019.

\bibitem{Whittle:1953}
P. Whittle, ``The analysis of multiple stationary time series,'' \emph{J. R. Statist. Soc.} B, vol. 15, no. 1, pp. 125-139, 1953.

\bibitem{BJ:2004}
F. R. Bach and M. I. Jordan, ``Learning graphical models for stationary time series,'' \emph{IEEE Trans. Signal Process.}, vol. 52, no. 8, pp. 2189–2199, 2004.

\bibitem{JHG:2015}
A. Jung, G. Hannak, and N. Goertz, ``Graphical LASSO based Model Selection for Time Series,'' \emph{IEEE Signal Process. Lett.}, vol. 22, no. 10, pp. 1781-1785, 2015.

\bibitem{SV:2010}
J. Songsiri and L. Vandenberghe, ``Topology selection in graphical models of autoregressive processes,'' \emph{J. Mach. Learn. Res.}, vol. 11, pp. 2671-2705, 2010.

\bibitem{TFF:2015}
A. Tank, N. J. Foti, and E. B. Fox, E. B. ``Bayesian structure learning for stationary time series,'' In \emph{Proc. UAI}, pp. 872-881, 2015.

\bibitem{IR:2005}
H. Ishwaran and J. S. Rao, ``Spike and slab variable selection: Frequentist and Bayesian strategies,'' \emph{Ann. Stat.}, vol. 33, no. 2, pp. 730-773, 2005.


\bibitem{Bishop:2006}
C. Zhang, J. Butepage, H. Kjellström, and S. Mandt, ``Advances in variational inference,'' \emph{IEEE transactions on pattern analysis and machine intelligence}, 41(8), 2008-2026, 2019.





\bibitem{MB:2010}
N.\ Meinshausen, P.\ B\"{u}hlmann, ``Stability Selection,''  \emph{J. Roy. Statist. Soc. B - Stat. Methodol.}, vol.\ 72, pp.\ 417-473, 2010.

\bibitem{LHPW:2013}
S. Li, L. Hsu, J. Peng, and P. Wang, ``Bootstrap inference for network construction with an application to a breast cancer microarray study,'' \emph{Ann. Appl. Stat.}, vol. 7, no. 1, pp. 391-417, 2013.

\bibitem{DYWVLJC:2012}
J. Dauwels, H. Yu, X. Wang, F. Vialatte, C. Latchoumane, J. Jeong, and A. Cichocki, ``Inferring Brain Networks through Graphical Models with Hidden Variables'', \emph{Mach. Learn. \& Interpretation in Neuroimaging, Lecture Notes in Comput. Sci., Springer}, pp.\ 194-201, 2012.

\bibitem{BGLP:2012}
M. Billio, M. Getmansky, A. W Lo, and L. Pelizzon, ``Econometric measures of connectedness and systemic risk in the finance and insurance sectors,'' Journal of financial economics, 104(3):535–559, 2012.

\bibitem{Williams:2010}
M. Williams. ``Uncontrolled risk: lessons of Lehman brothers and how systemic risk can still bring down the world financial system,'' McGraw Hill Professional, 2010.

\bibitem{GY:2016}
P. Glasserman, and H. P. Young, H. P, ``Contagion in financial networks,'' Journal of Economic Literature, 54(3), 779-831, 2016.

\bibitem{DDLY:2018}
M. Demirer, F. X. Diebold, L. Liu, and K. Yilmaz, ``Estimating global bank network connectedness,'' Journal of Applied Econometrics, 33(1), 1-15, 2018.





\bibitem{Constancio:2012}
V. Constancio, ``Contagion and the European debt crisis,'' Financial Stability Review, 16, 109-121, 2012.








\bibitem{DVMC:2010}
J. Dauwels, F. Vialatte, T. Musha, and A. Cichocki, ``A comparative study of synchrony measures for the early diagnosis of Alzheimer's disease based on EEG,'' \emph{NeuroImage}, vol. 49, pp. 668-693, 2010.

\bibitem{HIHGOWDV:2006}
G. Henderson, E. Ifeachor, N. Hudson, C. Goh, N. Outram, S. Wimalaratna, C. Del Percio, and F. Vecchio, ``Development and assessment of methods for detecting dementia using the human electroencephalogram,'' \emph{IEEE Trans. Biom. Eng}, vol. 53, pp. 1557-1568, 2006.

\bibitem{YD:2018}
H. Yu and J. Dauwels. ``Modeling functional networks via piecewise-stationary graphical models,'' In \emph{Signal Process. and Mach. Learn. for Biomedical Big Data}, pp. 193–208. CRC Press, 2018.

\bibitem{YDW:2012}
H.\ Yu, J.\ Dauwels, and X.\ Wang, ``Copula Gaussian Graphical Models with Hidden Variables,'' \emph{Proceedings of ICASSP 2012}, pp.\ 2177--2180, 2012.






\end{thebibliography}

\begin{thebibliography}{1}

\bibitem{Bishop:2006}
C. Zhang, J. Butepage, H. Kjellström, and S. Mandt, ``Advances in variational inference,'' \emph{IEEE transactions on pattern analysis and machine intelligence}, 41(8), 2008-2026, 2019.

\bibitem{HBWP:2013}
M. D Hoffman, D. M Blei, C. Wang, and J. W. Paisley, ``Stochastic variational inference,'' \emph{Journal of Machine Learning Research}, vol. 14, no. 1, pp. 1303-1347, 2013.

\bibitem{HRKTK:2010}
A. Honkela, T. Raiko, M. Kuusela, M. Tornio, and J. Karhunen ``Approximate Riemannian conjugate gradient learning for fixed-form variational Bayes,'' \emph{Journal of Machine Learning Research}, vol. 11, pp. 3235-3268, 2010.

\bibitem{KBLSS:2016}
M. E. Khan, R. Babanezhad, W. Lin, M. Schmidt, and M. Sugiyama, ``Faster stochastic variational inference using Proximal-Gradient methods with general divergence functions,'' In \emph{Proceedings of the Thirty-Second Conference on Uncertainty in Artificial Intelligence}, pp. 319-328, 2016.

\bibitem{SRG:2002}
R. Salakhutdinov, S. Roweis, and Z. Ghahramani, ``On the convergence of bound optimization algorithms,'' In \emph{Proceedings of the Nineteenth conference on Uncertainty in Artificial Intelligence}, pp. 509-516, 2002.

\bibitem{DVMC:2010}
J. Dauwels, F. Vialatte, T. Musha, and A. Cichocki, ``A comparative study of synchrony measures for the early diagnosis of Alzheimer's disease based on EEG,'' \emph{NeuroImage}, vol. 49, pp. 668-693, 2010.

\bibitem{HIHGOWDV:2006}
G. Henderson, E. Ifeachor, N. Hudson, C. Goh, N. Outram, S. Wimalaratna, C. Del Percio, and F. Vecchio, ``Development and assessment of methods for detecting dementia using the human electroencephalogram,'' \emph{IEEE Trans. Biom. Eng}, vol. 53, pp. 1557-1568, 2006.

\end{thebibliography}
\end{document}